\documentclass[lettersize,journal]{IEEEtran}
\usepackage{amsmath,amsfonts}
\usepackage{array}
\usepackage[caption=false,font=normalsize,labelfont=sf,textfont=sf]{subfig}
\usepackage{textcomp}
\usepackage{stfloats}
\usepackage{url}
\usepackage{verbatim}
\usepackage{graphicx}
\usepackage{cite}
\usepackage{soul}
\usepackage{color,xcolor}
\soulregister{\cite}7
\soulregister{\ref}7
\usepackage{float}
\usepackage{algorithm}
\usepackage{algorithmic}

\usepackage{cite} %for dealing with citations
\usepackage{graphicx} %for dealing with figures.
\usepackage[numbers,sort&compress]{natbib}
\usepackage{booktabs}

\begin{document}

\title{RCP-RF{:} A Comprehensive Road-car-pedestrian Risk Management Framework based on Driving Risk Potential Field}

\author{Shuhang Tan,
        Zhiling Wang
        and Yan Zhong% <-this % stops a space
\thanks{Shuhang Tan was with Hefei Institutes of Physical Science, Chinese Academy of Sciences, Hefei, 230031, China and University of Science and Technology of China, Hefei, 230026, China, e-mail: stan9177@mail.ustc.edu.cn;}% <-this % stops a space
\thanks{Zhiling Wang was with Hefei Institutes of Physical Science, Chinese Academy of Sciences, Hefei, 230031, China, Anhui Engineering Laboratory for Intelligent Driving Technology, Hefei, 230031, China and Application and Innovation Research Institute of Robotics and Intelligent Manufacturing
C.A.S., Hefei, 230031, China, e-mail: zlwang@hfcas.ac.cn; Corresponding author;}% <-this % stops a space
\thanks{Yan Zhong was with University of Science and Technology of China,Hefei, 230026, China, e-mail: zhongyan@mail.ustc.edu.cn}}

\markboth{Journal of \LaTeX\ Class Files,~Vol.~14, No.~8, August~2021}%
{Shell \MakeLowercase{\textit{et al.}}: A Sample Article Using IEEEtran.cls for IEEE Journals}


\maketitle

\begin{abstract}
Recent years have witnessed the proliferation of traffic accidents, which led wide researches on Automated Vehicle (AV) technologies to reduce vehicle accidents, especially on risk assessment framework of AV technologies. However, existing time-based frameworks can not handle complex traffic scenarios and ignore the motion tendency influence of each moving objects on the risk distribution, leading to performance degradation. To address this problem, we novelly propose a comprehensive driving risk management framework named RCP-RF based on potential field theory under Connected and Automated Vehicles (CAV) environment, where the pedestrian risk metric is combined into a unified road-vehicle driving risk management framework. Different from existing algorithms, the motion tendency between ego and obstacle cars and the pedestrian factor are legitimately considered in the proposed framework, which can improve the performance of the driving risk model. Moreover, it requires only $O(N^2)$ of time complexity in the proposed method. Empirical studies validate the superiority of our proposed framework against state-of-the-art methods on real-world dataset NGSIM and real AV platform.
\end{abstract}


\section{Introduction}\label{sec1}
\IEEEPARstart{T}{here} are more than 1 million people died due to the traffic accidents each year \cite{536}, which is mainly caused by driver action error at the feeling, judgment or operation \cite{Singh-496}. In order to reduce traffic accidents, intelligent vehicles, as an important and foreseeable part of the whole transportation system driven by the government \cite{Eskandarian-497,XuWang-571}, have attracted wide attentions in recent years, which can lower the probability of accidents and save the society economic loss. Therefore, the researches of Autonomous Vehicle (AV)\footnote{AV is a highly integrated system, it carries different type of sensors to realize perception, localization, decision and control tasks \cite{Eskandarian-497,LiLi-494}.}  technologies are of great value to the creation of safety and efficient traffic environment, and they also need to increase people's acceptance of these technologies \cite{ZhangTao-538}. Thus, how to guarantee AVs' driving safety is a crucial problem, and this is why building an efficient driving risk management framework is necessary.

To guarantee the reliability of decision-making and the driving safety for AVs, many driving risk management frameworks\footnote{Risk assessment framework is the link between perception and decision module, it takes the results from perception module as inputs, then evaluates the risk using these information, the risk evaluation result is generated as the input of decision module, to help AV make the best decision.} have been developed, trying to evaluate the potential risk around ego-car in the nearby environment accurately and effectively. The most widely used risk indicator is time-based. For example, time-to-collision (TTC) \cite{Lee-542,KatrakazasQuddus-504,KimKum-512}, time headway (THW) \cite{Vogel-539,Van}, time-to-reaction (TTR) \cite{ZhangAntonsson-544,KellerDang-543}, etc., which can be simply and efficiently calculated and applied into real traffic environment, but may over-estimate the risk and lead to a too conservative behavior pattern\cite{LiLi-494}.

To avoid this problem, some researches began to explore Vehicle-to-Everything (V2X) facilities to achieve more accuracy and abundant perception results in AVs with the development of 5G communication. And some potential field-based risk evaluation frameworks are proposed in recent years \cite{HongyuChi-492}, where the more comprehensive environment information can be learned \cite{WangWu-396,WangWu-421,LiGan-423,LiGan-397,Mullakkal-BabuWang-402,Mullakkal,TianPei-498}. Different from time-based risk assessment indicator, potential field-based methods can evaluate the environment risk more comprehensively, since the interaction between all the traffic participants and ego-car can be considered during risk assessment process. However, there are still some drawbacks in these methods.

\romannumeral1. It is assumed that ego-car's motion state has no effect on the whole risk field in these existing methods, and the risk values of each obstacle only depend on their own motion state, which is contrary to the fact that ego-car's motion and obstacle cars' motion are dependable in the real traffic scenarios. Additionally, the motion tendency influence of each moving objects on the risk distribution is ignored in these methods, leading to unreliable results.

\romannumeral2. The pedestrian risk field cannot be learned effectively in existing AV risk field models. Although the motion tendency of pedestrians is disorderly, their risk to ego-car cannot be neglected since pedestrian is an important part of the transportation and they are the most vulnerable road users \cite{RasouliTsotsos-547}. 

\romannumeral3. These works choose the field force as the quantized risk metric in the decision module \cite{LiGan-397,TianPei-498,WuYan-477}. However, this parameter is hard to be calculated accurately by program without professional numerical software in real traffic situation, leading to performance degradation of risk assessment indicator.

To address these problems, we propose a new quantification risk indicator and a comprehensive Road-Car-Pedestrian Risk Field model named RCP-RF in this paper. To be specific, based on previous work \cite{LiGan-423,LiGan-397,TianPei-498}, we firstly introduce the cosine similarity into car risk model to calculate the relationship between relative location and velocity vector of the ego and obstacle cars, according to which the value of obstacle car risk can be revised. The relationship between the above two vectors can reflect the relative motion tendency among ego and other cars, which can reflect the risk better, especially in the car cut-in scenarios. Then, simulate to \cite{ShenRaksincharoensak-553}, a simplified pedestrian model is proposed to be embedded into the whole risk field framework, so that the pedestrian risk field can be learned effectively. Finally, in order to simplify the coding and calculation complexity, and inspired by Complementary Cumulative Distribution Function (CCDF) as the risk curve \cite{CouncilStudies-566}, we introduce this parameter into risk indicator in our proposed framework.

We summarize the main contributions of this paper as follows:

\begin{itemize}
\item We propose a comprehensive road-car-pedestrian risk field model that integrates the pedestrian risk into the AV risk field model. The proposed framework can evaluate the environment risk at a microscopic level, and the interaction between ego-car with main transportation participants can be learned to provide a more minute risk information for ego-car, which could improve the reliability of decision-making. 

\item A cosine similarity-based method is proposed to analyze the relative motion tendency between ego-car and obstacle cars. Besides, we choose the Gaussian distribution to generate accelerate samples randomly based on accelerate get at that time, which can address the uncertainties caused by error and delay of hardware and software. 

\item The CCDF curve is introduced into the risk metric in the proposed method, which can simplify the calculation complex when converting the abstract risk field into a concrete risk metric. And extensive experiments on multiple benchmark datasets are conducted to validate the efficiency and effectiveness of our proposed framework.
\end{itemize}

The rest of the paper is organized as follows. The related work is reviewed in Section \ref{sec2}, and Section \ref{sec3} presents our comprehensive road-car-pedestrian risk field framework in detail. Then we conduct extensive experiments in Section \ref{sec4} to verity the advantage of our method. Finally, the conclusion is drawn in Section \ref{sec5}. 

\section{Literature review}\label{sec2}
\subsection{Common AV collision avoidance strategies}\label{subsec2.1}
The common strategies for autonomous vehicles to avoid accidents can be divided into three categories, motion planning based, learning based and risk assessment based \cite{LiYang-475} method. 

Motion planning based methods usually predict the vehicles' trajectories based on different motion model. Lefèvre \cite{Lefevre-476} investigated three types of the motion models, physical-based, maneuver-based and interaction-aware motion model, and provided the trajectories prediction methods based on each motion model. Wendi \cite{WendiYahang-523} proposed a DBSCAN based trajectories prediction method using lidar data, which can achieve two-way warning to pedestrians and drivers. Zhang \cite{ZhangFisac-533} used a hybrid zero-sum dynamic game theory to develop a new trajectory planning framework, which to help AV avoid obstacles. A limitation of this method is that  it is non-linear and non-convex, so it is easy to become a NP-hard problem \cite{LiYang-475}.

Profit from the development of Artificial Intelligence, another popular strategies is learning based, especially the supervised learning. Zhu \cite{ZhuMa-473} introduced the Machine Learning (ML) model, which used XGBoost model to classify the risk level for cars at merging situations. Xu \cite{xu2017end} applied the FCN-LSTM model on a large-scale vehicle action data, which can be used to classify drivers' action and use the result to evaluate the environment risk to ego-car. The benefit of the learning method is that it can make the networks generate realistic behaviors of human drivers \cite{LiYang-475}. Li \cite{LLi-424} proposed a detailed review about learning based threat assessment method in recent years, including machine learning, deep learning and reinforcement learning. Moverover, with the development of deep learning, unsupervised learning is also arisen. Khaled \cite{BayoudhHamdaoui-567} proposed a transfer learning method, which can imporve the detection accuracy of traffic sign and road in different environment to help build driving assistance system. However, the main drawback of these methods is that they are data-driven, so they have an unsatisfactory result at some scenarios, where related data is difficult to be collected.

Risk assessment based method is another type that is used widely, and it can be divided into time-based and potential field-based. And this part will discuss related works about time-based methods, and potential-based methods will be discussed in the next subsection. Time-based methods evaluate the risk from the time perspective, and they are the most popular metrics, since their simple and efficient qualities \cite{LLi-424}. There are several time-based indicators, time-to-collision (TTC) is the most popular one. Lee \cite{Lee-542} proposed the TTC theory used for vehicle brake control since 1976. Then TTC has been improved with other technologies. Kim \cite{KimKum-512} integrated TTC algorithm with lane-based probabilistic motion prediction of surrounding vehicles to evaluate the long-term risk. Katrakazas \cite{KatrakazasQuddus-504} proposed an new integrated method, which combined Dynamic Bayesian Networks (DBN) with TTC to realizing estimating risk in real time. Besides TTC, there are also works researches other time-based indicators. Chen \cite{Chen-506} proposed a forward collision probability index (FCPI), which considers the driving behavior to assist the driver to avoid the forward collision accident in highway driving. Noh \cite{Noh-514} used the criterion called time-to-enter (TTE) to judge the risk at an intersection. Time-based methods are simple and efficient, but they are limited by fixed scenarios, which impedes their development.

\subsection{Potential field based methods}\label{subsec2.2}
Potential field theory was first introduced by Khatib \cite{O-432} and used for robotic route planning and collision avoidance. Sattel \cite{SattelBrandt-545} transferred potential field to automotive path planning applications. Compared to time-based theory, potential field can generate a more high-level driving strategy, which can consider the interaction between ego-car and other traffic participants. And Ni \cite{Ni-428,Ni-429,Ni-430} proved the feasibility that the potential field theory can be applied in traffic scenario, which gave the theoretical basis for further related researches.

Many researchers devoted to building a vehicle risk field model based on potential field. Wu \cite{WangWu-396,WangWu-421,Wu-0} proposed a driving-vehicle-road driving safety field (DSF) system, which considered the factors of vehicle characteristics, road conditions and driver behaviors. This work provided a systemic DSF model, and it became the foundation for many later works. However, their work is under the assumption that all traffic participants' information can be obtained accurately in real-time, which has not be realized nowadays. Li \cite{LiGan-423,LiGan-397} proposed a model can be used for both car-following and cut-in scenario based on DSF theory. This model mainly considers lane boundary and obstacle cars' characteristics, and the model parameters are calibrated using NGSIM I-80 datasets. Meanwhile, they used the field force as the risk indicator to interpret the risk value. But they ignore the motion interaction between ego and obstacle cars, and only describe the static scenario. Meng \cite{Mullakkal-BabuWang-402,Mullakkal} built car risk model considering cars' physical motion laws. Although this helps the risk model more robust, the cost of realizing the model is huge and this model can not be applied in real traffic environment. Tian \cite{TianPei-498} considered the impacts of vehicle geometry on risk model, and used ellipse model to appropriately reflect the vehicle shape to improve the effect of the model, but the geometry relationship calculation is complex. 

All of the models mentioned above neither consider the interaction between ego and obstacle cars nor are efficient enough to apply in real traffic environment. So aiming at these defects, this paper introduces cosine similarity into the car risk model, to analyze the relative motion tendency between ego-car and obstacle cars. Also, CCDF risk curve is used to imply the environment risk, which is a more feasible metric than the current common risk indicator, risk field force.  Meanwhile, to our best knowledge, this paper is the first to integrate a specific form of pedestrian risk model into the whole driving risk management framework. The final framework performance is evaluated using real world open source vehicle dataset NGSIM. And compared to previous Li's \cite{LiGan-423} work, our model is more sensitive to the fine change of vehicles' motion state. Meanwhile, pedestrian risk metric and CCDF-based vehicle risk metric are also evaluated and proved to be valid compared to previous related work.

\section{Proposed  RCP-RF driving risk field model}\label{sec3}
\begin{figure}[h]%
\centering
\includegraphics[width=\linewidth]{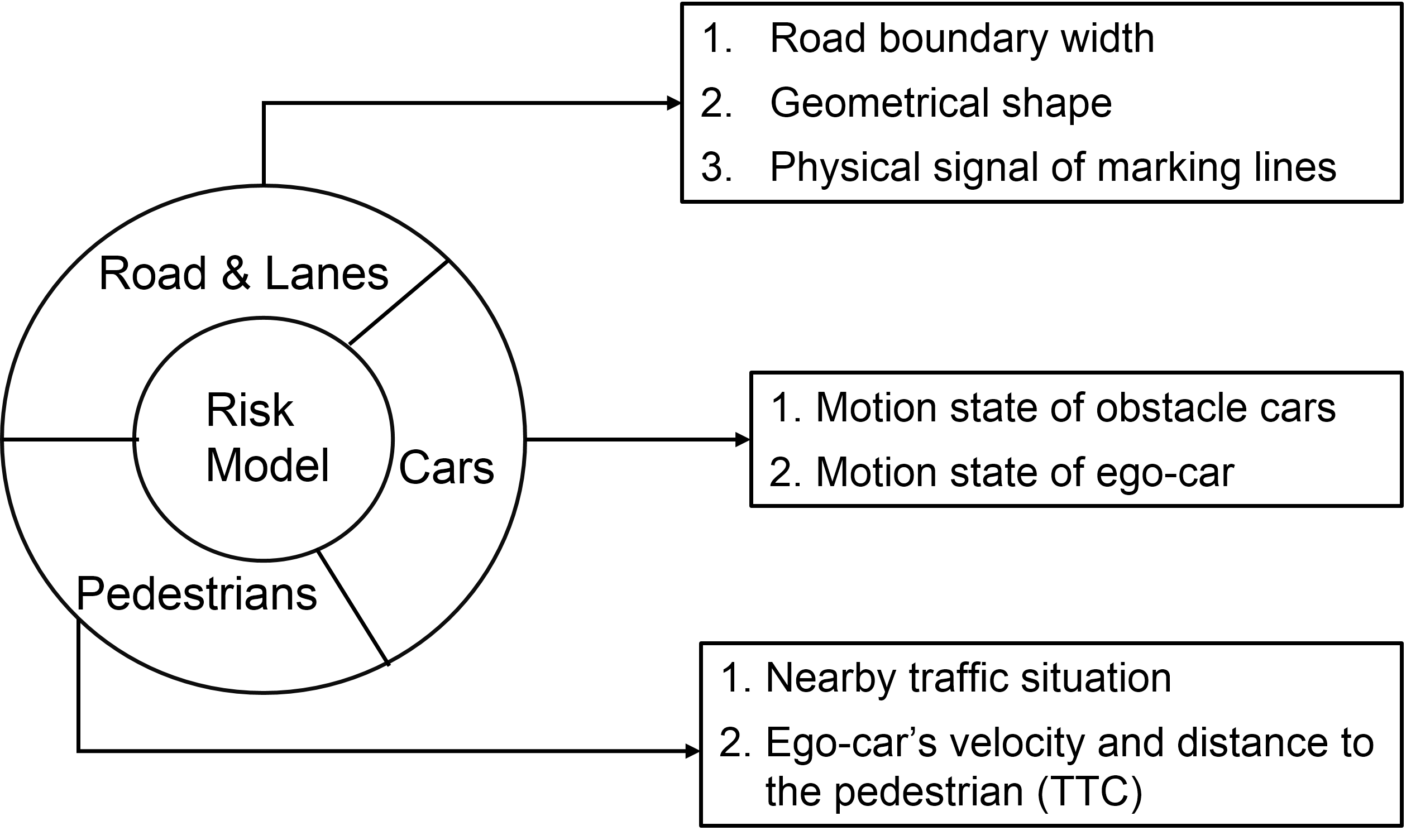}
\caption{Framework of the RCP-RF model}\label{fig1}
\end{figure}

The RCP-RF driving risk field model proposed in this paper is divided into three parts, roads, cars and pedestrians. Fig. \ref{fig1} shows the whole framework of the model, and the main factors that should be considered when estimating each object's risk model is given. 

The width and the geometrical shape of the lane boundary are the main characteristics in the lane risk model. Besides, the marking lines also show important signal in the urban traffic environment. To simplify the model, we only consider two kinds of marking lines \cite{LiGan-397}, one is the white dotted line which is used to separate vehicles running in the same direction, and the vehicles are allowed to go through the line to change lanes. The other one is double amber line that divides vehicles running in the opposite direction. Generally, vehicles are not allowed to travel across the double amber line. There are less attributes for car models, which need to be considered are ego and obstacle cars' motion state, including velocity, accelerate and position. For pedestrian model, this paper proposes a pedestrian risk metric which combines nearby traffic situation and ego-car's velocity and distance to the persons when they appears.

\begin{figure}[h]%
\centering
\includegraphics[width=\linewidth]{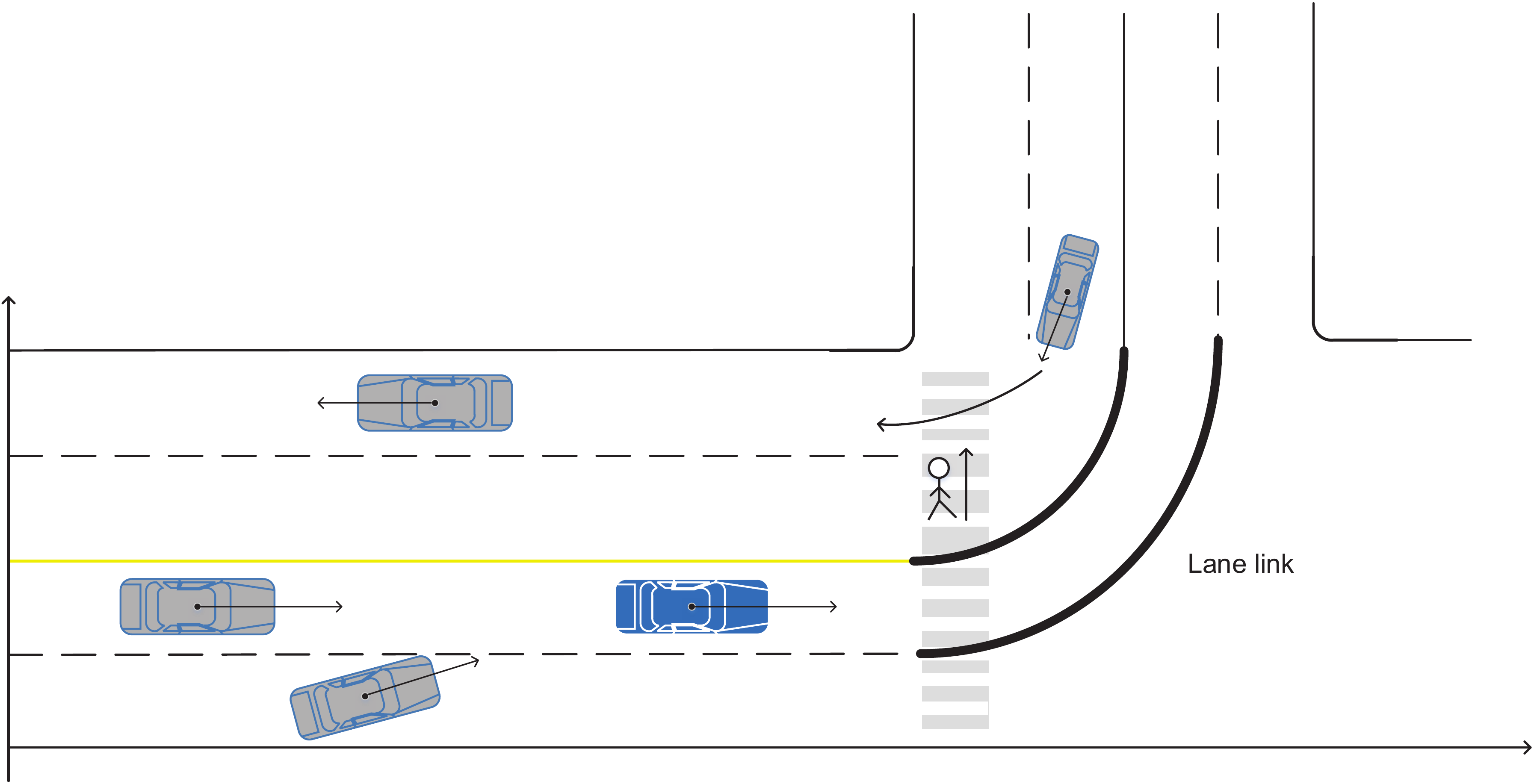}
\caption{Common urban traffic scenario}\label{fig2}
\end{figure}

Fig. \ref{fig2} shows a common traffic scenario in a urban road crossing, including roads, cars and pedestrians. To make risk model proposed in this paper can describe the general traffic scene, some special cases need to be generalized. First, when vehicles approach the intersection, the lane link has different attributes with the previous road, to simplify the model, the lane link is expressed as a lane rotates certain angle, and it is in the road part. Moreover, besides moving vehicles, there are static obstacles in the road, such as fault vehicles or roadblocks. Considering the similar characteristics between them and the vehicles, they are treated as the vehicle with $0$ velocity in the car part.

The following sections will give more specific description of each part. The final driving risk field is defined as the weighted sum of the risk strength generated by different traffic participants. Thus, the total risk field $E_{\text {total }}$ can be expressed as:

\begin{equation}
\lvert E_{\text {total }}\rvert = {\alpha}_{R}\lvert E_{R}\rvert + {\alpha}_{V}\lvert E_{V}\rvert  + {\alpha}_{P}\lvert E_{P} \rvert
\label{eq1}
\end{equation}

where $E_{\text{R}}$, $E_{\text{V}}$, $E_{\text{P}}$ represent the risk strength generated by roads, vehicles and pedestrians separately, and ${\alpha}_{R}$, ${\alpha}_{V}$, ${\alpha}_{P}$ are the weight factor for each objects' risk influence.

\subsection{Road risk potential field model}\label{subsec3.1}
\begin{figure}[h]%
\centering
\includegraphics[width=\linewidth]{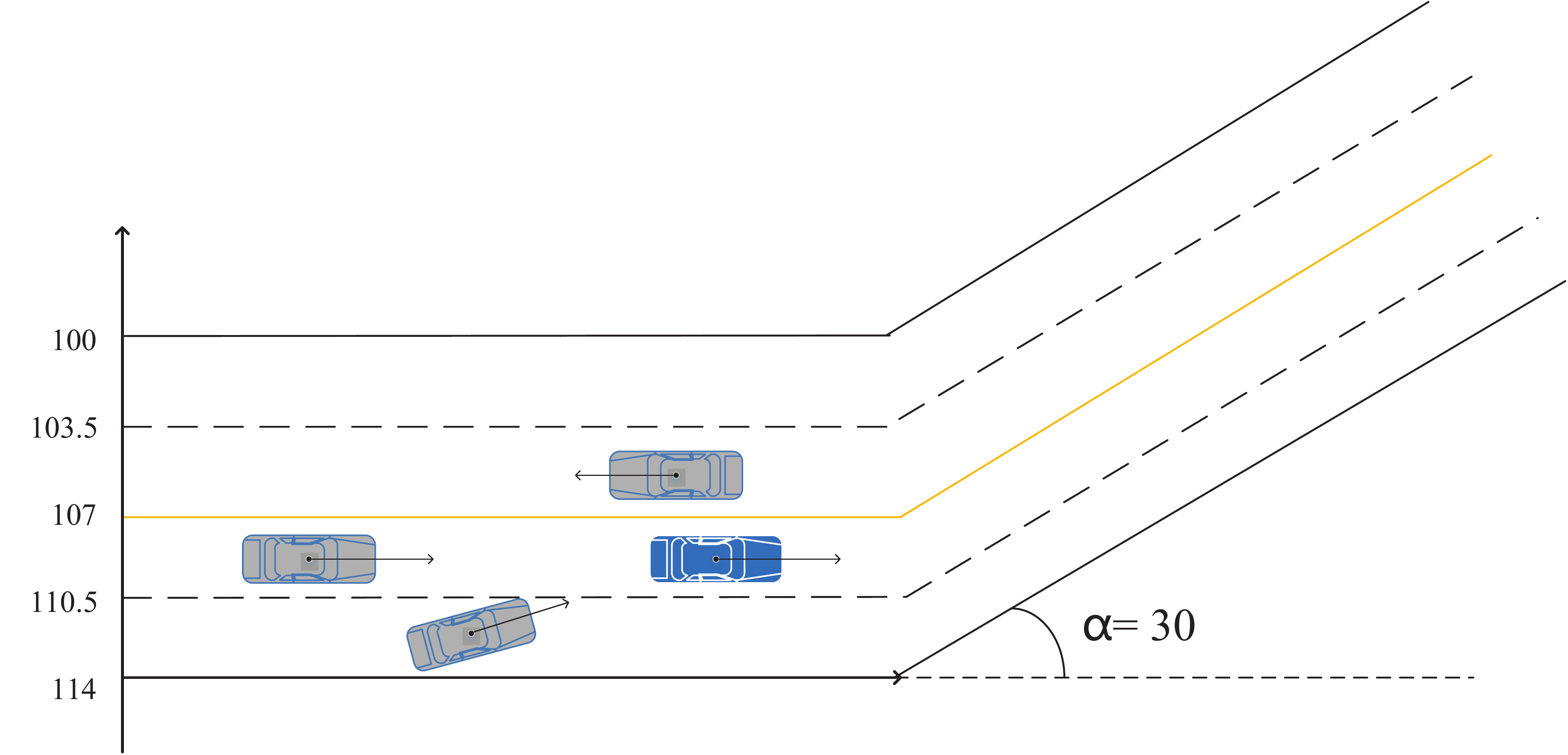}
\caption{Illustration of a simple traffic environment}\label{fig3}
\end{figure}

The whole lane risk model has two parts, the first one involves the the width of the lane boundary and the simple road geometrical shape to describe the general straightway and lane link when vehicles approach the crossing. And the second one focuses on the risk strength of marking lines. Fig. \ref{fig3} shows a schematic overview of simple road situation.

For lane model, this paper assumes that the risk field generated by the boundary and marking lines are only depended on their own  attributes,  and their risk influence on ego-car is only related to the distance between ego and specific boundary. The goal of the lane risk field model is to help ego-car keep the lane when driving.

\subsubsection{Road boundary risk model}\label{subsubsec3.1.1}
As mentioned above, the lane boundary model will consider the width of the road boundary and the angle of the lane. And for generalization, the lane boundary can be formed by any physical blocks, such as static road obstacles or isolation belts in the middle of the road. Using the road example in Fig. \ref{fig3}, the lane can be divided into two parts, the first part is a straight lane, whose angle is  $0^{\circ}$, while the second part has $30^{\circ}$ angle. In this part, we only build the model for the boundary, and the next section will give the model for marking lanes. 

The boundary risk tends to go to infinity when vehicles approaching gradually, the purpose of the model is to prevent vehicles collision with the boundary. Without considering the angle of the road, the lane boundary model function is built using an exponential function, and it can be expressed as:
\begin{equation}
E_{b}=\sum_{i=1}^{2} Amp_{bi} \cdot (e^{\frac{\lvert dis_{bi} - t_{bi}\rvert}{k_{b}}}-1).
\label{eq 2}
\end{equation}
where $Amp_{bi}$ controls the the amplitude of the risk peak value of the specific boundary, $dis_{bi}$ represents the distance from ego to that specific boundary, $t_{bi}$ is a empirical potential risk distance coefficient, if the distance from ego to the specific boundary is smaller than this threshold, it means the ego-car is entering the danger zone. $k_{b}$ is coefficient that determines the change rate of the risk field strength when ego approaches or leaves the boundary. Considering the exponential function's value range is in $[1, \infty)$ when independent variable is bigger or equal to $0$, so the expression minus one to convert the value range to $[0, \infty)$. Usually, a road has two boundaries, so the total boundary risk field is the summation of two boundaries' risk field.

\begin{figure}[h]%
\centering
\includegraphics[width=\linewidth]{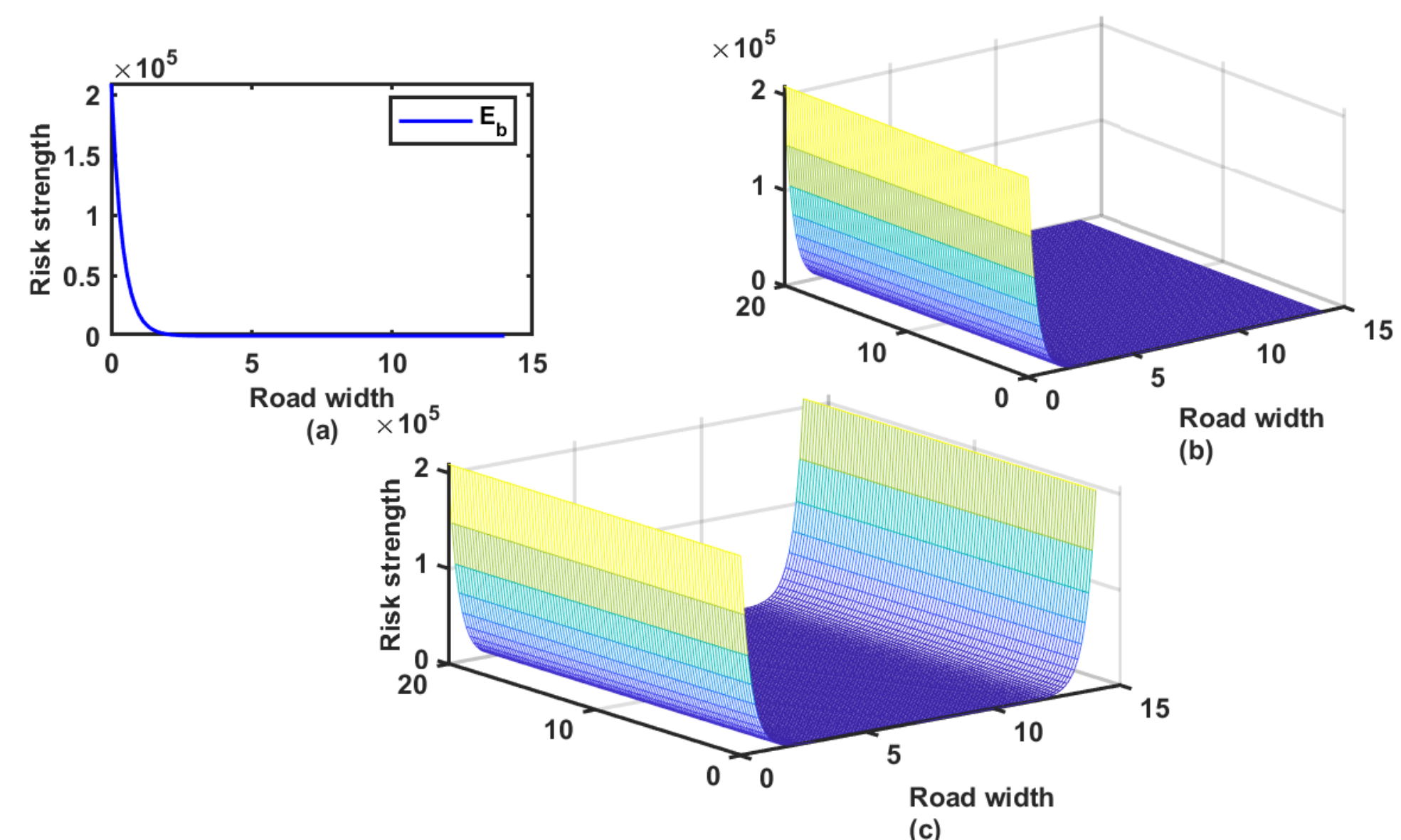}
\caption{Results of the lane risk model. (a) and (b) are the 2D and 3D format risk distribution results of single lane boundary; (c) is the 3D format risk distribution result of dual road boundaries.}\label{fig4}
\end{figure}

Using the example in Fig. \ref{fig3}, Fig. \ref{fig4} shows the results of the lane risk model for the first part of that example.
 In Fig. \ref{fig4},  $(a)$ and $(b)$ show the results of single boundary risk change tendency in both 2-D and 3-D format, and $(c)$ gives the 3-D format result for dual road boundaries. As these pictures show, the risk strength increases rapidly when car approaches the boundary, and decreases when vehicle moving away from the boundary, which is in consistent with common sense.
 
 Moreover, as shown in Fig.\ref{fig3}, the second part of the road has $30^{\circ}$ angle, to fit the shape of the road better, the model function is modified and the parameter $dis_{bi}$ and $t_{bi}$ need to be converted according to the angle.

\begin{figure}[h]%
\centering
\includegraphics[width=\linewidth]{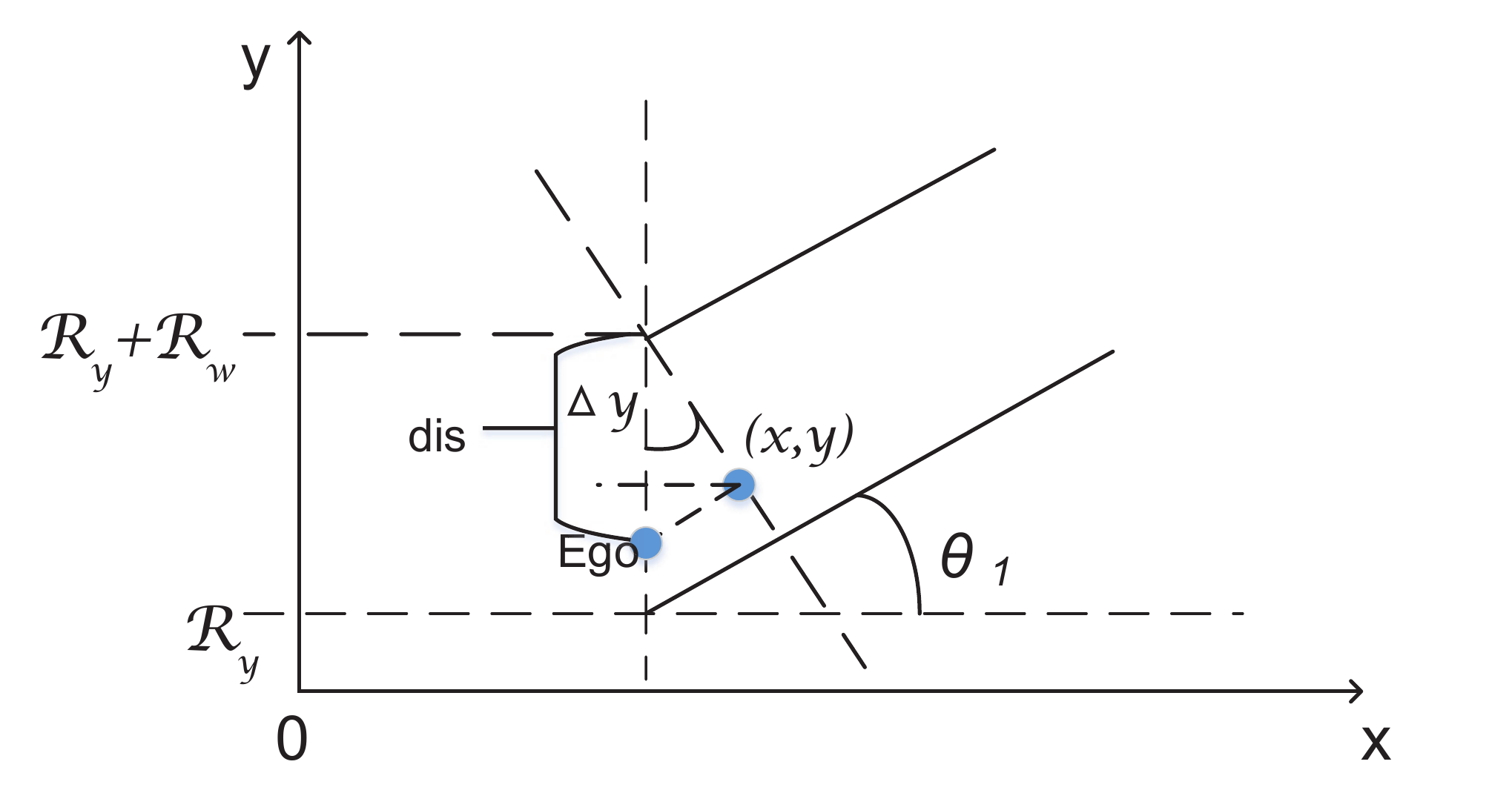}
\caption{Coordinates conversion illustration for lane model.}\label{fig5}
\end{figure}
 
 As shown in Fig. \ref{fig5}, the width of the road is $R_w$, angle is ${\theta 1}$. When ego car comes to the entrance of the second road, according to the trigonometric function, $y$ can be calculated by $y=R_y+R_w-\Delta y$, where $\Delta y = dis \cdot \cos \theta 1 \cdot \cos \theta 1 $. Thus, the conversion of $dis_{bm}$ and $t_{bm}$ can be generally expressed by equation \ref{eq 3} and \ref{eq 4}, where $R_{y}$ is road starts location coordinate, $R_{w}$ is the road width, ${\theta 1}$ is the road rotates angle to the horizontal line.

\begin{equation}
dis_{bim}=R_{y}+R_{w}-dis_{bi}\cdot \cos{\theta_1} \cdot \cos{\theta_1}
\label{eq 3}
\end{equation} 

\begin{equation}
t_{bim}=R_{y}+R_{w}-t_{bi}\cdot \cos{\theta_1} \cdot \cos{\theta_1}
\label{eq 4}
\end{equation} 

\begin{equation}
E_{b}=\sum_{i=1}^{2} Amp_{bi} \cdot (e^{\frac{\lvert dis_{bim} - t_{bim} \rvert}{k_{b}}}-1).
\label{eq 5}
\end{equation}

Equation \ref{eq 5} is the boundary risk model considering the rotated angle, which can be applied to model the line link at road crossing. $dis_{bim}$ and $t_{bim}$ represent the value after conversion using equation \ref{eq 3} and \ref{eq 4}.

\begin{figure}[h]%
\centering
\includegraphics[width=\linewidth]{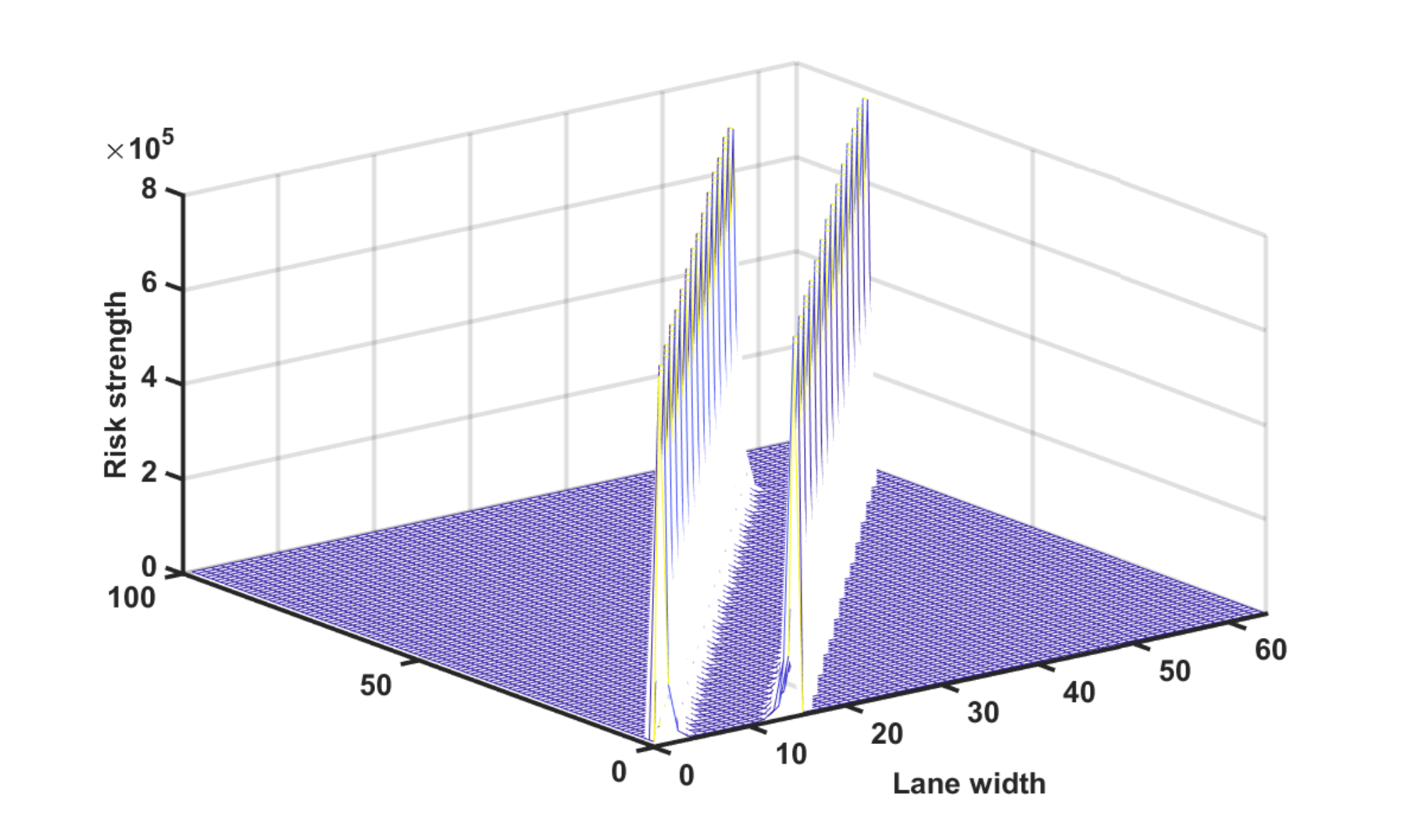}
\caption{Road risk model after coordinate conversion, $30^{\circ}$ rotation acts on the road risk field.}\label{fig6}
\end{figure}

Fig. \ref{fig6} shows the result of model considering coordinate conversion, from which can see clearly there is a $30^{\circ}$ rotation of the road risk field.
\subsubsection{Marking lines risk model}\label{subsubsec3.1.2}
For this part, the model deals with the physical  signal of marking lines, including white dotted line and  double amber line. The general expression for making line as followed equation \ref{eq 6}:
\begin{equation}
E_{l}=\sum_{i=1}^{n} Amp_{li} \cdot e^{\frac{-(dis_{li}-l_{posi})^{2}}{k_{l}}}
\label{eq 6}
\end{equation}

Where $Amp_{li}$ is the parameter controls the risk field amplitude related to different marking lines' physical signal, for example, the risk strength peak value of amber lines have a higher value than white dotted lines, so $Amp_{al}$ is bigger than $Amp_{wdl}$, which represent the coefficient value of amber lines and white dotted lines separately. $dis_{li}$ is the distance form ego to the specific marking line, and $l_{posi}$ is the position of specific marking line in image coordinate system. $k_{l}$ is coefficient that determines the change rate of the risk field strength when ego approaches or leaves the specific marking line. The total $E_{l}$ is the summation of all the marking lines risk field, the $n$ depends on the number of detected marking lines. The total road risk strength $E_{R}$ is the superposition of risk field of boundary and marking lines in equation \ref{eq 7}:
\begin{equation}
E_{R} = E_{b} + E_{l}.
\label{eq 7}
\end{equation}

\begin{figure}[h]%
\centering
\includegraphics[width=\linewidth]{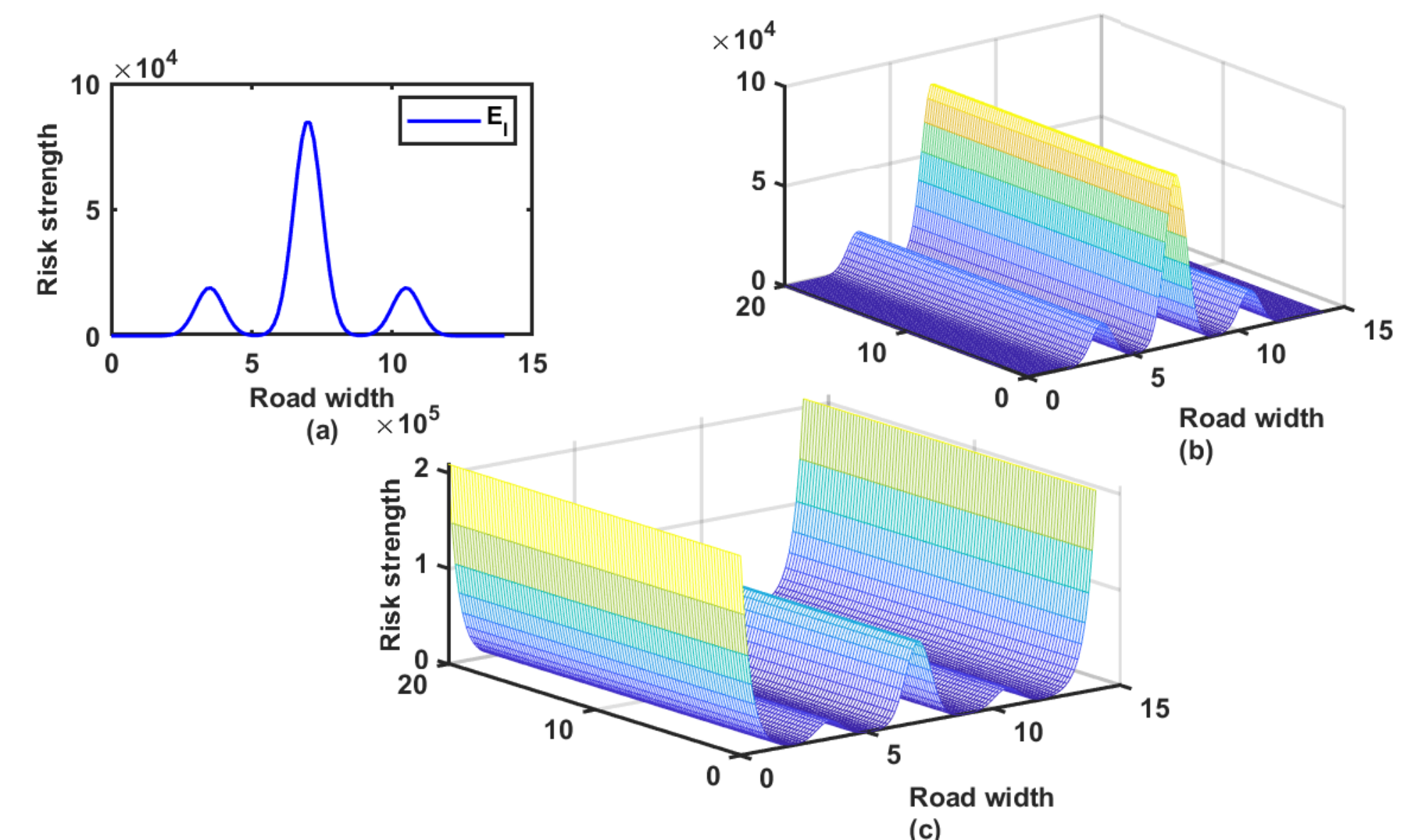}
\caption{Lane risk model. (a) and (b) are the risk distribution of marking lines; (c) is the total risk strength of lane model.}\label{fig7}
\end{figure}

Fig. \ref{fig7} $(a)$ and $(b)$ gives the results of marking lines' risk distribution and $(c)$ shows the total risk strength of lanes. Lane risk strength has peak value at the actual position of the boundary or marking lines, and decreases when vehicles approaching the middle of two lines. The peak risk strength of each line changes according to different signals of that line. Specifically, boundaries have the highest risk to ego-car, because it is a physical block. Cars are not allowed to cross the double amber line, so amber line has the second highest risk strength. White dotted line has the lowest risk to cars, since cars can cross the line to change lanes.

\subsection{Vehicle risk potential field model}\label{subsec3.2}
The vehicle risk potential field model is built to calculate the risk generated by obstacle cars to ego-car. The following sections will description the basic model formula and then give the detailed algorithm flow of how to calculate the vehicle risk strength.

\subsubsection{Vehicle risk model}\label{subsubsec3.2.1}
The input parameters of the model are the physical attributes and motion state of ego and obstacle cars, including dimension, location, velocity, accelerate and yaw for each car. Following will describe the detail of these parameters.

\romannumeral1. Vehicle physical attribute: \cite{Wu-0} acclaimed that the type and the mass of the vehicle may influence its the risk strength. Thus, they first proposed the virtual mass concept, which combines velocity and mass. Then, this concept is used in many paper \cite{LiGan-423,LiGan-397,WangWu-396}. However, precise vehicles' mass can only be obtain under perfect V2X environment, which have not realized nowadays. Therefore, the vehicle attribute used in this paper is only the dimension (e.g., width, length) of each vehicle, which can be attained based on current perception system.

\romannumeral2.  Motion state: The risk field generated  by specific vehicle to ego-car is closely related to their motion state. Specially, the different relative velocity and position will generate different risk distribution. Velocity $v$, accelerate $a$, heading angle $yaw$ and location (e.g., x, y) are all key factors affecting the risk field. This paper assumes the anticlockwise around the ego-car is positive direction.

\begin{figure}[h]%
\centering
\includegraphics[width=\linewidth]{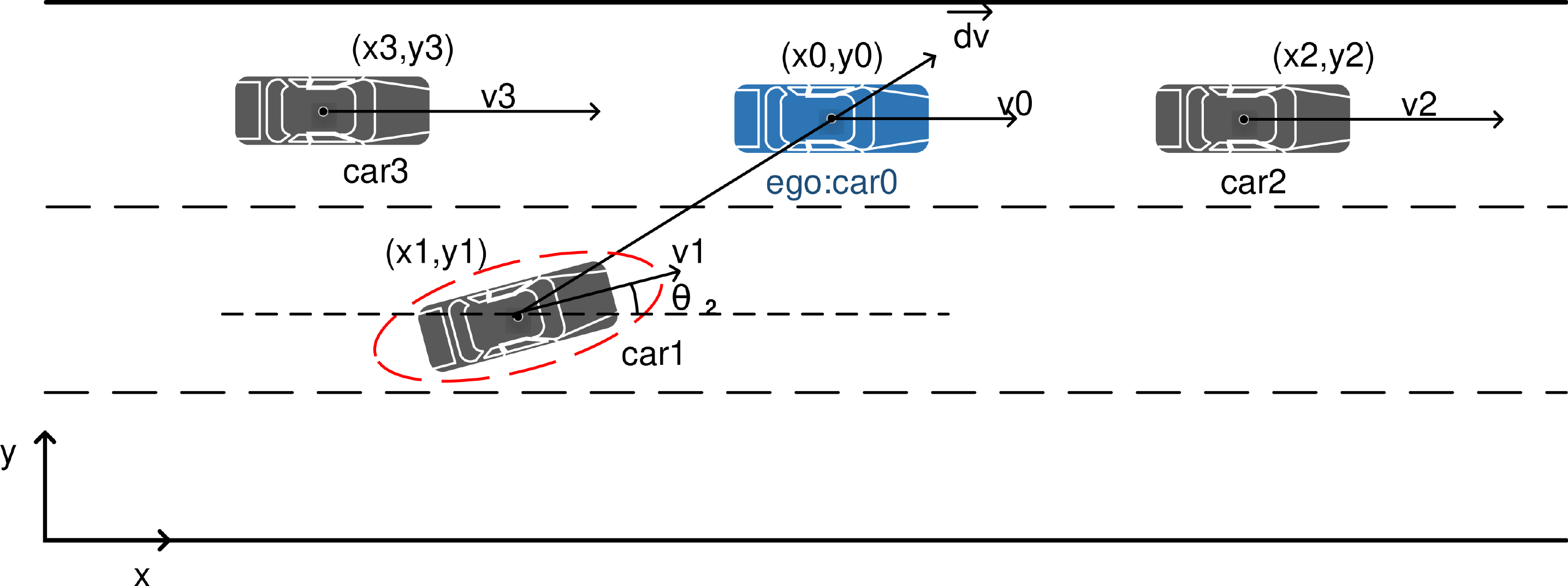}
\caption{Illustration of different vehicle motion states in the street.}\label{fig8}
\end{figure}

For vehicles which are in the same line with ego-car, and all of them do not have the tendency to change lines, the main factor affects the risk field is the moving direction and relative speed between ego and obstacle cars. In Fig. \ref{fig8}, assuming blue car as the ego, if the velocity of its preceding car is greater than its velocity, the risk generated by that preceding car is smaller than that if its velocity slower than ego-car's.

For vehicles are in the different line with ego-car, whether the obstacle car has intention to change lane is a key factor to affect the risk field distribution. This paper uses their relative velocity and position to estimate this intention. All of the motion state parameters should be considered in this part, and detailed model will be discussed in the following part.

The risk strength generated by obstacle vehicle to ego-car is inversely proportional to the distance from the center of that car to ego-car, the risk strength will increase when ego coming close to the obstacles. The traditional method to calculate the Euclid distance between any two points is 
\begin{equation}
dis_{p} = \sqrt{\left(x_{1}-x_{0}\right)^{2}+\left(y_{1}-y_{0}\right)^{2}}.
\label{equ 8}
\end{equation}

However, if we use $1/{dis}$ directly to represent the negative coorelation in car risk model, it can be found that the risk intensity is same at any points with the same distance to ego-car, no matter if the obstacles are in the same lane with ego or not, which violates the reality. In practice, the risk distribution should extend with the velocity direction of the obstacle cars, shown as the dotted line in Fig. \ref{fig8}. Meanwhile, relative motion tendency between obstacle and ego cars also have important effect on risk field. This paper uses cosine similarity between relative location and velocity vector of the ego and obstacle cars to estimate this tendency. Thus, to deal with the two factors mentioned above, this paper proposes a new virtual distance, which can be expressed using the following three equations:

\begin{equation}
dis_{virtual} = \sqrt{\frac{\left(x_{1}-x_{convt}\right)^{2}}{\left(obs_{l}\cdot k)^{2}\right.}+\frac{\left(y_{1}-y_{convt}\right)^{2}}{obs_{w}^{2}}},
\label{equ 9}
\end{equation}

\begin{equation}
    \begin{array}{l}
    x_{convt}=\cos (\theta_2) \cdot\left(x-x_{1}\right)-\sin (\theta_2) \cdot\left(y-y_{1}\right)+x_{1} \\
    y_{convt}=\sin (\theta_2) \cdot\left(x-x_{1}\right)+\cos (\theta_2) \cdot\left(y-y_{1}\right)+y_{1}
    \end{array}
    \label{equ 10}
\end{equation}

\begin{equation}
k=\left\{\begin{array}{l}
1+\log_2 \left(1+re_{v}\right), \quad \text { approaching } \\
(1+e^{-\frac{re_{v}}{ego_{v}}})* \frac{obs_w}{obs_l}, \quad \text { leaving }
\end{array}\right\}.
\label{equ 11}
\end{equation}

Equation \ref{equ 9} gives the general formula to calculate the virtual distance, where $obs_{l}, obs_{w}$ are the length and width of the obstacle vehicle that used to revise the risk distribution range shape; $k$ is the parameter that reflects the relative motion tendency, $x_{convt}$ and $y_{convt}$ are the coordinates rotates around the center of the obstacle car according to the heading angle. Equation \ref{equ 10} is the formula that convert the coordinates, where $\theta_2$ is the obstacle car's velocity angle to the road, $(x_{1}, y_{1})$ is the center coordinate of the obstacle car, $(x, y)$ is any points in the coordinate around the obstacle car. Equation \ref{equ 11} gives the selection strategy of value of $k$, $re_{v}$ is the relative velocity of ego and obstacle car, $ego_{v}$ is ego's velocity, $obs_{w}, obs_{l}$ are the width and length of obstacle car. $k$'s value is related to whether the obstacle has the moving tendency to approach the ego, the detail will be discussed in the next section in Algorithm \ref{algo1}.

As mentioned above, the risk field is inversely proportional to the distance, so the generally vehicle risk model can be written as the following formulas:
\begin{equation}
E_{V}=\frac{e^{{\delta}\cdot{S\cdot acc}\cdot {cos(\theta_3)}}}{dis_{virtual}}.
\label{equ 12}
\end{equation}

\begin{equation}
\begin{split}
\text {S}&=\frac{d_v \cdot yaw_v}{\|d_v\|\|yaw_v\|} \\
&=\frac{\sum_{i=1}^{n} d_{vi} \times yaw_{vi}}{\sqrt{\sum_{i=1}^{n}\left(d_{vi}\right)^{2}} \times \sqrt{\sum_{i=1}^{n}\left(yaw_{vi}\right)^{2}}}
\end{split}
\label{equ 13}
\end{equation}

Equation \ref{equ 12} gives the formula of $E_{V}$, where ${\delta}$ is an undetermined coefficient, $S$ is the cosine similarity between relative location and velocity vector, equation \ref{equ 13} shows the normal cosine similarity calculation formula, $d_v$ and $yaw_v$ are the position and velocity angle vectors between obstacle and ego cars, the calculation method is mentioned in algorithm \ref{algo1}. $acc$ represents the sample acceleration at that time, and ${\theta_3}$ is the clock-wise angle from any point in the map around the obstacle car to the mass center of the obstacle car with obstacle car's motion direction.

Using the general traffic situation example in Fig. \ref{fig8}, we assume car0 is the ego-car. $v2$ and $v3$ is faster than $v0$, these three cars are in the same lane and have no intention to change lanes. $v1$ is slower than $v0$, and it has a tendency to change lanes.
\begin{figure}[h]%
\centering
\includegraphics[width=\linewidth]{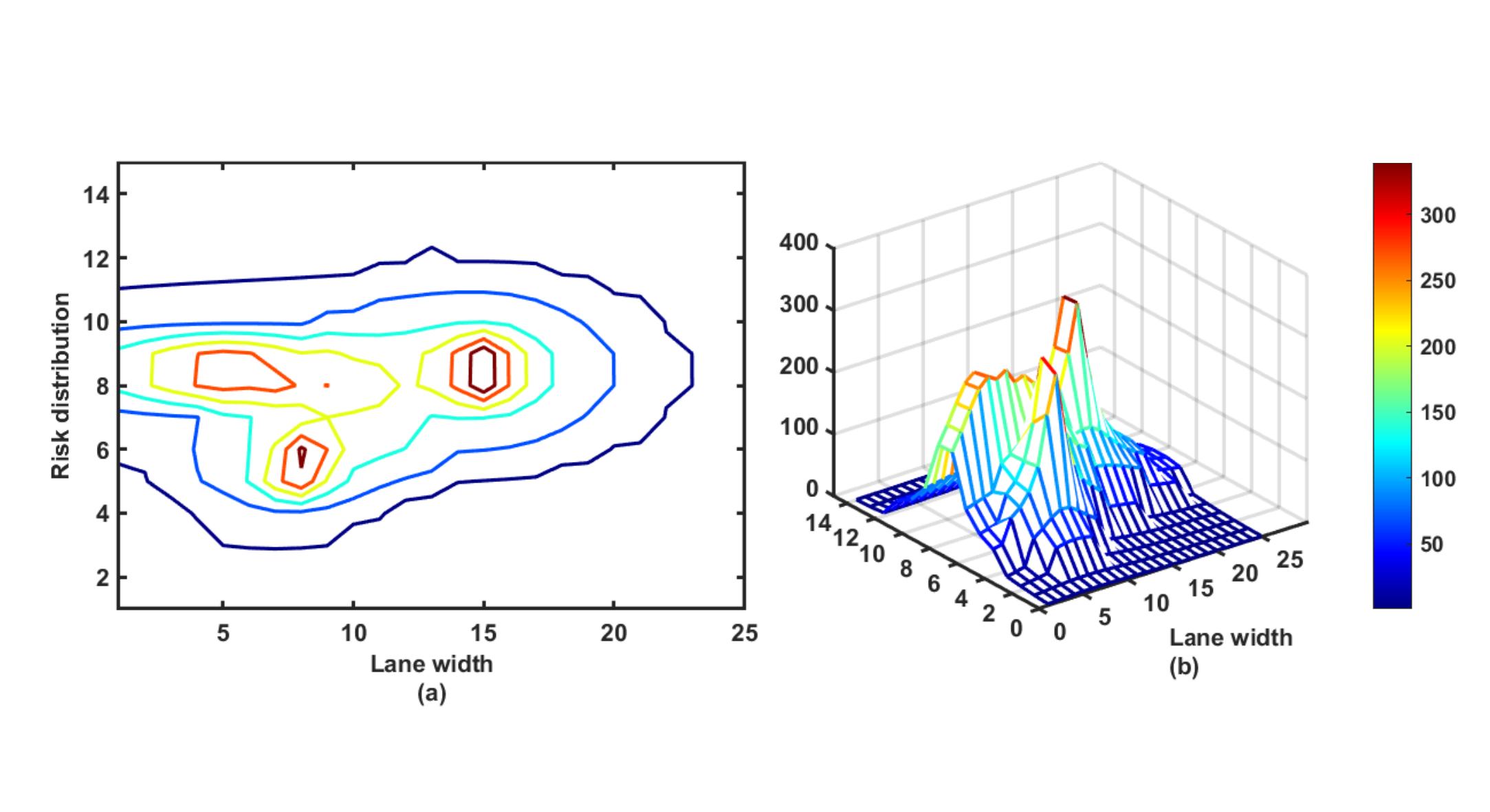}
\caption{The risk distribution of three obstacle cars in Fig. \ref{fig8}. (a) is the contour map of the risk field, (b) is the 3D format.}\label{fig9}
\end{figure}
\begin{figure}[h]%
\centering
\includegraphics[width=\linewidth]{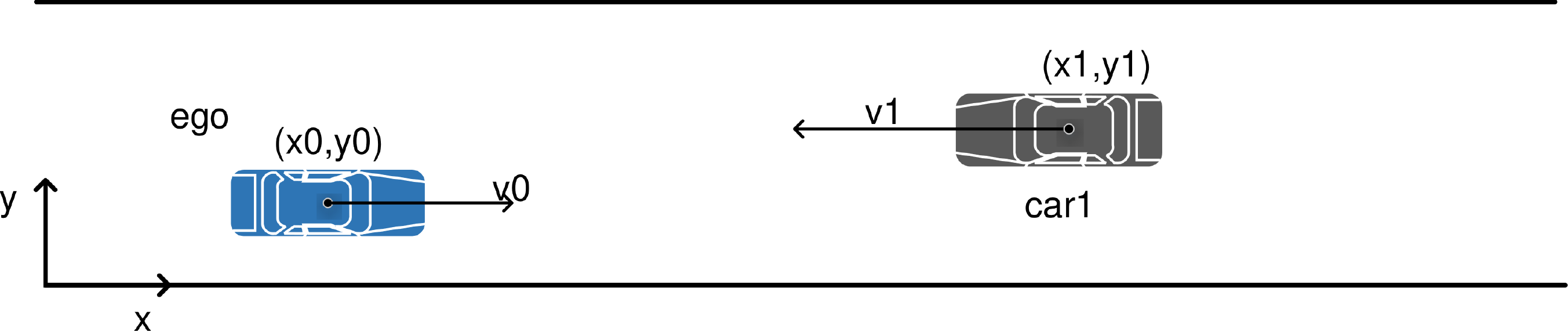}
\caption{Special traffic scenario. Car2 moves to the ego in opposite direction in the same lane.}\label{fig10}
\end{figure}

\begin{figure}[h]%
\centering
\includegraphics[width=\linewidth]{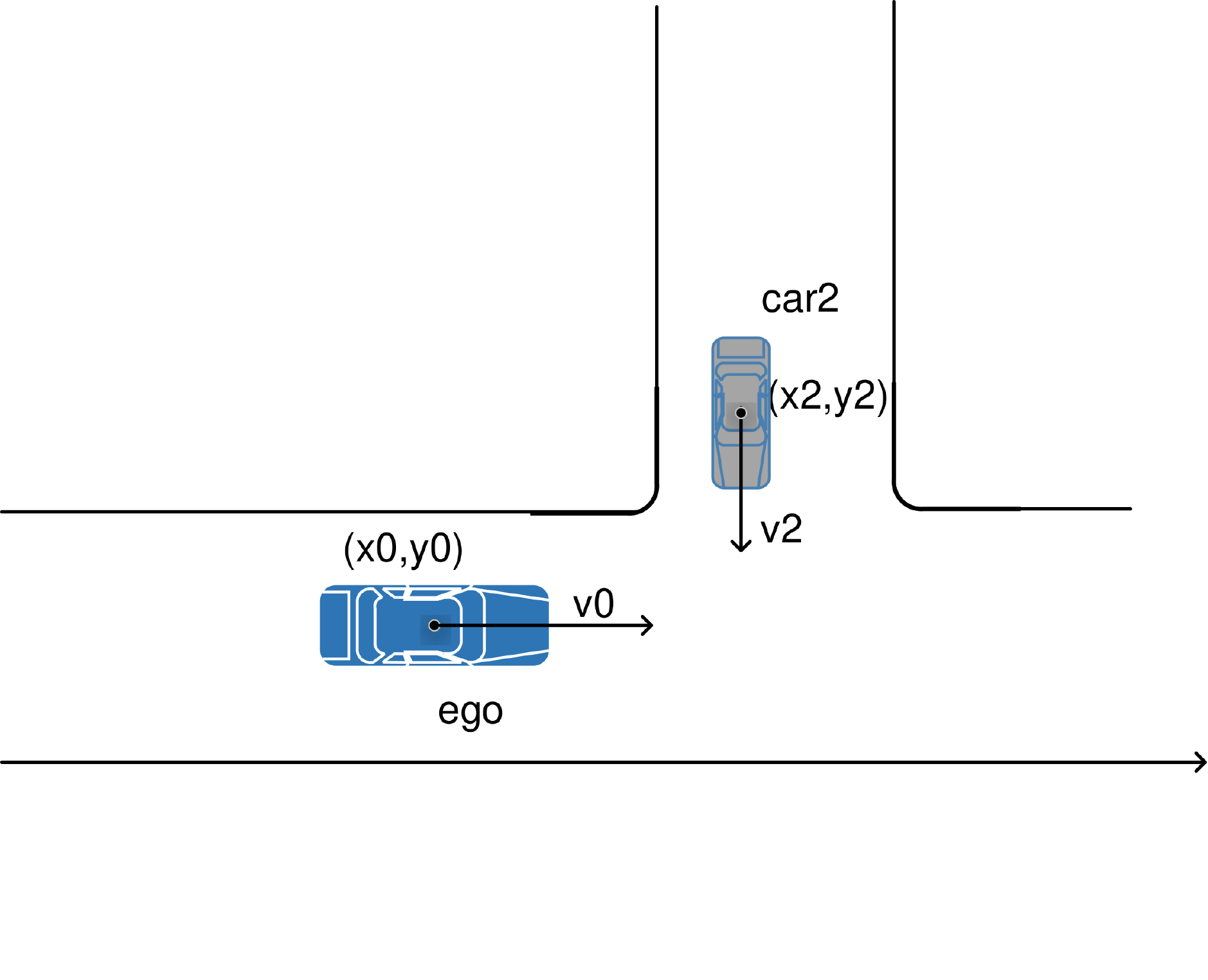}
\caption{Special traffic scenario. Car1 has a $90$ degree to change the line.}\label{fig11}
\end{figure}

Fig. \ref{fig9} shows the risk distribution of three obstacle cars in Fig. \ref{fig8}. 
According to the color bar, if the color of the risk field is closer to red, the risk is higher; otherwise, if the color is close to blue, the risk is lower.
The risk distribution means the potential risk influence range of specific obstacles. $(a)$ in Fig. \ref{fig9} gives the 2-D format results, the highest risk is generated by the obstacles' locations. Then the risk distributes to the surroundings continuously and its value decreases with the distance, forming the risk field distribution. $(b)$ is the 3-D format of the risk result.

In the top lane, the preceding car and following car's velocity are faster than the ego's. 
Although according to common sense, the following car seems to have no effect on ego car when all of them obey the traffic rule and no vehicle will change its motion state, we cannot ignore its risk since all the traffic participants may influence ego car's decision when it tends to change motion state.
Besides, we cannot deny the probability that some vehicles will violate the traffic rule. Therefore, all vehicles' risks are calculated by the car model, and
the risk generated by the followed car is bigger than the preceding car, though they have the same motion state and distance to the ego. 
Also, the risk generated by the preceding car is compressed compared to the followed car, which is controlled by parameter $k$ in equation \ref{equ 11}. 
For car in the second lane, its heading angle is bigger than $0$, which causes the risk field also rotates certain angle, and this is reflected in Fig. \ref{fig9}.

\begin{figure}[h]%
\centering
\includegraphics[width=\linewidth]{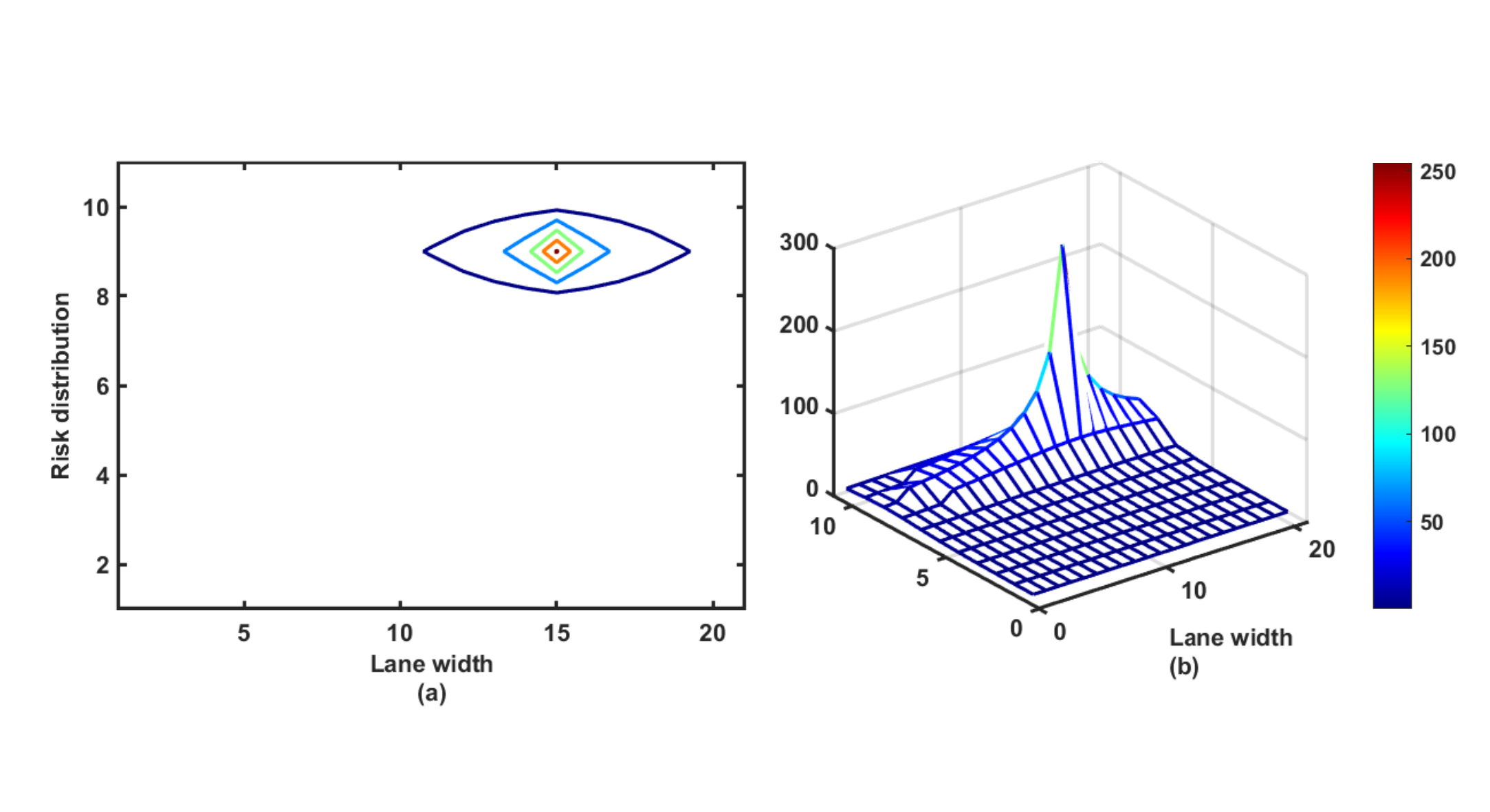}
\caption{Risk field of special traffic scenario in Fig. \ref{fig10}.}\label{fig12}
\end{figure}

\begin{figure}[h]%
\centering
\includegraphics[width=\linewidth]{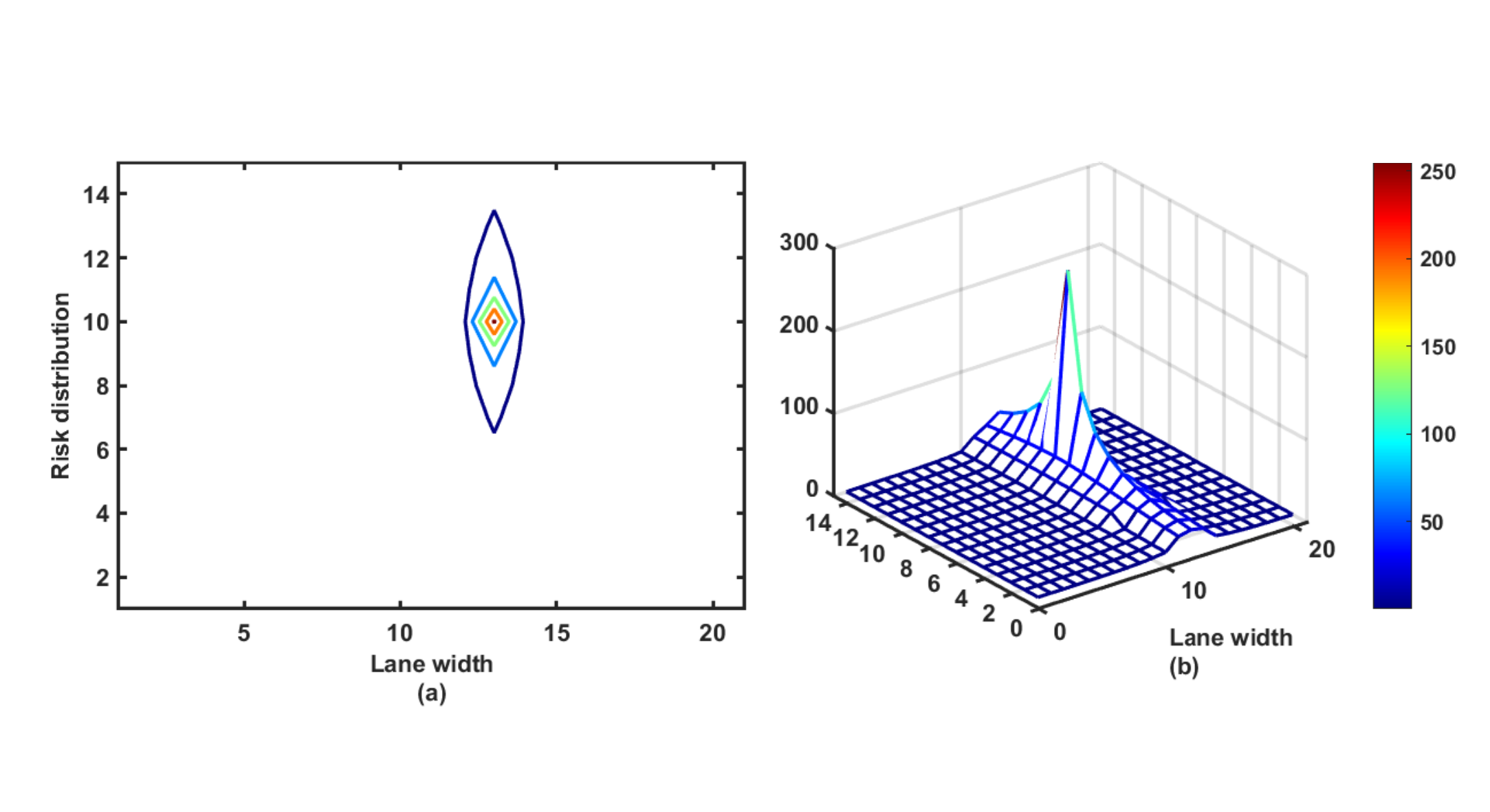}
\caption{Risk field of special traffic scenario in Fig. \ref{fig11}.}\label{fig13}
\end{figure}

Especially, there are some specific scenarios should be considered into the model, such as when obstacle car moving towards the ego in the same lane with opposite motion direction or when the obstacle car has a $90$ degree to the ego in a crossing, shown as car1 and car2 in Fig. \ref{fig10} and \ref{fig11}. 

Fig. \ref{fig12} and \ref{fig13} show the risk field results for car1 and car2. The meaning of the risk distribution in results is same to Fig. \ref{fig9}.
The risk distribution of car1 in Fig. \ref{fig10} enlarges since it approached the ego directly in an opposite direction and has a non-negligible risk to the ego. 
And for car2 in Fig. \ref{fig11}, its risk field is $90$ degree rotated, since its relative moving direction to the ego. 
All of the results are consistent with the driving risk field theory and common sense in the real traffic world.

\subsubsection{Algorithm flow}\label{subsubsec3.2.2}

This section will give a detailed explanation of the calculation method for each parameter in equation \ref{equ 12}, Algorithm \ref{algo1} and \ref{algo2} are the  pseudo-code that show a rough algorithm framework for handling the vehicle risk model. 

For clarification, the $ego$ and $obs$ in the pseudo-code are all objects that contain the attribute of each car (e.g., dimension, location, v, a, yaw), and the $obs_{list}$ is a list that contains multiple $obs$ objects.
%Algorithm 1
\begin{algorithm}
\caption{Calculate $k$ in equation \ref{equ 9}}\label{algo1}
\begin{algorithmic}[1]
\REQUIRE $ego$, $obs_{list}$
\ENSURE $k$
\FOR {$obs$ in $obs_{list}$}          %FOR循环结构 
    \STATE $d_{v} = [ego_{x}-obs_{x}, ego_{y}-obs_{y}];$
    \STATE $re_{v} = obs_{v}-ego_{v};$
    \STATE $re_{yaw} = obs_{yaw}-ego_{yaw};$
    \STATE $a_{sample} = normrnd(obs_{a}, var, num);$
    \STATE $yaw_{v} =[cos(re_{yaw}), sin(re_{yaw})];$
    \STATE $similarity = cosine_{similarity}(d_v, yaw_v);$
    \IF{$in$ $same$ $lane$}
    			\STATE $sign $ $=$ $re_{yaw} == 0  ? true : false;$
                \STATE $re_v$ $=$ $similarity>=0 ? re_{v} : -re_{v};$
                \IF {$similarity > 0$ $and$ $velocity$ $is$ $opposite$}
                    \STATE $Approaching$;
                  \ELSE 
                  \IF {$!sign$ $or$ $re_v < 0$} 
                		\STATE $Leaving$;
                    \ENDIF
                  \ELSE
                  		 \STATE $Approaching$; 
                   \ENDIF
      \ELSE
      				\IF{$not$ $($ $d_v[1]$ $XOR$ $re_{yaw}$ $)$}
                    			\STATE $sign = true$;
                      \ELSE 
                      			\STATE $sign = false$;
                       \ENDIF
                       \IF{$sign$} 
                       			\STATE $Approaching$;
                         \ELSE
                         		\STATE $Leaving$;
                          \ENDIF
      \ENDIF
      \IF{$risk$ $yes!$} 
      			\STATE $k$ $=$ $1+log_2(1+re_v)$;
      \ELSE
      			\STATE $ratio$ $=$ $abs(re_v)/ego_v$;
      			\STATE $k$ $=$ $(1+e^{-ratio})\cdot(obs_w/obs_l)$;
      \ENDIF
      \STATE jump to $Algorithm2$ to calculate risk field;
\ENDFOR
\end{algorithmic}
\end{algorithm}

% Algorithm 2
\begin{algorithm}
\caption{Calculate $E_{V}=\frac{e^{{\delta}\cdot{S\cdot acc}\cdot {cos(\theta_3)}}}{dis_{virtual}}$}\label{algo2}
\begin{algorithmic}[1]
\REQUIRE $ego$, $obs$, $k$, $a_{sample}$, $similarity$
\ENSURE $E_{V}$
\FOR {$i$ $=$ $0 < size(a_{sample})$}
			\STATE $dis[i]$ $=$ $obs_v\cdot t + 0.5\cdot a_{sample}\cdot t^2$;
\ENDFOR
\STATE rotate coordinate according to $obs_{yaw}$;
\FOR {$i$ $=$ $0 < size(a_{sample})$}
			\STATE $dis_{virtual}$ $=$ $dis_{virtual} > dis[i]$ $?$ $e^{dis_{virtual}}$   $:$  $dis_{virtual}$ ;
            \STATE calculate $E_{V}$ using equation \ref{equ 12};
\ENDFOR
\STATE Mean value: $E_{V}$ $=$ $E_{V}/size(a_{sample})$;
\STATE Normalization: $E_{V}$ $=$ $E_{V} / max(E_{V})$;
\end{algorithmic}
\end{algorithm}

Algorithm \ref{algo1} shows the flow of how to select the $k$, which is the key parameter in equation \ref{equ 9} to calculate the virtual distance. The core idea is to  find if there is any possibility for obstacle and ego car to collide. To figure out this, the algorithm first get the position and velocity vector $d_{v}$ and $yaw_{v}$, the cosine similarity $similarity$ is calculated using this two vectors. Then, it divides the traffic situation into two parts. For cars in the same lane,  $sign$ is used to judge if they are in the following scenario and $re_v$ is modified considering their relative position and velocity. For example, $car2$ and $ego$ in Fig. \ref{fig8}, $re_v$ is bigger than $0$, while $similarity$ is smaller than $0$, after going through line 9 in Algorithm \ref{algo1}, $re_v$ is negative, which will make the program jump to line 11. Since this paper assumes obstacle cars should have their own safety system, so in the following scenario, when they try to change lanes, we assume they will avoid collision and have less risk to ego. Thus, in line 10, when $sign$ is false or $re_v < 0$, the risk is smaller than usual situation. For cars in the different lanes, when the position vector and $re_{yaw}$ have the same sign, parameter $sign$ is $true$, which is used to decide if these cars have the tendency to move closer. If $sign$ is $true$, then the risk should be attach importance to. According to the judgement flow, the strategy of calculating the $k$ is based on the attributes of the ego and obstacle cars, and $k$ can be used to revise the risk distribution.

When we get the $k$, Algorithm \ref{algo2} can be used to calculate the final $E_V$. To deal with the uncertainty of perception and make short time obstacle car's motion prediction, this paper uses Gaussian Distribution $X \sim \mathcal{N}\left(\mu, \sigma^{2}\right)$ to generate accelerate, the mean $\mu$ is the obstacle car's accelerate at that time, $\sigma^{2}$ is calculated by specific traffic datasets. In this paper experiment part, we choose NGSIM data to get $\sigma^{2}$, the time interval is $0.1s$ considering the program delay. $a_{sample}$  is used to calculate the future reaching distance using $d=v \cdot t+\frac{1}{2} \cdot a \cdot t^{2}$. Using equation \ref{equ 9}, we can get the $dis_{virtual}$ , line 6 in Algorithm \ref{algo2} is used to control the risk distribution range. In the end, the final $E_V$ can be attained by equation \ref{equ 12}.

The time complexity of the whole risk framework algorithm is limited to $O(n^2)$, and the real time consume depends on the perception range and the number of nearby obstacles. The detailed analysis is discussed in appendix \ref{secA1}.

\subsection{Pedestrian risk potential field model}\label{subsec3.3}
Pedestrian is a non-negligible part in urban traffic system. However, due to the difficulty to get precise pedestrian's attributes and its unpredictable motion state, pedestrian has not be considered as a part in driving risk field model before. Although the reasons mentioned above, pedestrian-car collision accounts for  a large percentage of traffic accidents nowadays \cite{Quintero}, and analyze the interaction between vehicles and pedestrians is essential. Therefore, this paper tries to integrate the pedestrian risk model into traditional driving risk field model. Shen \cite{ShenRaksincharoensak-553} proposed a new pedestrian risk assessment method that they proved the risk generate by pedestrian is not only related to the traditional metric (TTC), but also related to pedestrian intention and the intensity of near-accident event. Inspired by their work, this paper proposes a pedestrian risk model also considers those factors. 

Pedestrians have their own unique characteristics, for example, they can Vchange their motion state instantly, and they are always much smaller that vehicles. Considering these features, unlike vehicles, which need to consider the shape and kinetic motion model, pedestrians are tend to be modeled as points, which can move flexible and the shape can be ignored. 
In Shen's paper, the pedestrian risk is related to TTC, the intensity of near-accident event and pedestrian intention. 
Since the pedestrian intention prediction is based on Machine Learning, and need enough reliable data to let model converge, which is difficult to obtain, so this paper remove this factor in the model. 
According to the definition of $TTC=\frac{d}{v}$, when TTC is small, the risk is relative high. So the risk is in inverse proportion to TTC.
However, TTC is too simple to decide the pedestrian risk totally, other factors need to be considered into the model. Near-accident event is related to the probability of traffic event appearance, which may cause damage.
When the probability is high, the potential risk is also high.
Thus, the two factors are combined for a more comprehensive model to decide the pedestrian risk, and the formula can be expressed:
\begin{equation}
E_{P}=\eta_{1} e^{\lambda}+\eta_{2} e^{-TTC},
\label{equ 14}
\end{equation}
where $\eta_{1} > 0$ and $\eta_{2} > 0$ are the weights of two parts and $\eta_{1} + \eta_{2} = 0$, ${\lambda}$ is the event intensity at that position, which may influence the motion tendency of the pedestrian. And  TTC depends on the ego-car's motion state at the current time.

To get ${\lambda}$, this paper generated $v$ and $distance$ data based on the method mentioned in paper \cite{ShenRaksincharoensak-553}, and uses the data to calculates the intensity ${\lambda}$. Referring to the lab results in Shen's paper, Pearson model has better fitting performance than bivariate Gaussian model, so this paper chooses Pearson model, then the ${\lambda}(v,s)$ \cite{ShenRaksincharoensak-553} is expressed as:
\begin{equation}
\lambda\left(v, s\right)= \gamma\left(\beta+\frac{\left(v-\mu_{v}\right)^{2}}{\delta_{v}^{2}}+\frac{\left(s-\mu_{s}\right)^{2}}{\delta_{s}^{2}}-\right.\\
\left.\times \frac{2 \rho\left(v-\mu_{v}\right)\left(s-\mu_{s}\right)}{\delta_{v} \delta_{s}}\right)^{-b}.
\label{equ 15}
\end{equation}

The parameter vector needs to obtained using Pearson system model-based intensity \cite{ShenRaksincharoensak-553} is denoted as:

\begin{equation}
\theta_{p o}=\left[\mu_{v_{0}}, \mu_{s}, \delta_{v_{0}}, \delta_{s}, \rho, \gamma, \beta, b\right]^{Tlign}.
\label{equ 16}
\end{equation}
\begin{figure}[h]%
\centering
\includegraphics[width=\linewidth]{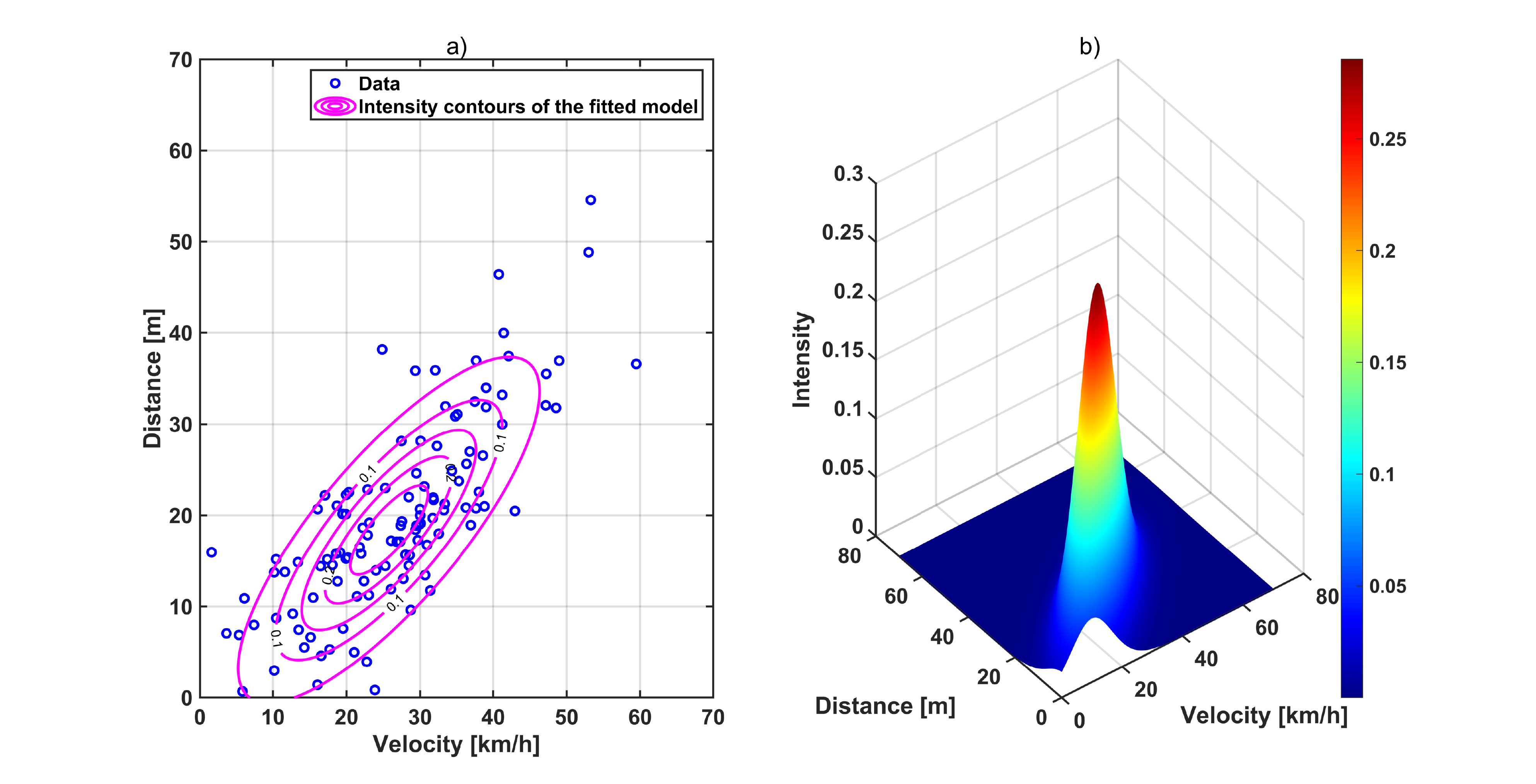}
\caption{(a) is the contour map result of intensity calculated based on Pearson model; (b) is the 3D format intensity of the result.}\label{fig14}
\end{figure}

The meaning of each parameter is same as \cite{ShenRaksincharoensak-553}, this paper will not discuss much about this. When the intensity distribution is obtained, the risk of person can be calculated by equation \ref{equ 14}, since the $TTC$ is definite when the state of ego-car and pedestrian are decided. The result of intensity calculated based on Pearson model is shown in Fig. \ref{fig14}.

\subsection{CCDF based risk metric}\label{subsec3.4}
CCDF curve $F(X)$ is the integration of  a real random variable $X$'s probability density function (pdf), it describes the summation of the probability that $X$ is higher than specific value $a$.  And is has been used for risk in many areas \cite{WangChen-511,CouncilStudies-566}, since it can reflect the probability that a variable exceeds a certain value, which is referred to in risk assessment as "the exceedance question." Thus, this paper introduces this factor as the driving risk metric. The following shows the formulas to calculate the CCDF curve:
\begin{equation}
F(X)=\int_{-\infty}^{a} P(X \leq a),
\label{equ 17}
\end{equation}

\begin{equation}
CDF=P(X \leq a)
\label{equ 18}
\end{equation}

Probability Density Function ($pdf$) is the probability density of independent variable $X$, and Cumulative Distribution Function (CDF) is the derivative of $pdf$, which means the summation of the probability when variable $X$ is less than or equal to a constant $a$.
CCDF represents the sum of the probability when variable $X$ is greater than a constant $a$ and is equal to $1-CDF$, shown in equation \ref{equ 19}.
Also, since the value of CCDF is related to the summation of an independent event's probability, according to the probability theory, the value range of it is always between $[0,1]$.
\begin{equation}
CCDF=F(a)=P(X>a)=1-CDF.
\label{equ 19}
\end{equation}

\begin{figure}[h]%
\centering
\includegraphics[width=\linewidth]{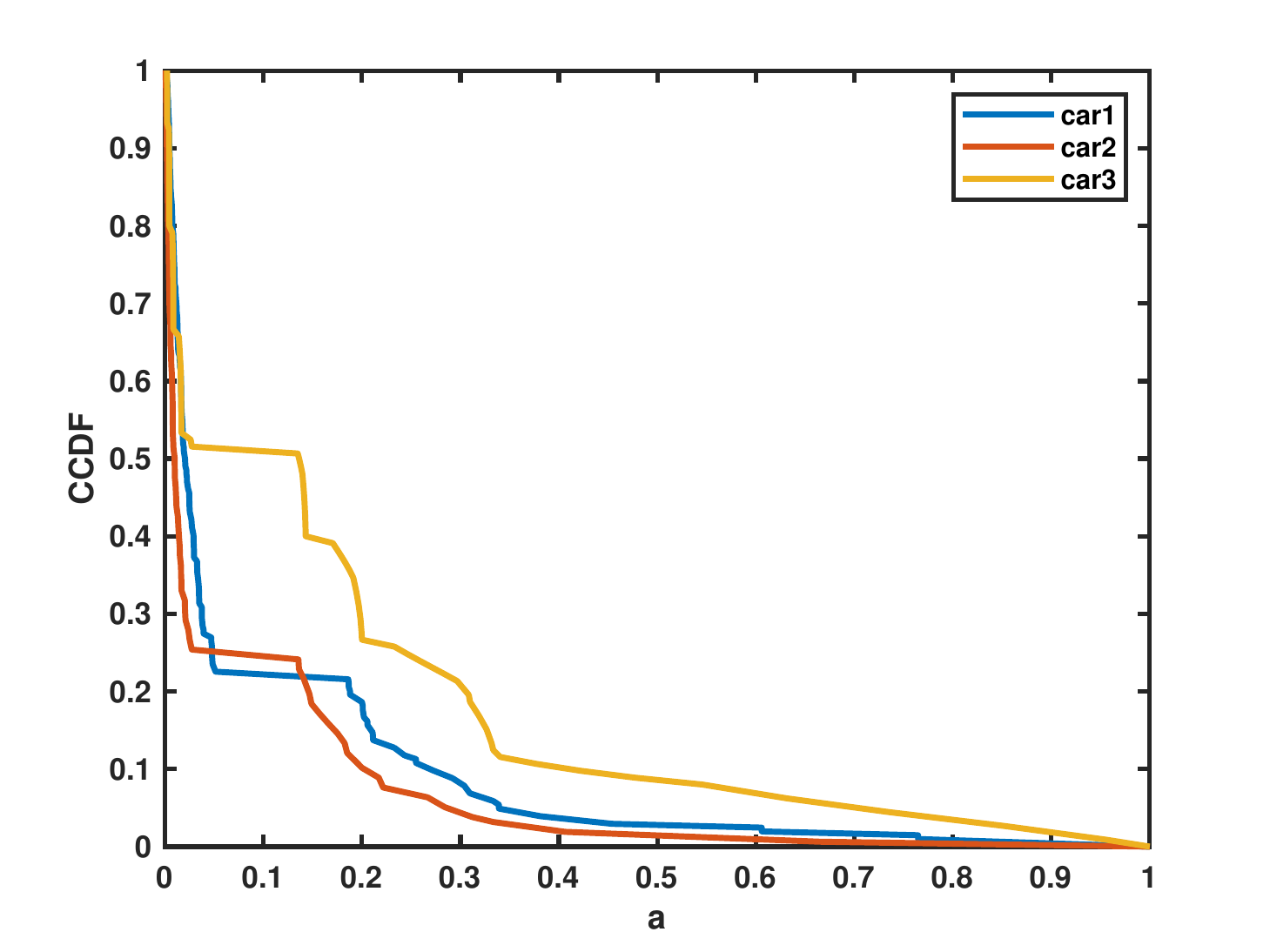}
\caption{CCDF curve for each car in Fig. \ref{fig8}.}\label{fig15}
\end{figure}

In this paper, when the risk strength of a specific obstacle in the traffic map has been calculated, the risk value distribution probability in the grid map can be obtained. Then the CCDF curve of this object can be calculated by its definition, equation \ref{equ 19}.
Using the traffic example in section \ref{subsubsec3.2.1} in Fig. \ref{fig9}, the CCDF curve of each obstacle car is shown in Fig. \ref{fig15}.
The horizontal axis is the constant value $a$, and the vertical axis is the value of CCDF, which means the sum of the probability when the risk value is greater than $a$.
In other words, for all the CCDF curves in one frame, the higher curve always means that the object's risk has more influence on ego than others.

As discussed in section \ref{subsubsec3.2.1},  car$1$ is in the different lane, car$2$ and car$3$ are in the same lane as ego-car, and they are in the following scenario with ego. Inspecting Fig. \ref{fig15}, the curve of car$3$ is higher than other two cars all the time, which means the probability of its risk strength exceeds $a$ is highest, for all the certain value $a$. In other words, car$3$ has the highest risk to ego-car, which is consistent with the result in section \ref{subsubsec3.2.1}. For curve$1$ and curve$2$, they have much less risk to ego than car$3$, however, the picture also shows that car$1$ has a little higher risk than car$2$. Considering it tends to change lane and moves closer to ego, while car$2$'s velocity is faster than ego, the CCDF curve gives a correct hint of the risk strength relationship for obstacles. 

The goal of the risk field is to guide AV's decision, which helps it avoid the obstacles and move along low-risk areas, taking the risk example in Fig. \ref{fig9}. The CCDF curve results can give two aspects of references for AV's decision system. The first aspect is that the unit of the area between specific curve and the axes reflects the risk distribution range. The larger the area is, the further risk influence may the object have on ego.
The second aspect is that for specific $a$, the higher the CCDF value is, the higher the risk level of the object may have.
With more real traffic data for model training, more particular details of guidance can be decided.

\section{Experiment results and analysis}\label{sec4}
In this section, we will verify the driving risk field model proposed in this paper can describe the risk distribution law generated by traffic participants in real traffic situation using open source real world traffic dataset NGSIM\footnote{Dataset is available in https://catalog.data.gov/dataset/next-generation-simulation-ngsim-vehicle-trajectories-and-supporting-data} and traffic data collected by real-world AV platform.
With NGSIM dataset, we select three representative traffic scenarios in this experiment, including common road traffic environment, car-following scenario and car cut-in scenario, CCDF-based risk metric results of obstacles are also provided for each scenario. All of the traffic scenarios risk field results are compared to risk field generated by Li's model \cite{LiGan-423} with the total same data, and the results discussion is provided. 
With real-world data, car-following and cut-in scenarios are evaluated and CCDF curves results generated by Li's and RCP-RF model are provided. The areas changes between CCDF curves based on RCP-RF model and coordinates are also given to prove the effectiveness of proposed risk metric.
The effectiveness of proposed pedestrian risk metric is  proved in a separate section in the end. All related program code can be found in this Github repository (https://github.com/SH-Tan/AV\_risk\_model).

\subsection{Vehicle Model Simulation experiments}\label{subsec4.1}

\subsubsection{Dataset description and preprocess}\label{subsubsec4.1.1}
Under the CAV environment,  autonomous car can get the accurate motion information of its nearby vehicles' in real time. Then the autonomous car can use the numerical valuable information to analyze the risk around the environment, effective driving decisions can be made with the analysis. To simulate the CAV environment, this paper uses the open source dataset NGSIM. NGSIM is collected via Next Generation Simulation project, it contains vehicle data from 4 different areas, and it has been adopted by many researchers for micro-driving behavior studies. We select the trajectory data on the southbound direction of US-101 freeway in California, which contains 5 lanes and has a total length of 640 meters, to verify the effectiveness of proposed vehicle model.

The original data contains the vehicle trajectory collected from all four areas, and includes some information we do not need. To facilitate the data can be suitable for this experiment, the data needs to be preprocessed. According to the model necessary input data, we first select 18 among total 25 of the information types, including vehicle id, velocity, location, accelerate, etc. Then, the time interval is converted from millisecond to 0.1 second, and the length unit is converted from feet to meter, which are more suitable for commonly using in China. Further more, 'data split' strategy has been applied for the data. Among the selected location and lane range, we split the data into small piece with sample time, which is set to 10 second. The final data we use focuses on the necessary vehicle information which is necessary the risk model, and small piece data is more easily for feature extraction. By the way, the information of velocity angle is not given in the original data, so the angle of vehicles at each frame is calculated by using the position information.

\subsubsection{Analysis of vehicle risk model for common traffic scenario}\label{subsubsec4.1.2}

\begin{figure}[h]%
\centering
\includegraphics[width=\linewidth]{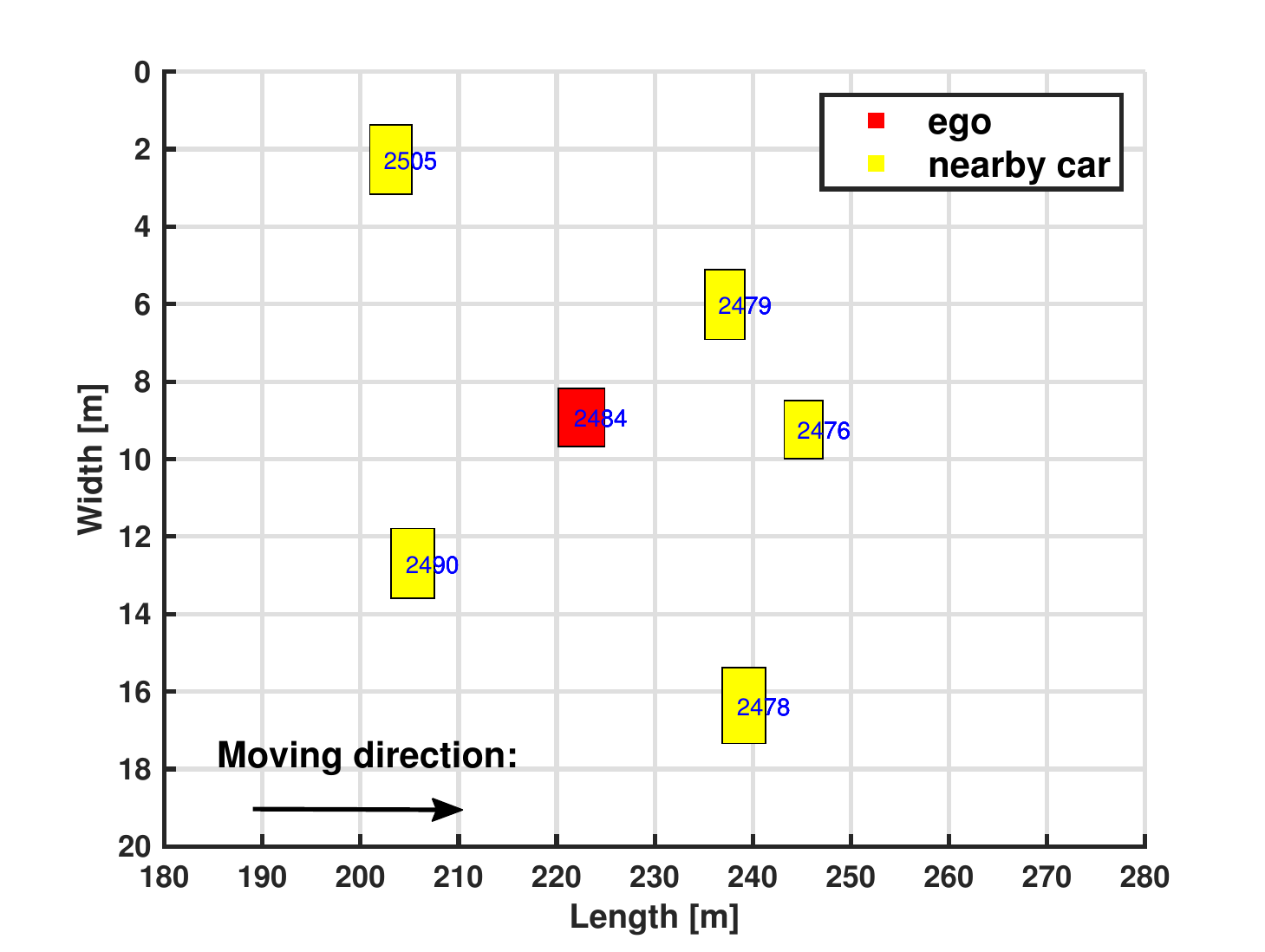}
\caption{Common urban road traffic scenario. The red car is the ego and the nearby yellow cars are obstacles.}\label{fig16}
\end{figure}

Section \ref{subsec3.2} gives the result example with made-up data and describes the algorithm flow of the model. In this section, the consecutive traffic data in NGSIM is used to verify the model is effective for real traffic environment. And the model proposed in \cite{LiGan-423} is also reproduced with the same data, and the difference of the results produced by two models are discussed.

To verify the proposed RCP-RF model is effective to describe the real world traffic environment, a common urban road traffic scenario is shown in Fig. \ref{fig16}.
We extract one static frame from a consecutive vehicle movement scene, all the vehicles keep driving straight during the scene.
And the model is first tested in this scenario. 
The vehicle data is extracted in NGSIM, and table \ref{tab1} gives the detailed motion state information of all the six cars in the pictures. 
The red car is considered as the ego-car, the nearby yellow cars are the obstacle cars.

\begin{table}[h]
\begin{center}
\begin{minipage}{174pt}
\caption{Vehicle motion state in Fig. \ref{fig16}}\label{tab1}%
\setlength{\tabcolsep}{4.2mm}{  %7
\begin{tabular}{@{}lllll@{}}
\toprule
id & Y $[m]$  & X $[m]$ & $car_{l}$ $[m]$ & $car_{w}$ $[m]$ \\
\midrule
2505    & 205.21  & 2.26  & 4.27 & 1.80   \\
2476    & 247.15  & 9.24  & 3.96 & 1.49  \\
2478    & 241.31  & 16.36  & 4.42 & 1.95  \\
2479    & 239.21  & 6.01 & 4.11 & 1.80 \\
2484   & 224.90  & 8.92 & 4.72 & 1.49 \\
2490   & 207.54  & 12.69 & 4.42 & 1.80\\

\end{tabular}}
\setlength{\tabcolsep}{3.6mm}{  %7
\begin{tabular}{@{}lllll@{}}
\toprule
id & v $[km/h]$ & acc $[m/s^2]$ & lane id & yaw $[\circ]$ \\
\midrule
2505    & 32.73   & 0.03048  & 1 &  -0.91 \\
2476    & 32.92   & 0.0091  & 3 & -0.36  \\
2478    & 39.83  & 0.2743  & 5 & -0.34 \\
2479    & 27.43  & 0  & 2 &  -0.36\\
2484    & 27.43  & 0 & 3 & -0.34\\
2490    & 32.69  & 1.0150 & 4 & -0.36\\
\bottomrule
\end{tabular}}
\end{minipage}
\end{center}
\end{table}
\begin{figure}[h]%
\centering
\includegraphics[width=\linewidth]{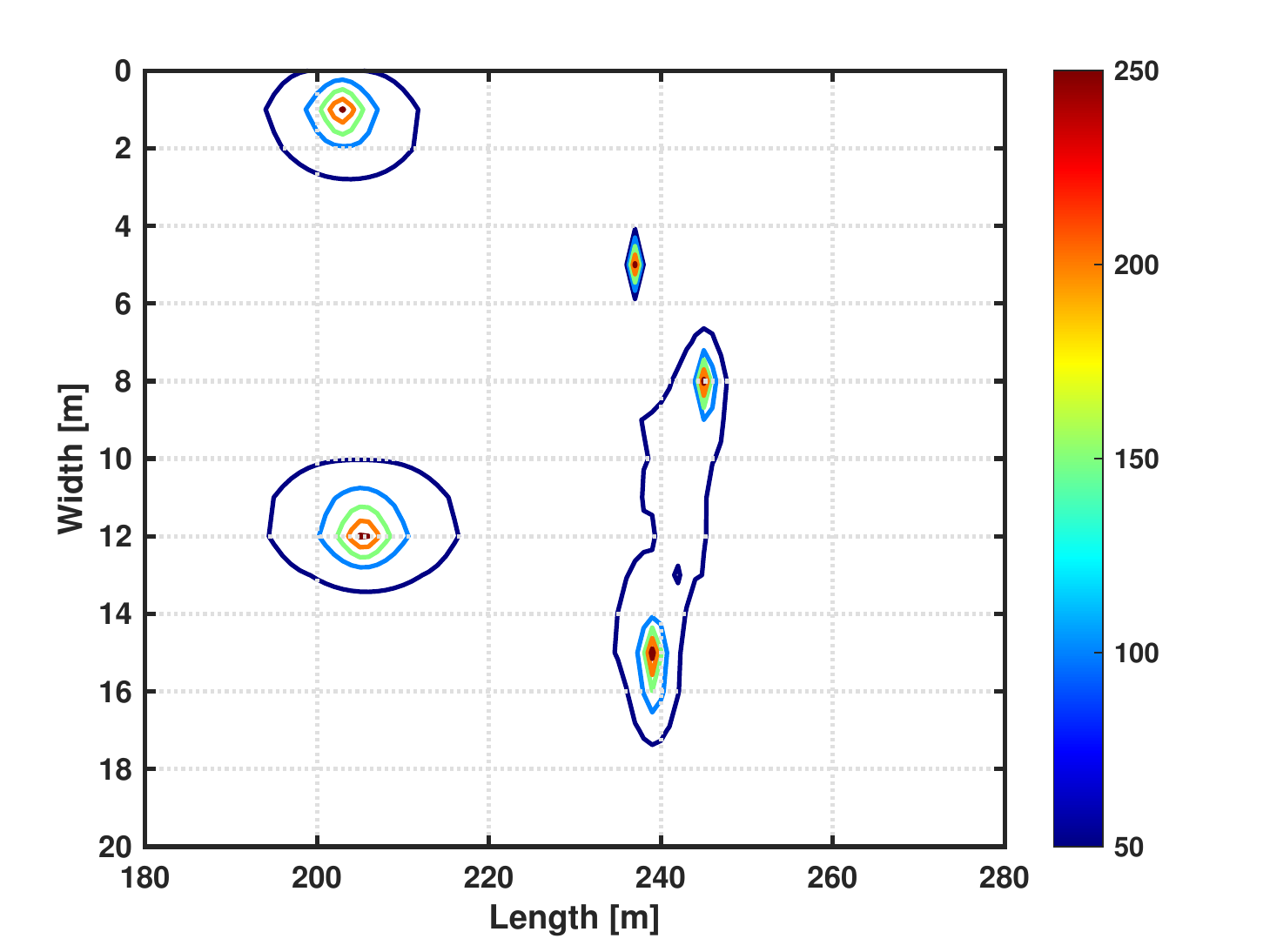}
\caption{Risk field for obstacles cars in Fig. \ref{fig16} using RCP-RF.}\label{fig17}
\end{figure}
\begin{figure}[h]%
\centering
\includegraphics[width=\linewidth]{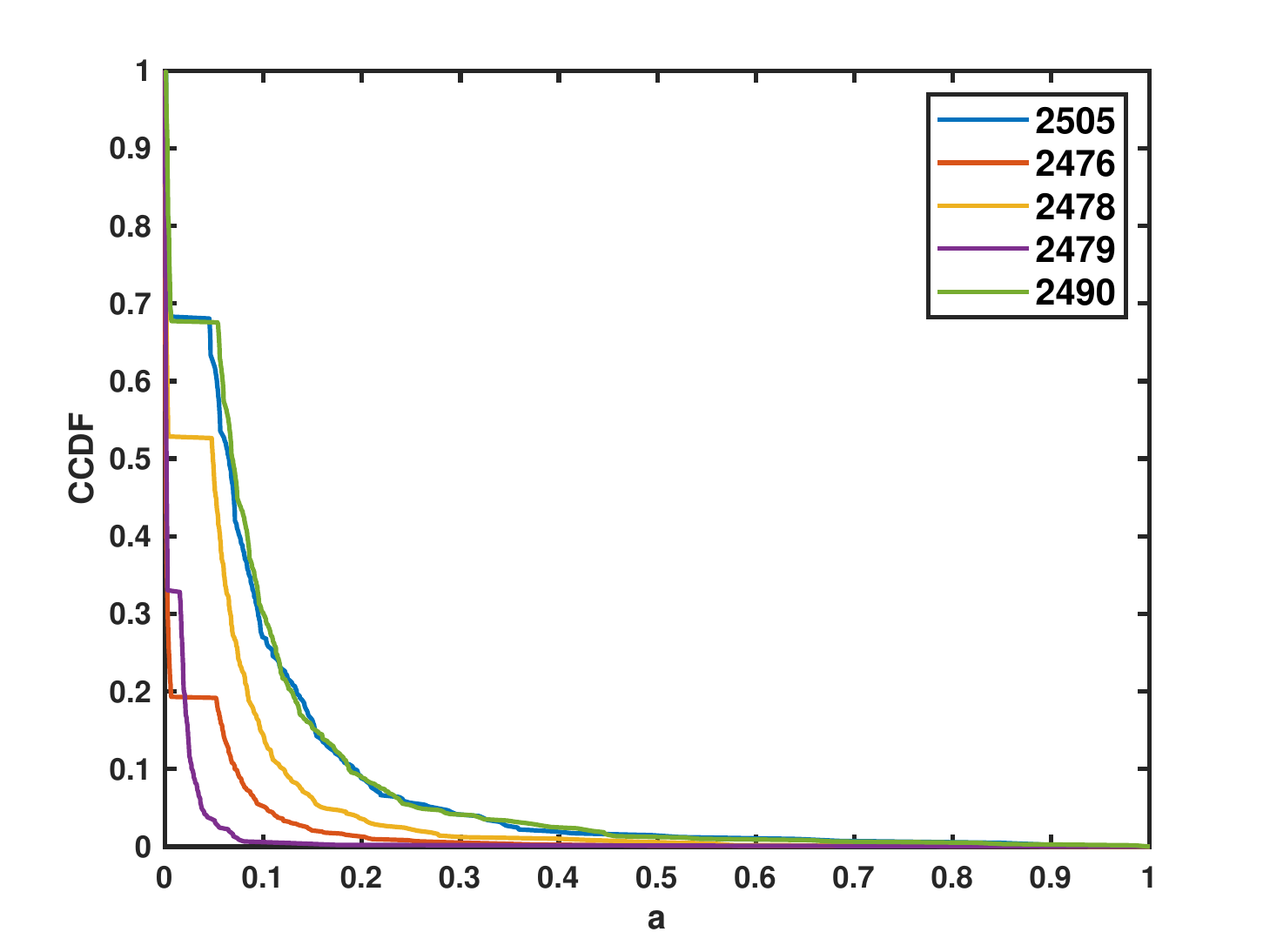}
\caption{CCDF-based risk metric for obstacles cars in Fig. \ref{fig16} using RCP-RF.}\label{fig18}
\end{figure}
Fig. \ref{fig17} and Fig. \ref{fig18} are the risk field and CCDF-based risk metric results of each obstacle cars in the Fig. \ref{fig16} using RCP-RF model proposed in this paper. In Fig. \ref{fig17}, is it obvious that our model can reflect the different risk distribution of each obstacle cars, considering their different motion states. And Fig. \ref{fig18} shows the CCDF risk metric for each obstacle, the relationship of value of each curve is consistent to the risk field of related obstacle cars, for example, the car2479 has the smallest risk field range, so its CCDF curve decreases fastest. It also shows a clear law that when the $a$ is decided, the higher the CCDF probability, the higher the risk the vehicle may generated to ego-car, that is why the curve for car2490 and 2505 have the highest value all the time, and this conclusion is consistent to the results in section \ref{subsec3.4}. To further verify the performance of our RCP-RF model, we select Li's model as the baseline and their risk field results are provided in Fig. \ref{fig19} and Fig. \ref{fig20}.
\begin{figure}[h]%
\centering
\includegraphics[width=\linewidth]{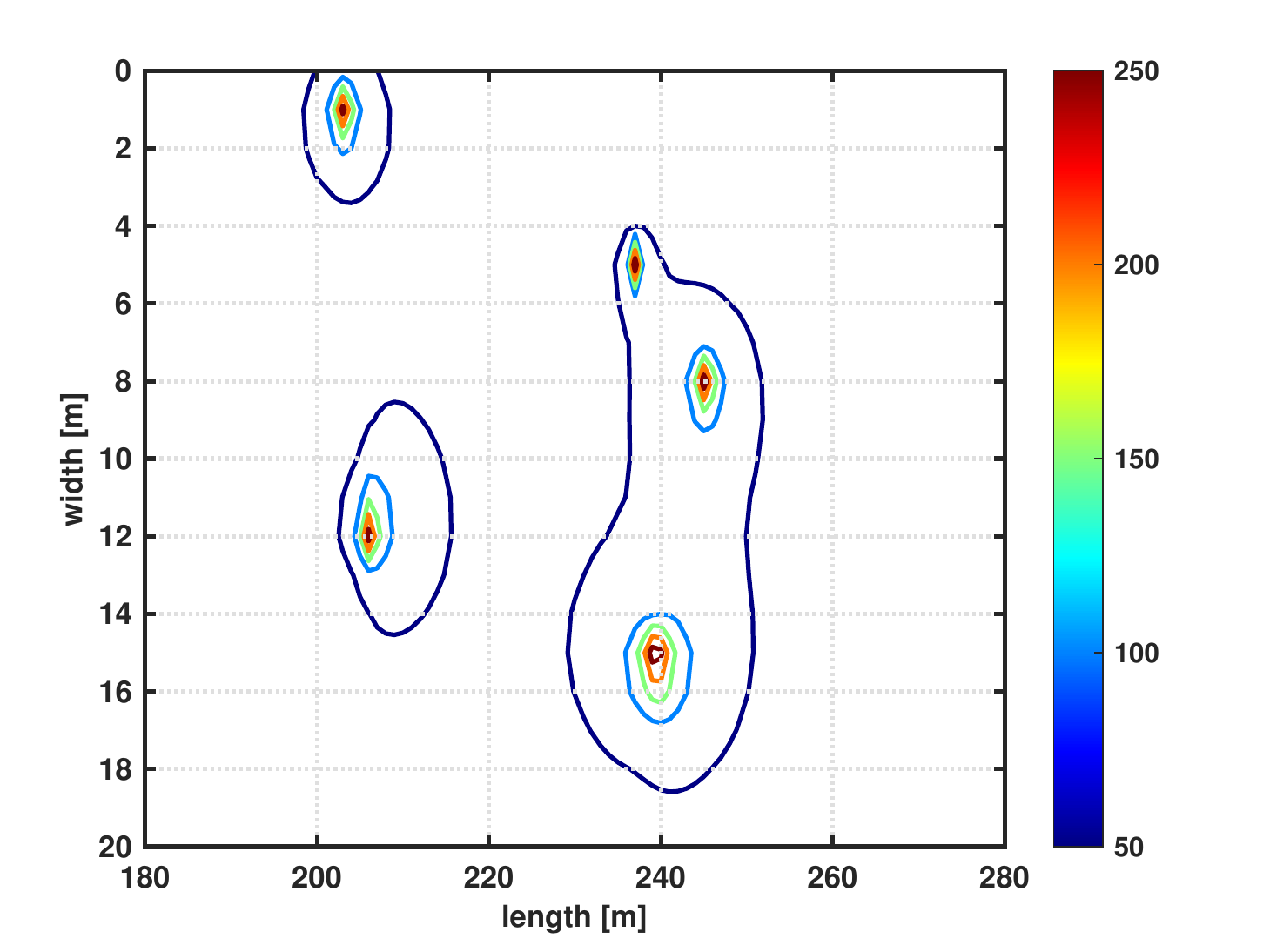}
\caption{Risk field for obstacles cars in Fig. \ref{fig16} using Li's model.}\label{fig19}
\end{figure}
\begin{figure}[h]%
\centering
\includegraphics[width=\linewidth]{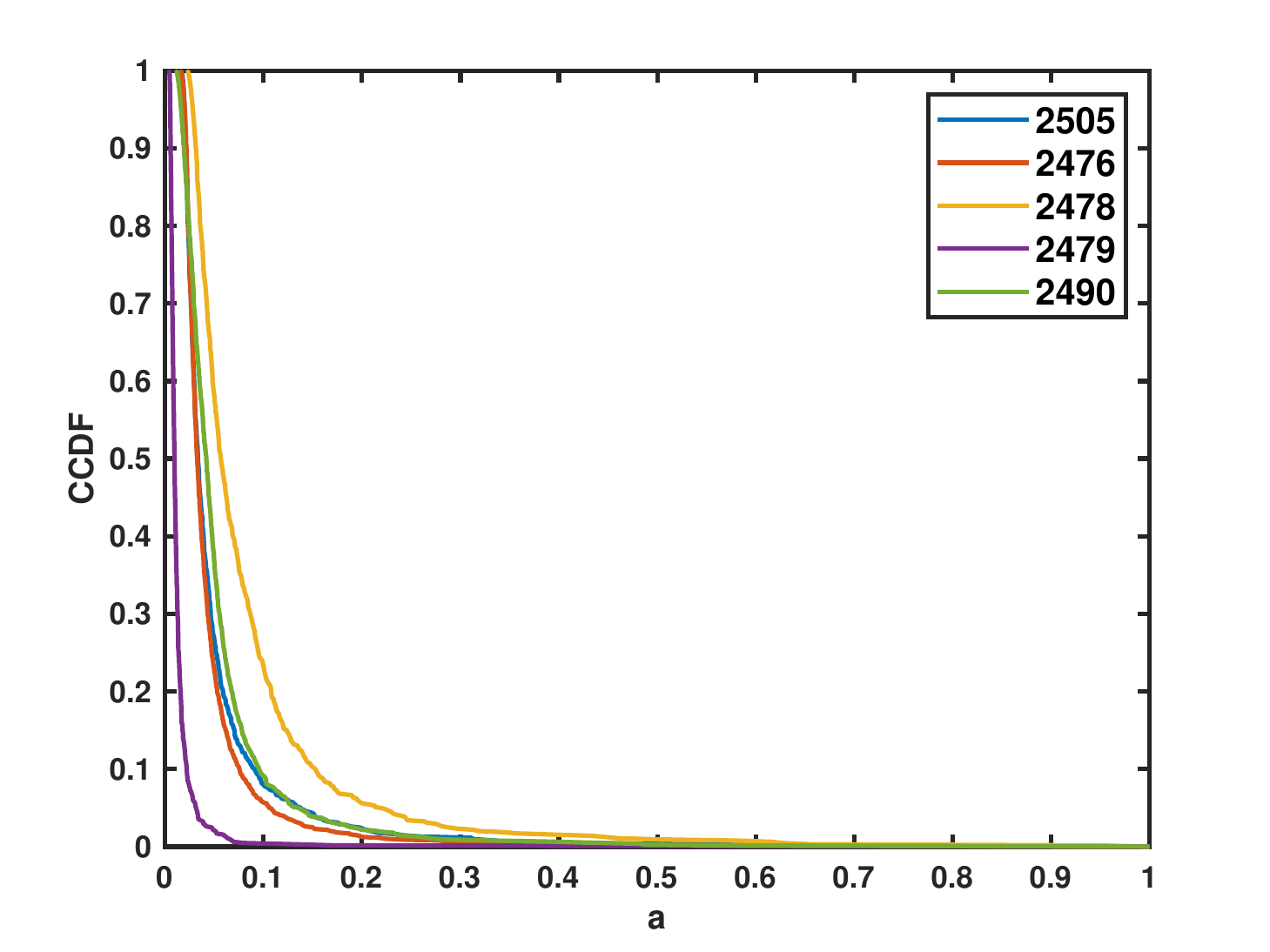}
\caption{CCDF-based risk metric for obstacles cars in Fig. \ref{fig16} using LI's model.}\label{fig20}
\end{figure}

Li's model also can give the rough risk distribution of each obstacle cars, but the difference of two results is obvious, our RCP-RF model can provide a more refined risk field distribution result of specific vehicle than Li's, which is helpful for autonomous car avoid making too conservative driving decisions. Fig. \ref{fig20} is the related CCDF risk curve generated using same vehicle data with Li's model, the algorithm to calculate the CCDF curve is totally same for two models. Compare to Fig. \ref{fig18}, the CCDF risk curve in Fig. \ref{fig20} has little discrimination, and it is hard to estimate  the risk difference of nearby obstacle cars.

\subsubsection{Analysis of vehicle risk model for car following scenario}\label{subsubsec4.1.3}

\begin{figure}[H]%
\centering
\includegraphics[width=\linewidth]{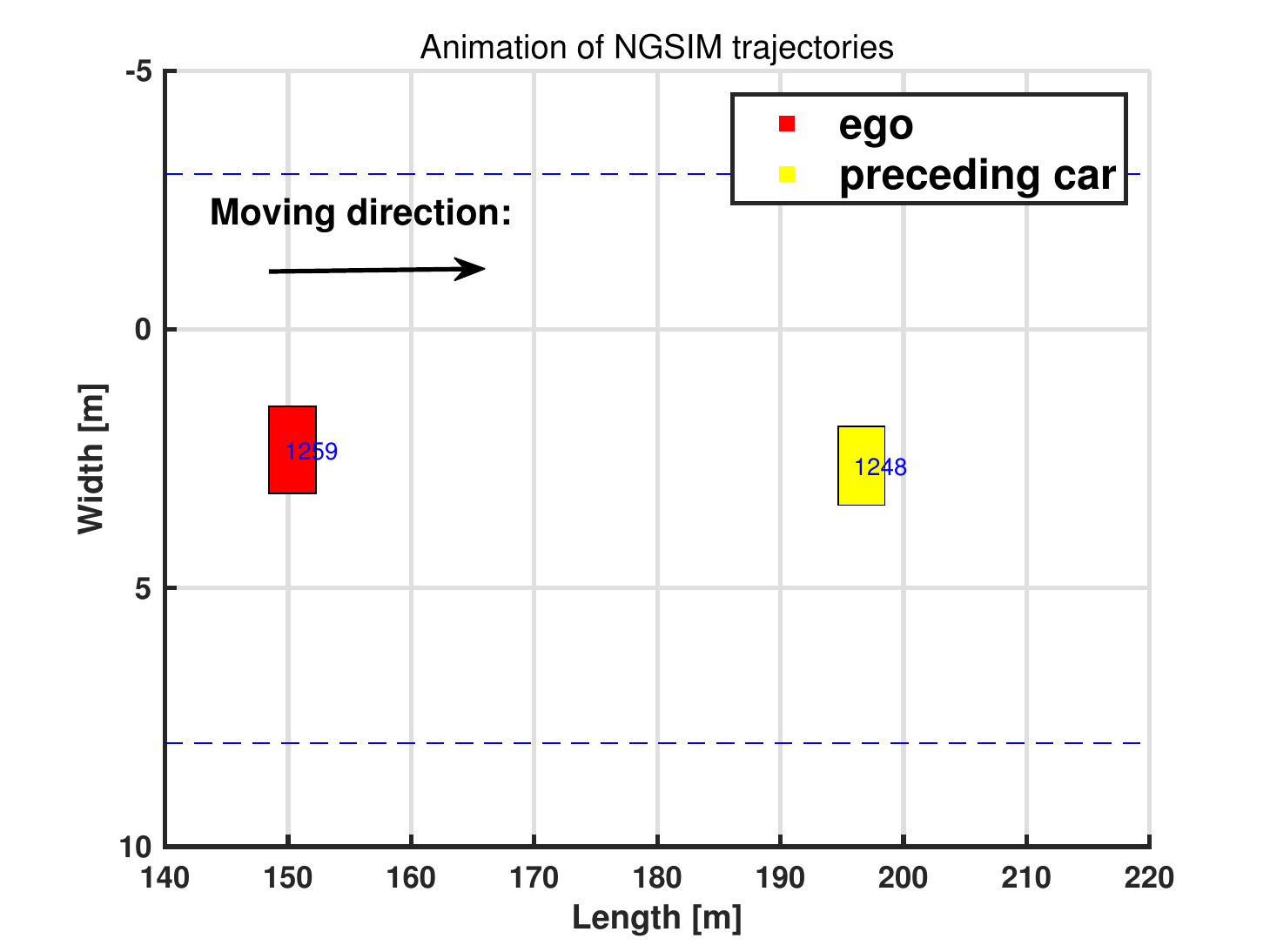}
\caption{Traffic following scenario, the red car1259 is the ego car, yellow car1248 is the preceding car.}\label{fig21}
\end{figure}
\begin{figure}[h]%
\centering
\includegraphics[width=\linewidth]{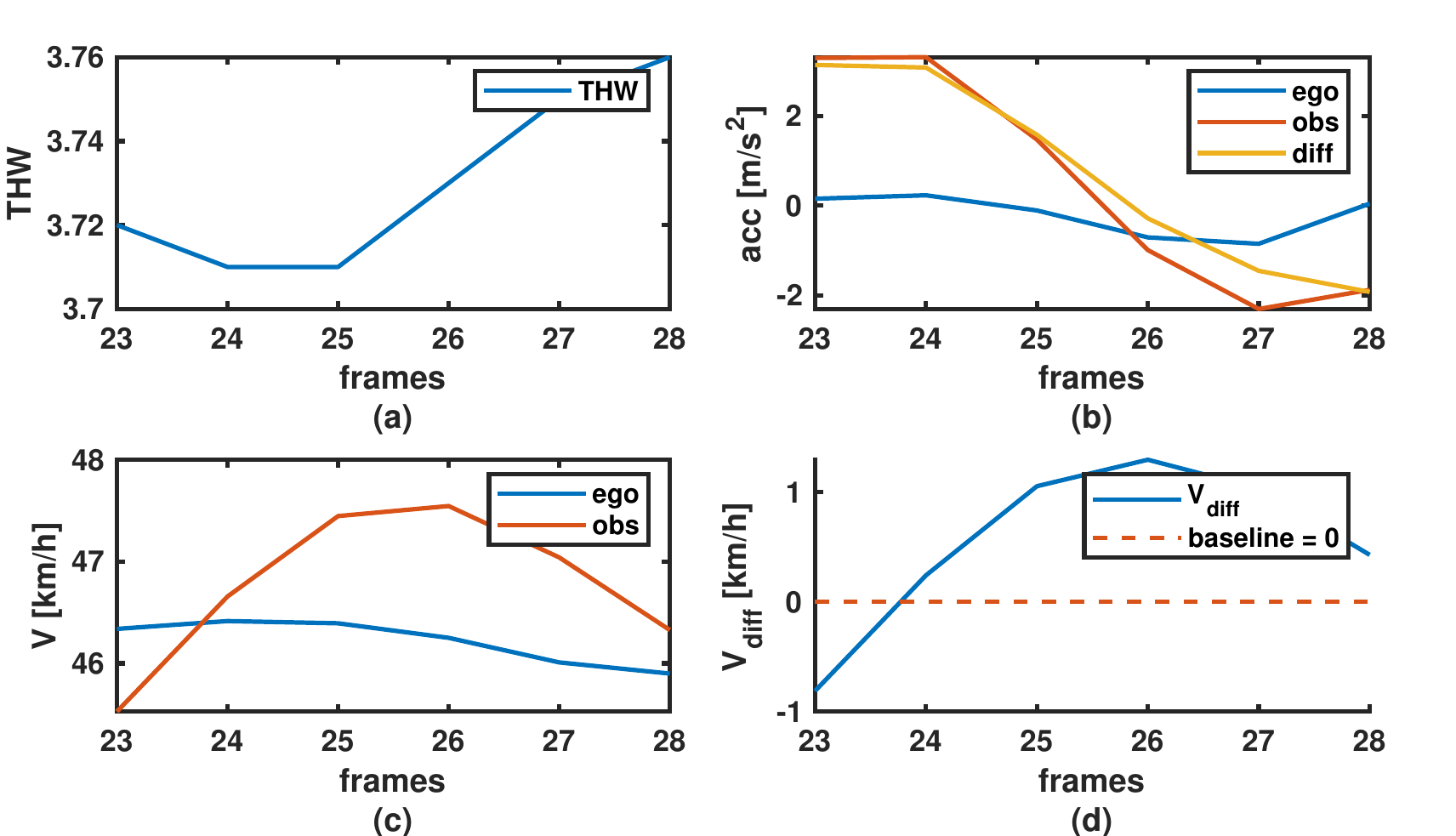}
\caption{Related motion state relationship of ego1259 and 1248. (a) is the THW from 1259 to 1248; (b) is the accelerate relationship between two cars; (c) is the velocity relationship; (d) is the relative velocity change curve.}\label{fig22}
\end{figure}
\begin{figure*}[h]
\centering
\includegraphics[width=\linewidth]{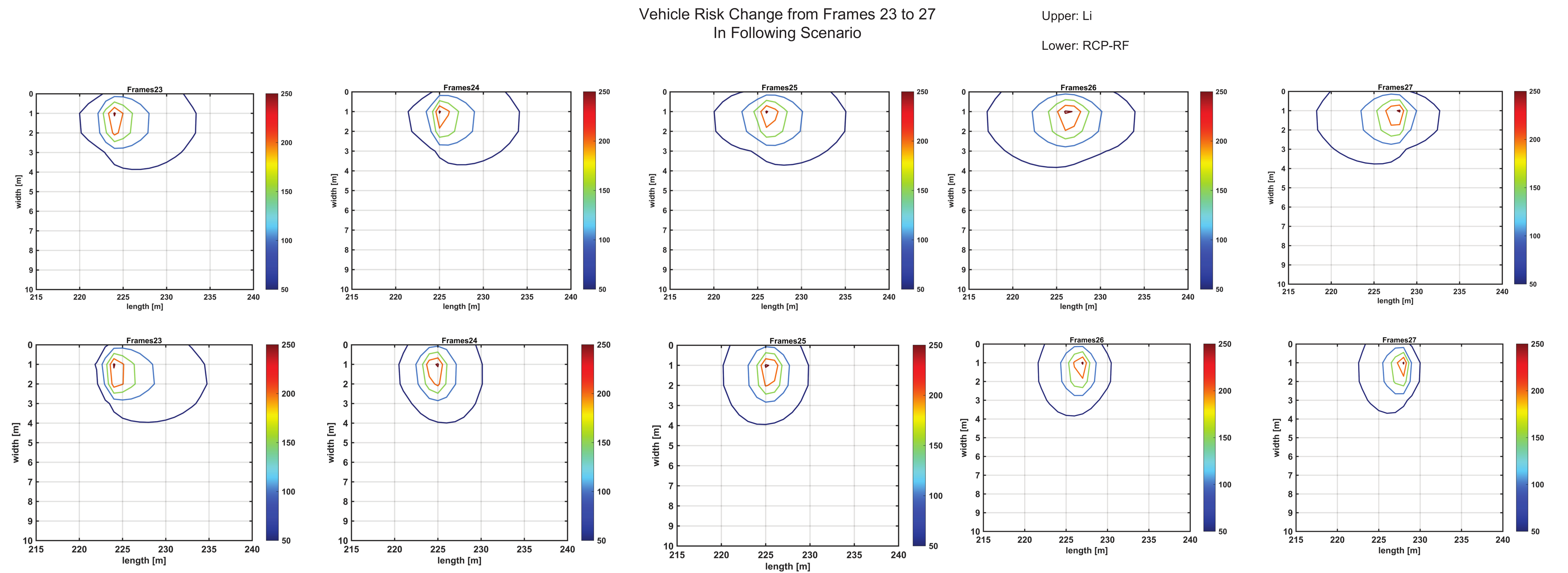}
\caption{Risk model results for following scenario in consecutive five frames, the upper one is from Li \cite{LiGan-423}; the lower lines shows the results of RCP-RF model of this paper.}\label{fig23}
\end{figure*}

Car following is one of the most usual traffic scenario in the road, a specific traffic following scenario is given in Fig. \ref{fig21}, the red car1259 is the ego-car, and the other is the obstacle car. To analyze the vehicle risk model performance in the car following situation, ego1259 and car1248 are selected, and the risk strength distribution of the obstacle1248 is calculated in five consecutive frames. To exemplify the example, car1248 and ego-car are in following situation, the accelerate and velocity of the ego maintain roughly smooth, while car1248's change obviously. Fig. \ref{fig22} illustrates the change of THW, accelerate, velocity and relative velocity of the two cars in five frames.

Using the data mentioned above, the risk results is shown in Fig. \ref{fig23}, the upper line shows the risk results generated by Li \cite{LiGan-423}, and the second line is the results generated by the RCP-RF model proposed in this paper. According to the model proposed in this paper, the risk strength is depended on both the obstacle car's motion state and the relative motion tendency between ego and obstacle, which is also reflected in the results in Fig. \ref{fig23}.  Combined Fig. \ref{fig22} and model theory in section \ref{subsec3.2}, the relative velocity $v_{re} = v_{obs} - v_{ego}$ changed from negative to positive from frame23 to 25, which leads the risk generated by car1248 decreases. Meanwhile, the accelerate of car1248 is positive all the time, so the risk distribution of the vehicle risk potential field is obviously inclined to front, which means the movement of car1248 will have more effect to its front area in its direction of movement. And this is totally in consistent with the real driving situation and drivers' feeling. From frame 26 to 27, the accelerate of car1248 changes from positive to negative, while the relative velocity stays positive. As a result, the risk field distribution still maintains a limited range, and shows a "backward incline" characteristic, which is completely opposite to the field shape in frame 23 to 25. Under this situation, the movement of car1248 is more likely to generate risk for the area in the back of car1248's movement direction, and this is also consistent with common people's acknowledge.

The first line in Fig. \ref{fig23} is the risk strength generated by Li \cite{LiGan-423} model using the same vehicle data. Compared to RCP-RF's results of this paper,  at frame23, when the obstacle car has a non-negligible effect to the ego, the risk field is almost same. However, when $v_{re}$ is bigger than zero, the leading car1248 should have less risk to the ego, and this is reflected in our model results by limiting the risk distribution range, but Li's model can not show the relationship of this. As show in the result pictures from frame 24 to 27, the risk field has a almost same distribution range with frame23, which leads to a over-conservative driving decision.

\begin{figure}[h]%
\centering
\includegraphics[width=\linewidth]{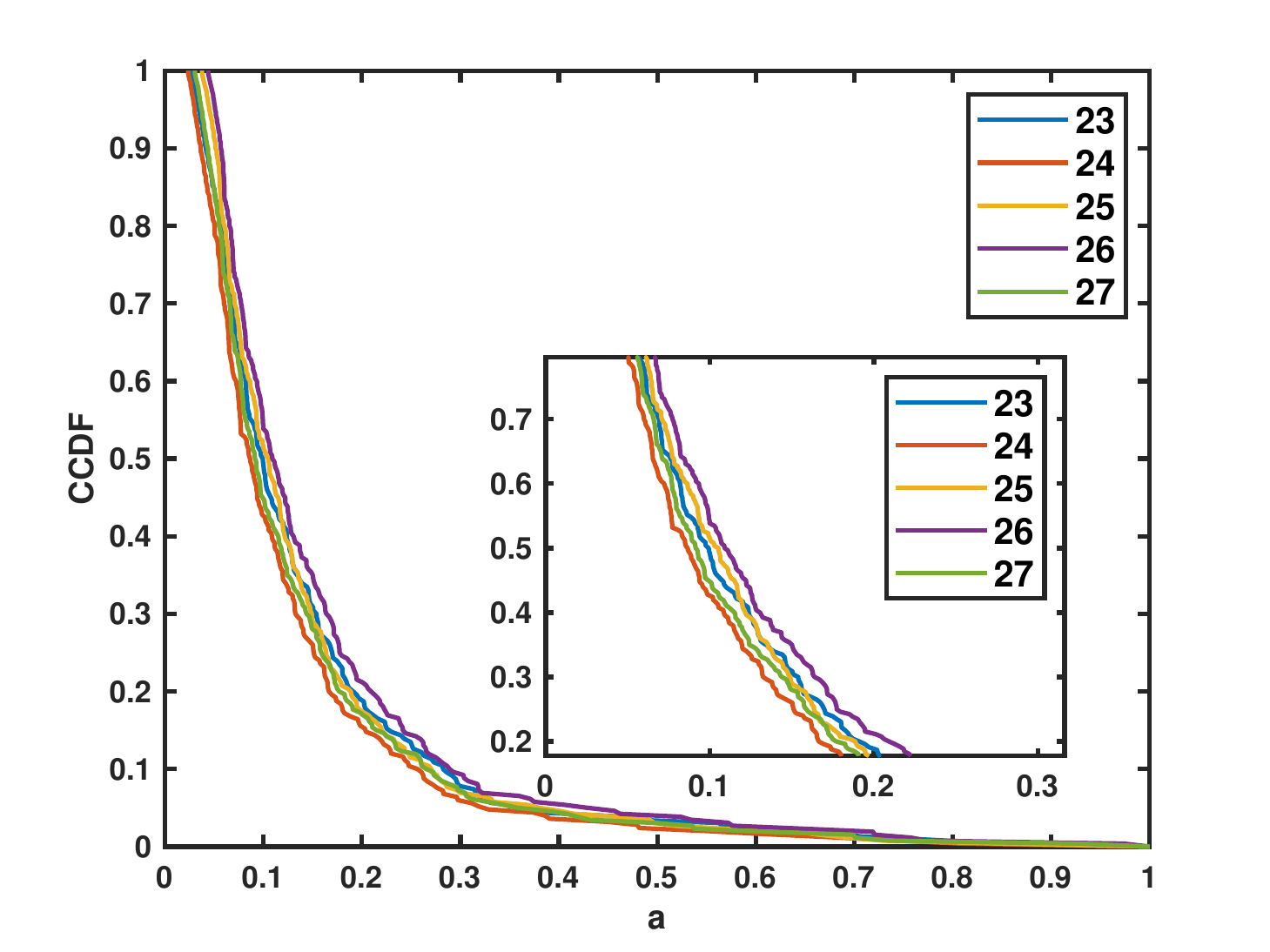}
\caption{CCDF risk metric of car1248 for car following scenario in consecutive frames in Fig. \ref{fig21} using Li model.}\label{fig24}
\end{figure}
\begin{figure}[h]%
\centering
\includegraphics[width=\linewidth]{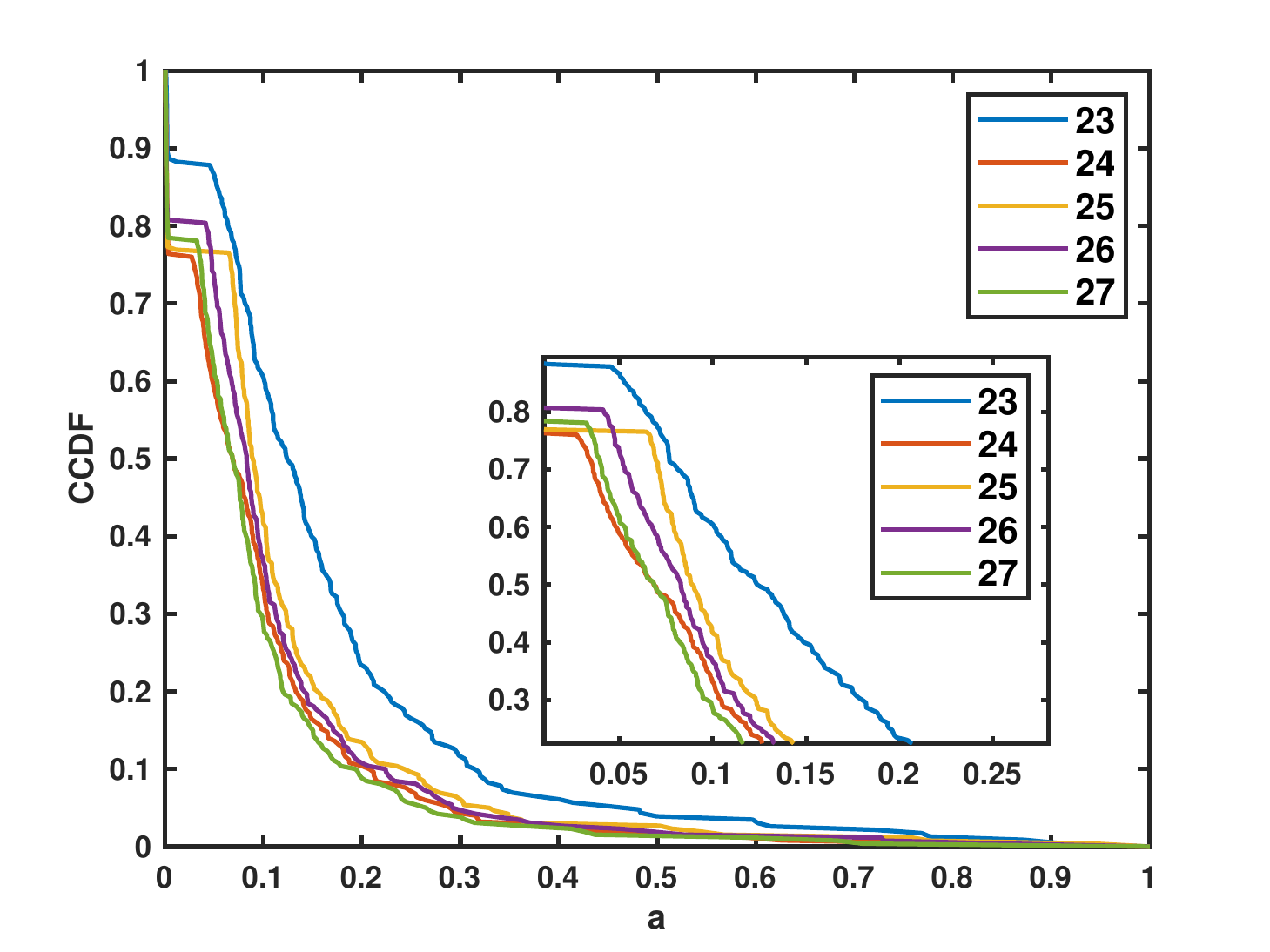}
\caption{CCDF risk metric of car1248 for car following scenario in consecutive frames in Fig. \ref{fig21} using RCP-RF.}\label{fig25}
\end{figure}

Fig. \ref{fig24} and Fig. \ref{fig25} represent the CCDF-based risk metric for the obstacle in the follow scenario, Fig. \ref{fig24} is the results using Li risk field model, and Fig. \ref{fig25} is the results using RCP-RF model. The difference of two model's results based on CCDF curve is obvious, from RCP-RF results, the risk metric change for obstacle car in the consecutive five frames is clear, while the CCDF risk curves based on Li model have little discrimination, and they are hard to estimate  the risk difference of nearby obstacle cars.

\subsubsection{Analysis of vehicle risk model for car cut-in scenario}\label{subsubsec4.1.4}

\begin{figure}[h]%
\centering
\includegraphics[width=\linewidth]{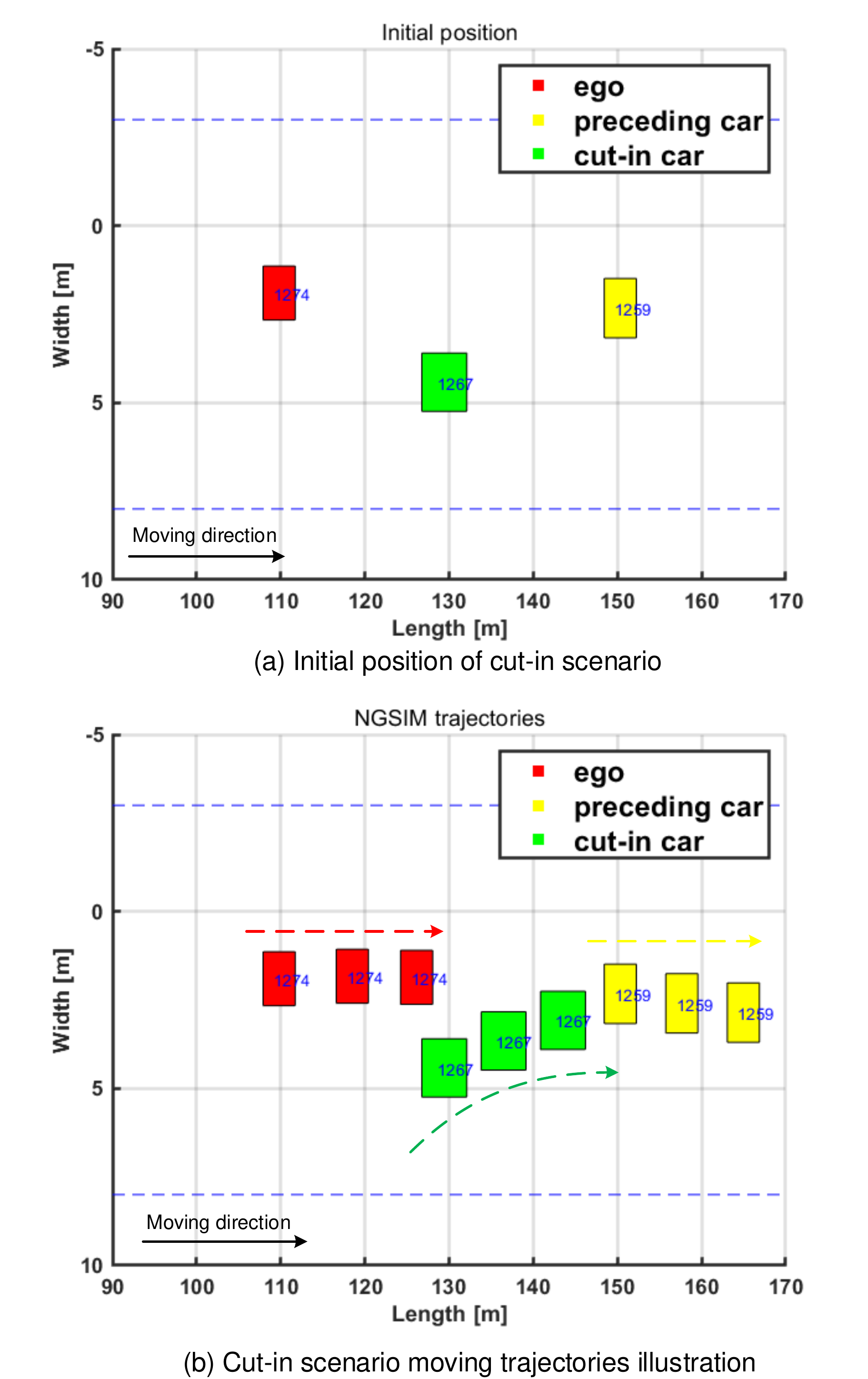}
\caption{Vehicle cut-in scenario, green car1267 is the cut-in car, red car1274 and yellow car1259 keep the same lane.}\label{fig26}
\end{figure}

\begin{figure}[h]%
\centering
\includegraphics[width=\linewidth]{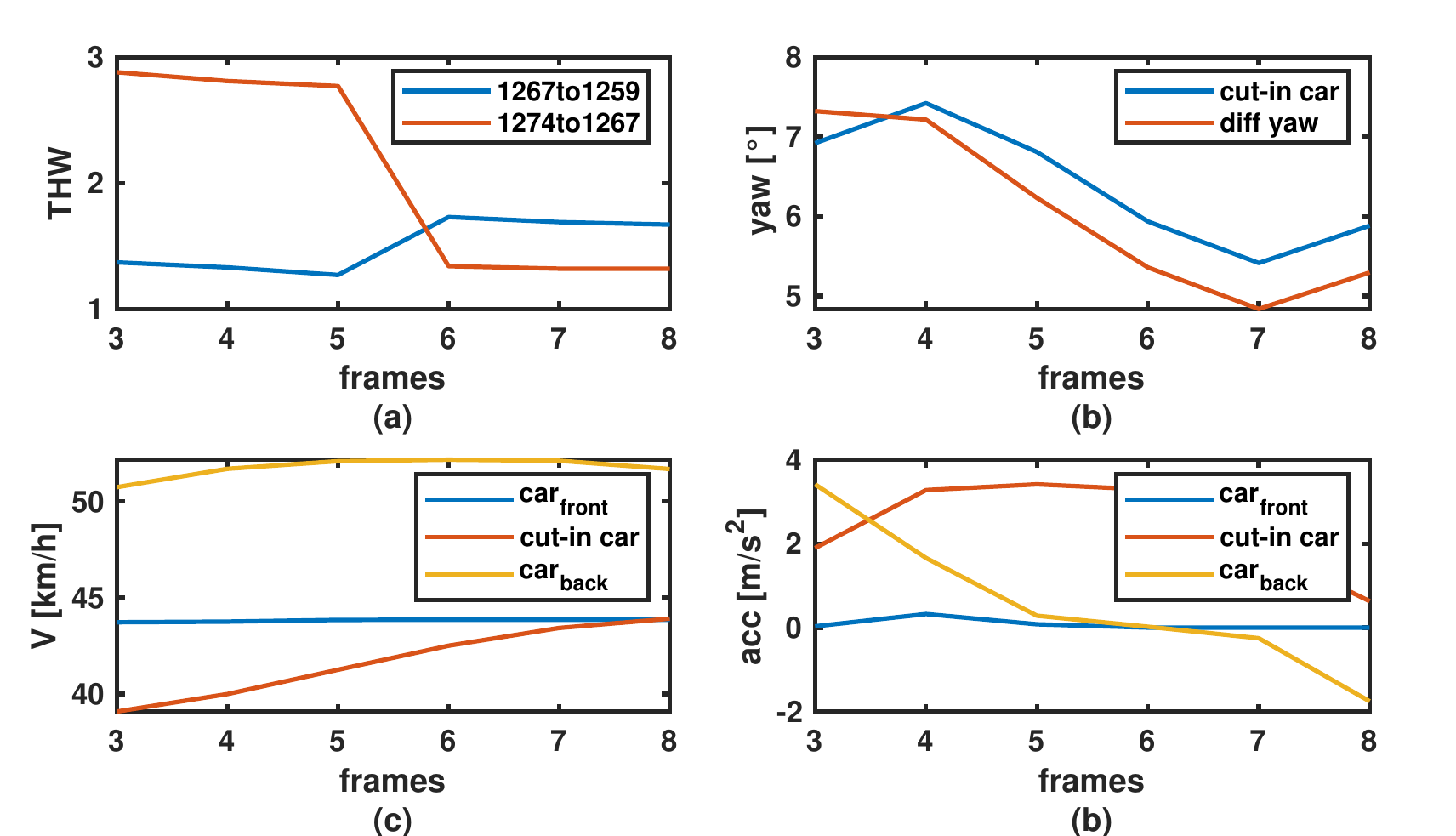}
\caption{Related motion state relationship of three cars. (a) is the THW from 1267 to 1259 and 1274 to 1267 separately; (b) is the heading yaw change curve for 1267, and relative yaw between 1267 and 1274; (c) is the velocity of three cars; (d) is the accelerate of three cars.}\label{fig27}
\end{figure}

Besides the car following scenario, cut-in situation is also tested with the model, Fig. \ref{fig26} gives the example. $(a)$ gives the vehicles' initial position, and $(b)$ shows the moving trajectories of three cars in this cut-in scenario.
Car1274 and car1259 are in the same lane all the time, while green car1267 is in the different lane at the beginning, and will cut into the middle of car1274 and car1259. 
Fig. \ref{fig27} provides the motion state of the three cars in successive frames, and the green car change to the same lane at frame6. 
As it shows, the car in the front is little affected by the cut-in action, which happens in the back of it, its velocity and accelerate maintain basically stable. 
Therefore, at this point, we set red car1274 as the ego, and analyze the risk distribution generated by cut-in car1267, since the front cut-in car have more influence on the car behind it.

\begin{figure*}[ht]
\centering
\includegraphics[width=\linewidth]{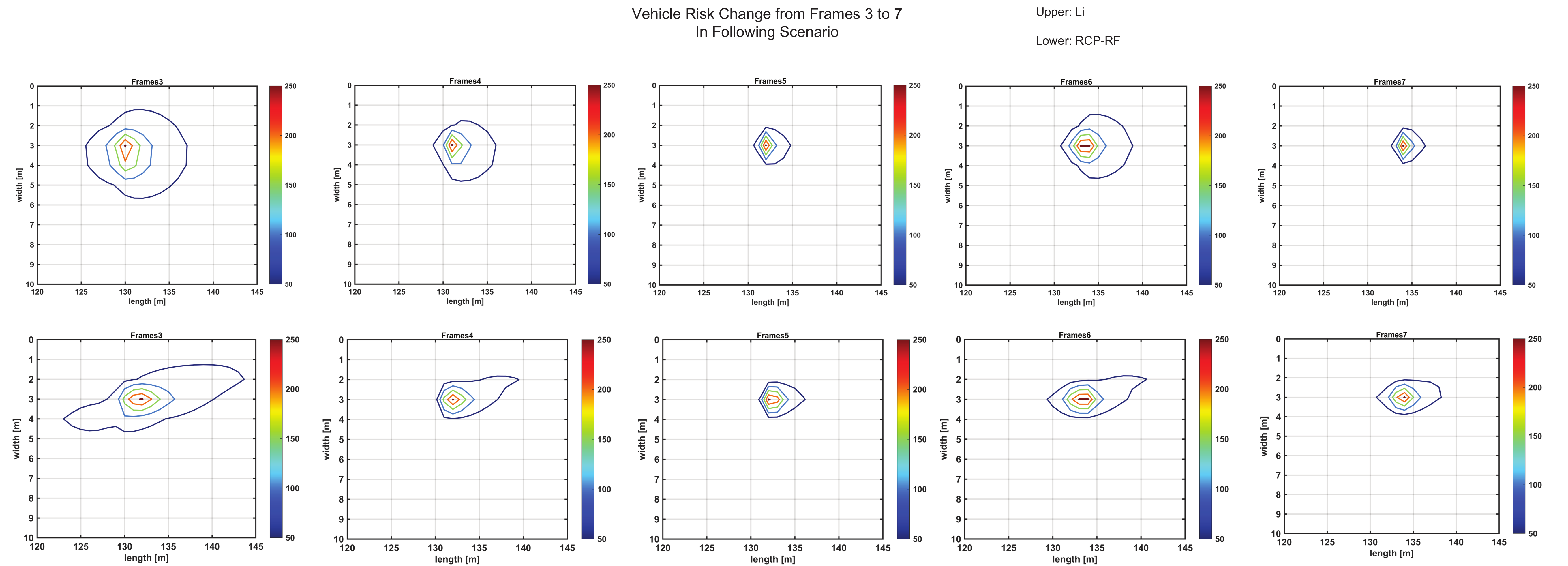}
\caption{Risk model results for cut-in scenario in consecutive five frames, the upper one is from Li \cite{LiGan-423}; the lower lines shows the results of RCP-RF model of this paper.}\label{fig28}
\end{figure*}
\begin{figure}[h]%
\centering
\includegraphics[width=\linewidth]{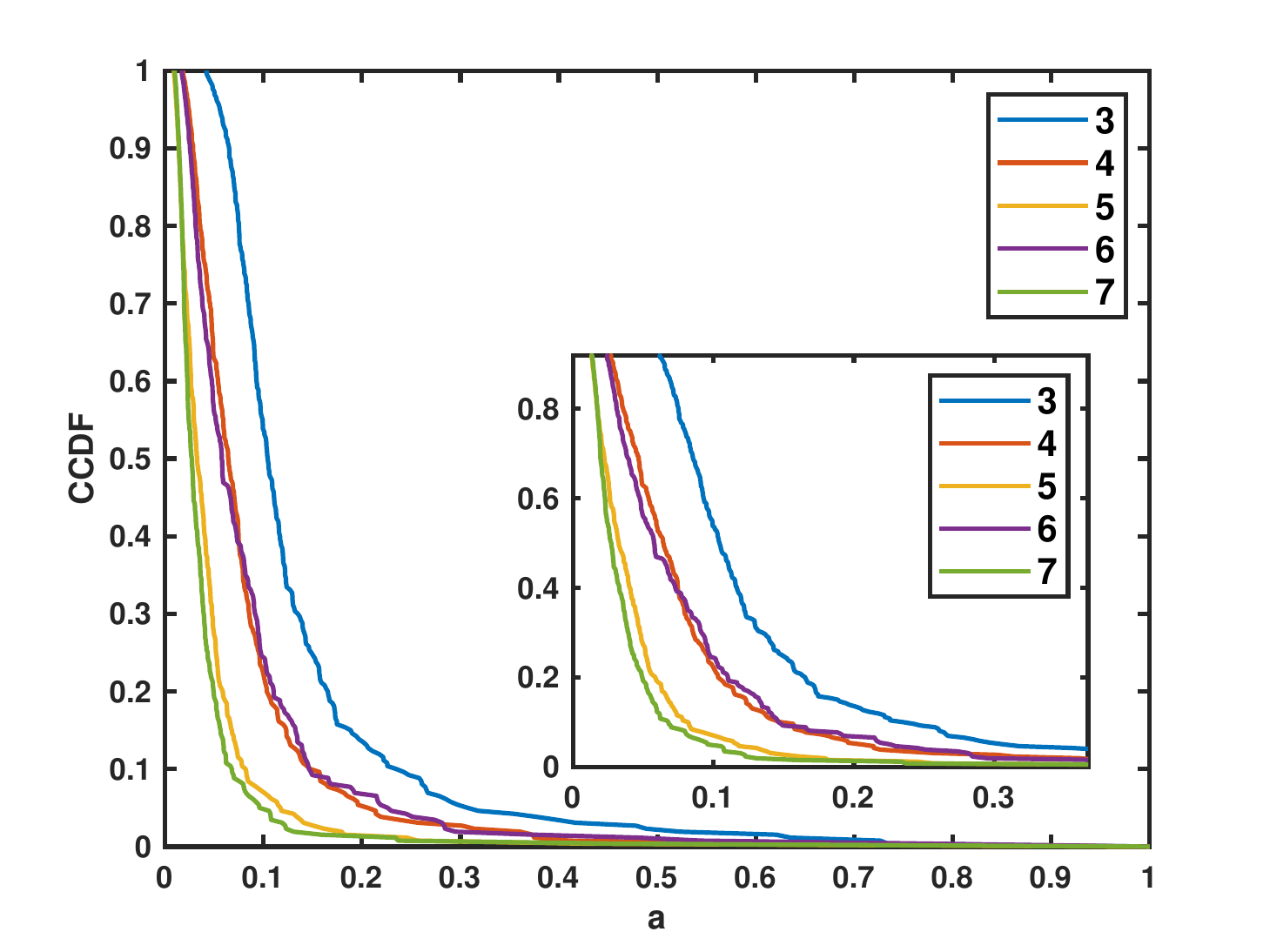}
\caption{CCDF risk metric of car1267 for car cut-in scenario in consecutive frames in Fig. \ref{fig26} using Li model.}\label{fig29}
\end{figure}
\begin{figure}[h]%
\centering
\includegraphics[width=\linewidth]{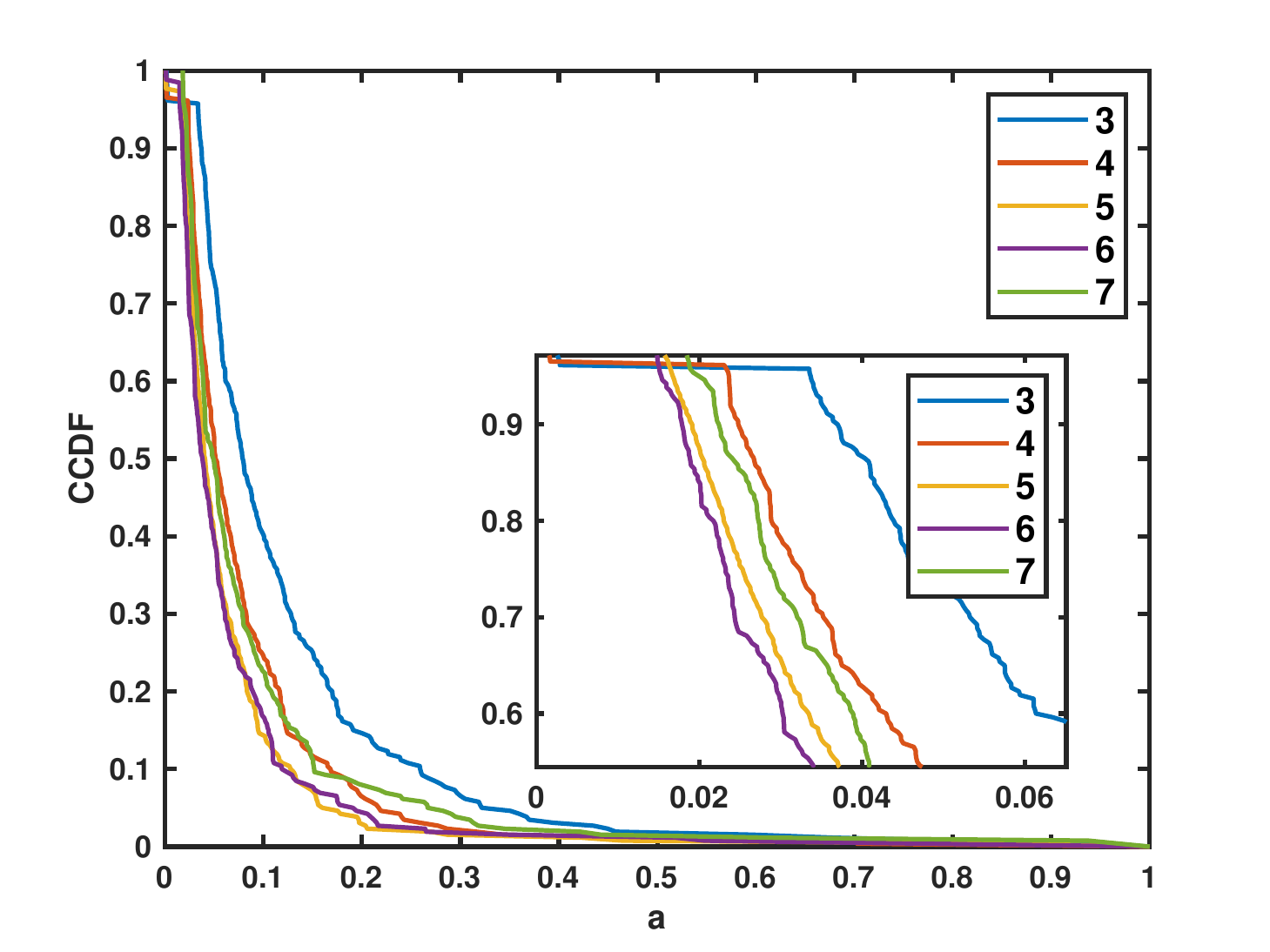}
\caption{CCDF risk metric of car1267 for car cut-in scenario in consecutive frames in Fig. \ref{fig26} using RCP-RF model.}\label{fig30}
\end{figure}

With the same settings as Fig. \ref{fig23}, Fig. \ref{fig28} shows the two risk model results using the same vehicle data mentioned above. Different from the following situation, in the cut-in scenario, the heading angle of the object car is not equal to zero, which leads the rotation of the risk field, and the result pictures show the difference clearly. From frame3 to 5, car1267 is at the different lane with the car1274, and tries to cut in to the same lane in front of car1274. As described in algorithm \ref{algo1}, the main factor that may affect the risk field when two cars are at different lane is the relative motion tendency between them. In other words, if the obstacle car has the tendency to approach the ego, it will generate more risk to ego-car than when it dose not have such motion tendency. Therefore, at Frame3-6, when car1267 has the tendency to change lane, it approaches to the ego, which makes its risk field presents an extension shape. The extension range is related to the relative velocity $v_{re} = v_{obs} - v_{ego}$, which is smaller than zero all the time at this scenario but decreases in this progress, that is why the risk distribution range becomes smaller from Frame3 to 5. At Frame6 and 7, car1267 finished its lane changing behavior, and it goes back to following scenario with car1274. Thus, the risk distribution obeies to the rule mentioned in the following scenario, and the pictures Frame6 and 7 give the risk results considering the relative velocity and accelerate. And the results conform to the expectation.

The first line of Fig. \ref{fig28} also shows the results using Li's model. Compare to the pictures in the first line, they show the obviously difference risk distribution characteristics. The risk field generated by Li \cite{LiGan-423} only depends on the velocity and accelerate, while the risk field generated by this paper's model can reflected the motion state and tendency better. 

Fig. \ref{fig29} and Fig. \ref{fig30} also show the CCDF-based risk metric results of two models for car cut-in scenario. And it is also obvious that the risk curves generated by RCP-RF model can reflect the imperceptible change of the vehicle risk better, while the curves generated by Li's model always have the roughly same tendency. The conclusion is consistent with all previous sections'.

\subsection{Vehicle Model experiments on real AV platform}\label{subsec4.2}
\subsubsection{Experiments setup}\label{subsubsec4.2.1}
For better evaluating the effectiveness and superiority of RCP-RF, we also conducted experiments based on real-world AV driving platforms.
Two same-type cars were used and each of them installs necessary sensors for autonomous running. One is as the ego car and the other is the obstacle car.

\begin{figure}[hptb]
  \centering
  \includegraphics[width=\linewidth]{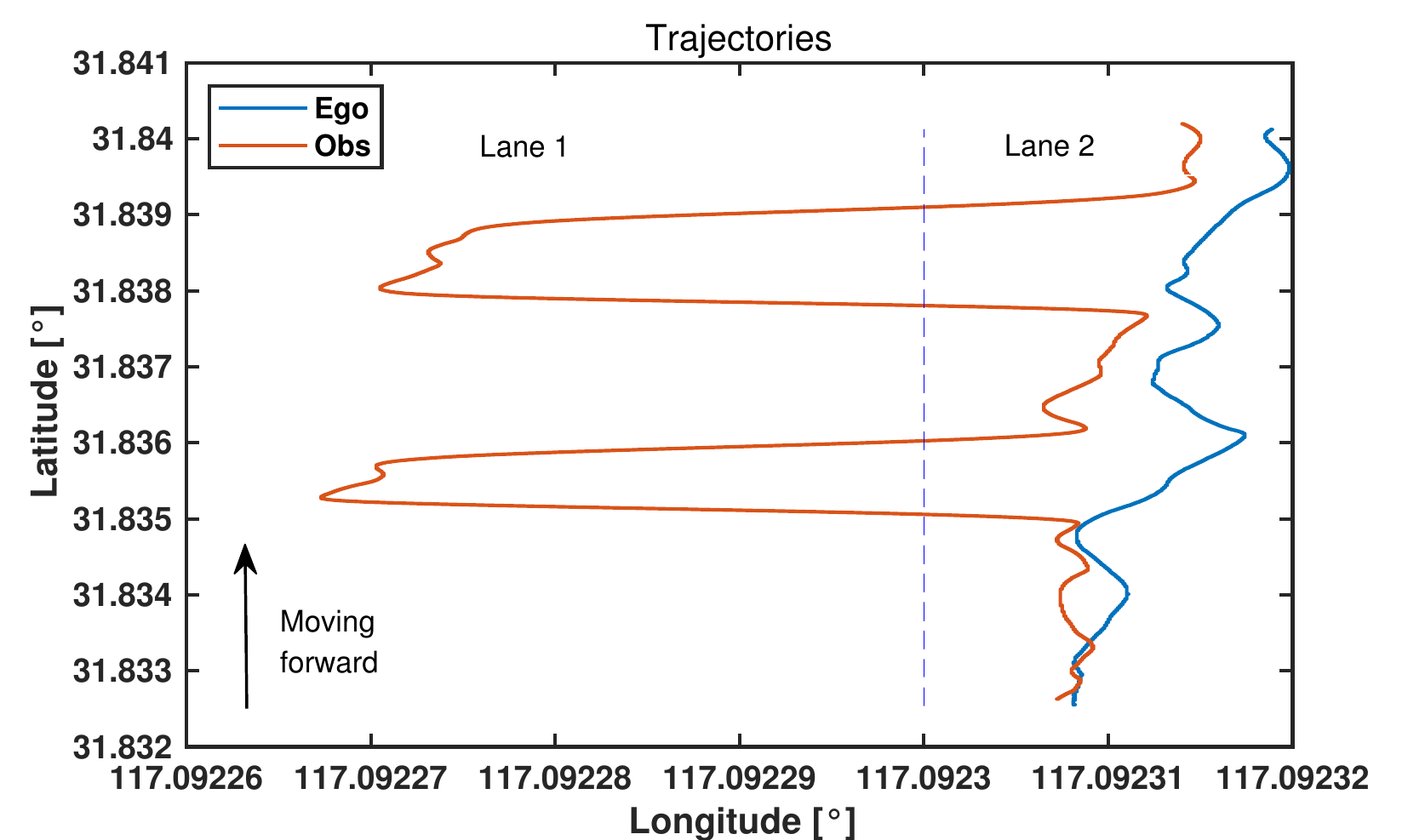}
  \caption{The whole driving trajectories of the experiment.}\label{fig32}
\end{figure}

The whole traffic experiment was conducted on one common urban road in Hefei and contains car following and cut-in scenarios.
Fig. \ref{fig32} shows the whole trajectories of the experiment. 
Two cars moved in the north-south direction. Blue lines show the trajectory of the ego car, it kept moving forward in $Lane 2$;
The red one is obstacle's trajectory, it started moving at $Lane 2$, then changed to $Lane 1$, and moved back to $Lane 2$ to finish the cut-in process, this process repeated twice.

The driving data collected contains the necessary information for RCP-RF, such as the motion state of two cars.
To better evaluate the performance of RCP-RF, we select two representative car following and cut-in segments from the whole driving data. 
And the following two sections will give more details about the results.
Similar to the previous experiment in section \ref{subsec4.1}, all the results of RCP-RF are compared with Li \cite{LiGan-423} to prove RCP-RF model's superiority.

\subsubsection{Real-world car following analysis}\label{subsubsec4.2.2}
\begin{figure}[hptb]
  \centering
  \includegraphics[width=\linewidth]{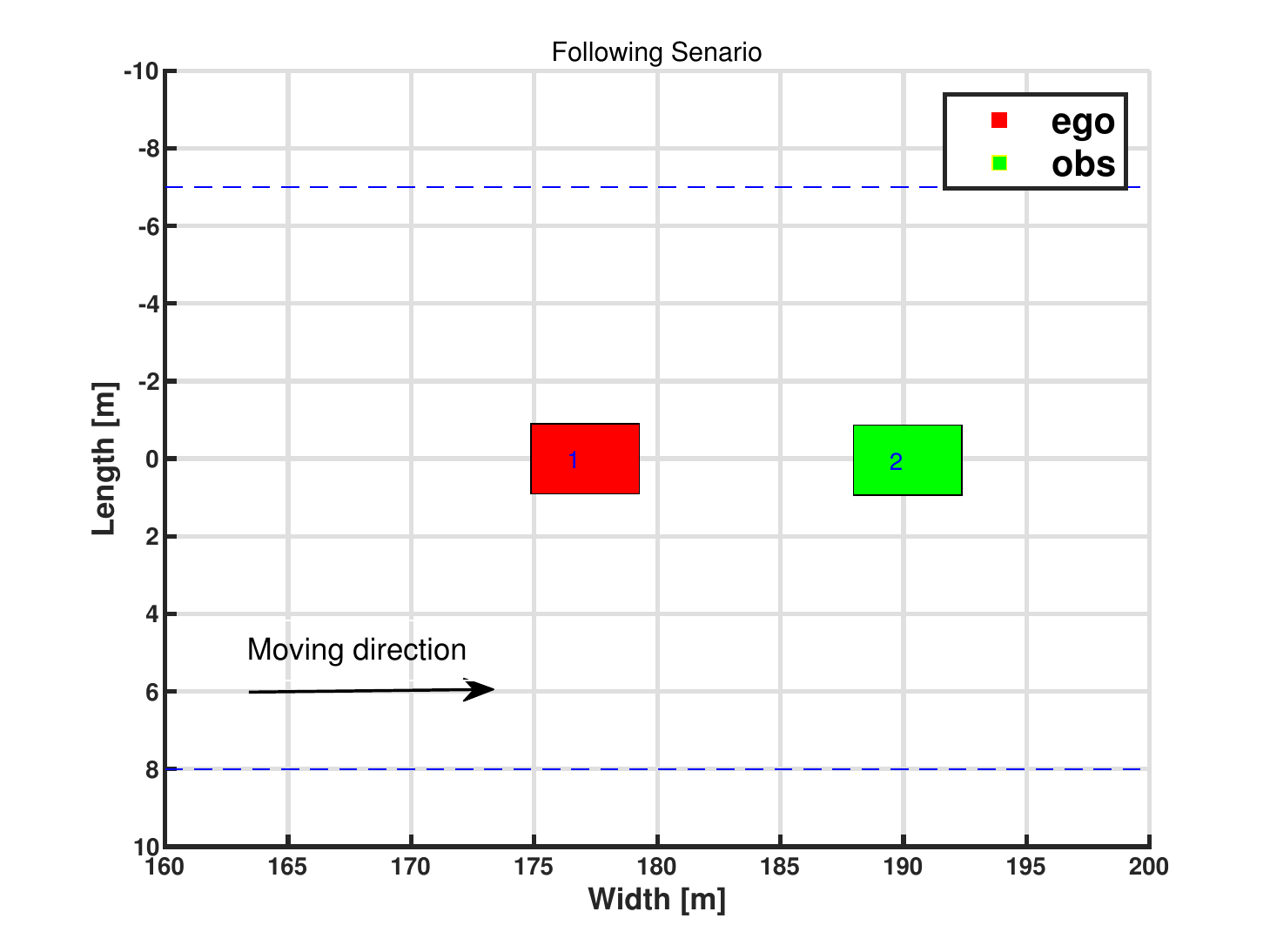}
  \caption{Real-world car following situation. Red car labeled $1$ is ego; green car labeled $2$ is obstacle.}\label{fig33}
\end{figure}

\begin{figure}[hptb]
  \centering
  \includegraphics[width=\linewidth]{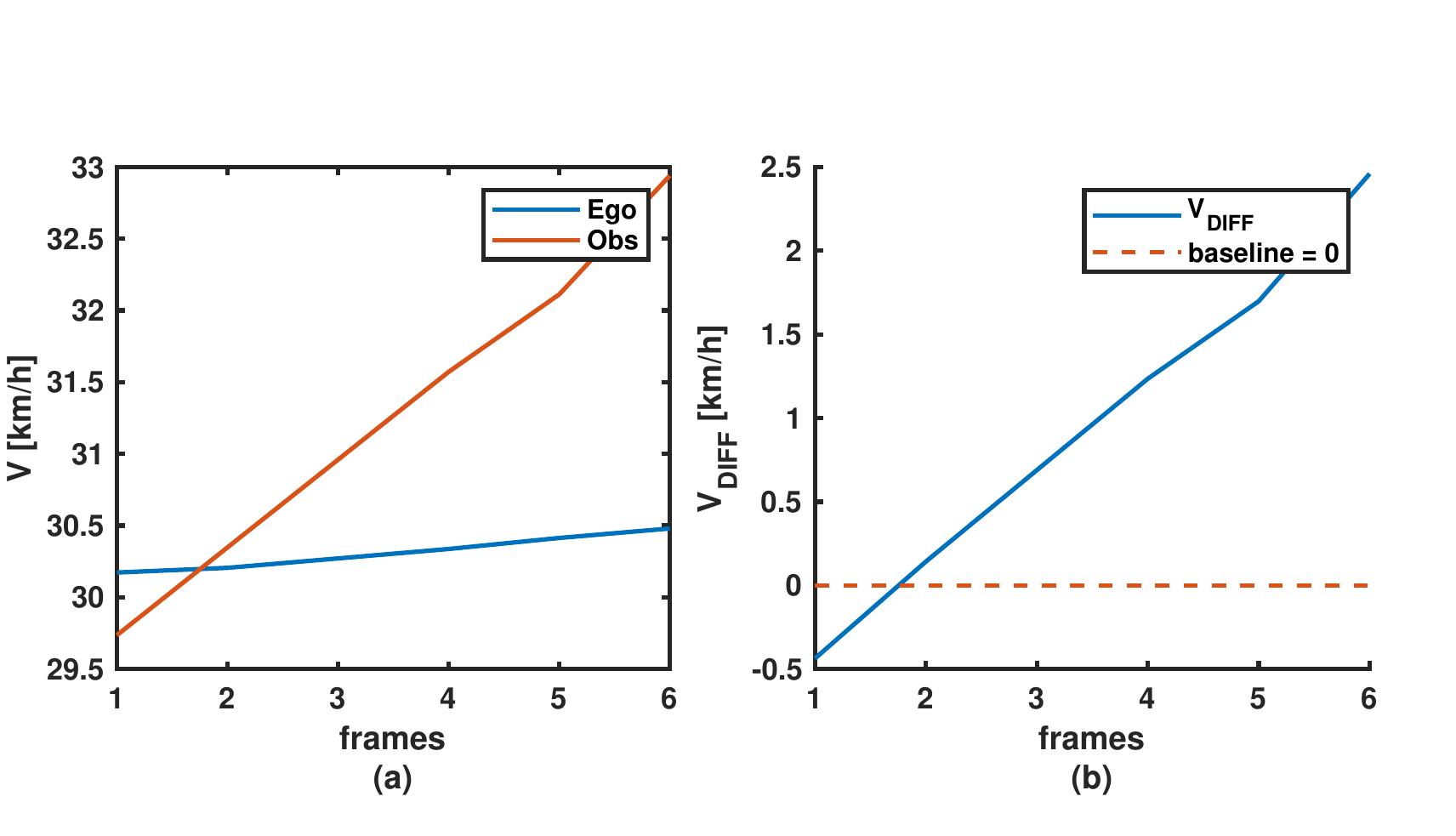}
  \caption{Relative velocity relationship. $(a)$ is the velocty of two cars; $(b)$ is the velocity difference change.}\label{fig34}
\end{figure}

The illustration of car following scenario is show in Fig. \ref{fig33}, red ego car followed the green preceding car. 
Fig. \ref{fig34} shows the velocity relationship of two cars during the following proess.
The ego car kept a realtive same velocity, while obstacle car started at a slower speed, then kept accelerating in the process. 

As discussed in the section \ref{subsec3.2} and section \ref{subsubsec4.1.3}, 
since the obstacle car was approaching the ego when its velocity was slower then the ego;
after the obstacle speeded up, it tended to leave away the ego,
so the risk generated by obstacle car should reach highest at the beginning, and then descending.

\begin{figure}[hptb]
  \centering
  \includegraphics[width=\linewidth]{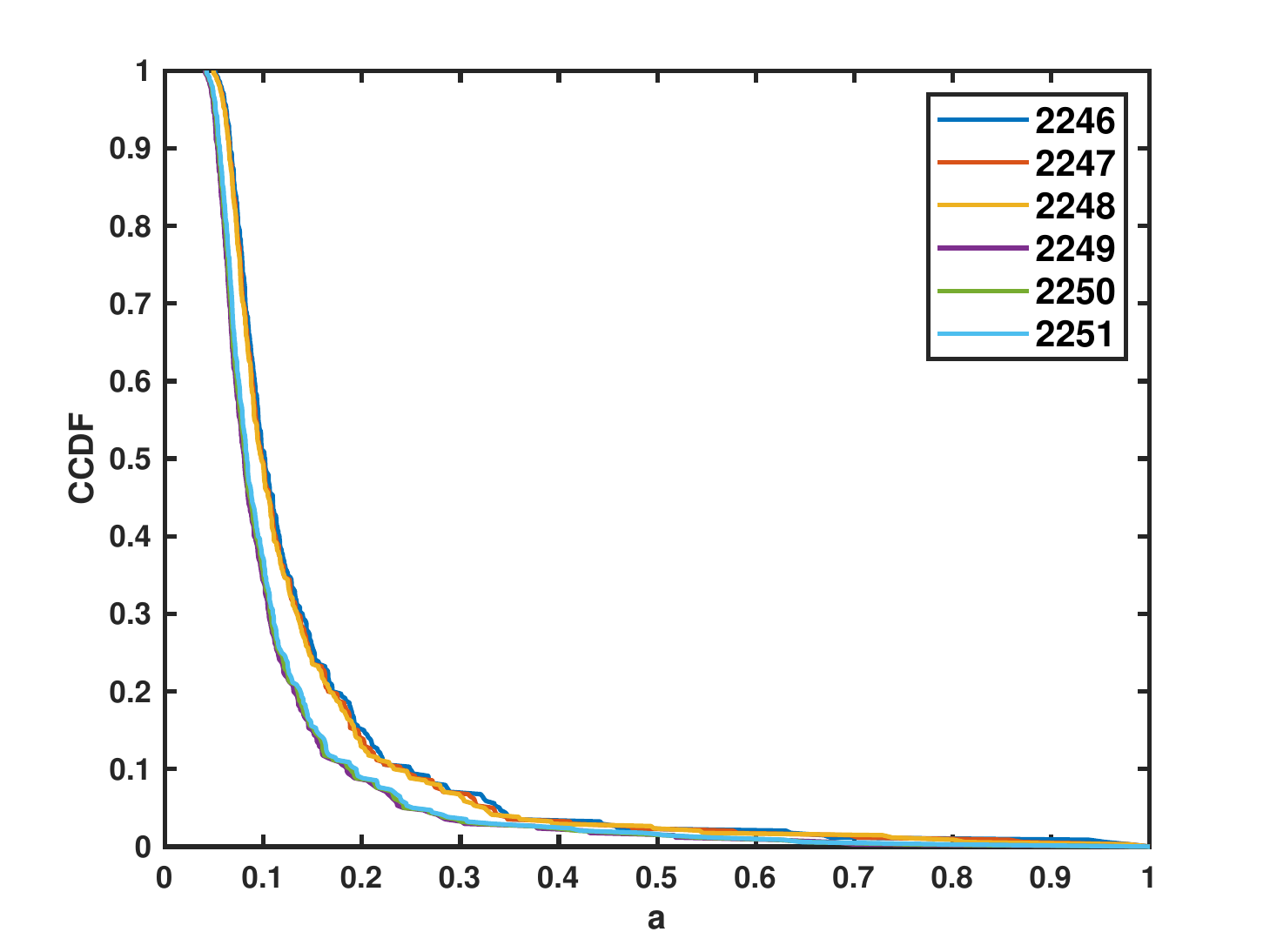}
  \caption{CCDF curve results for car following scenario in Fig. \ref{fig33} using Li model.}\label{fig35}
\end{figure}
\begin{figure}[hptb]
  \centering
  \includegraphics[width=\linewidth]{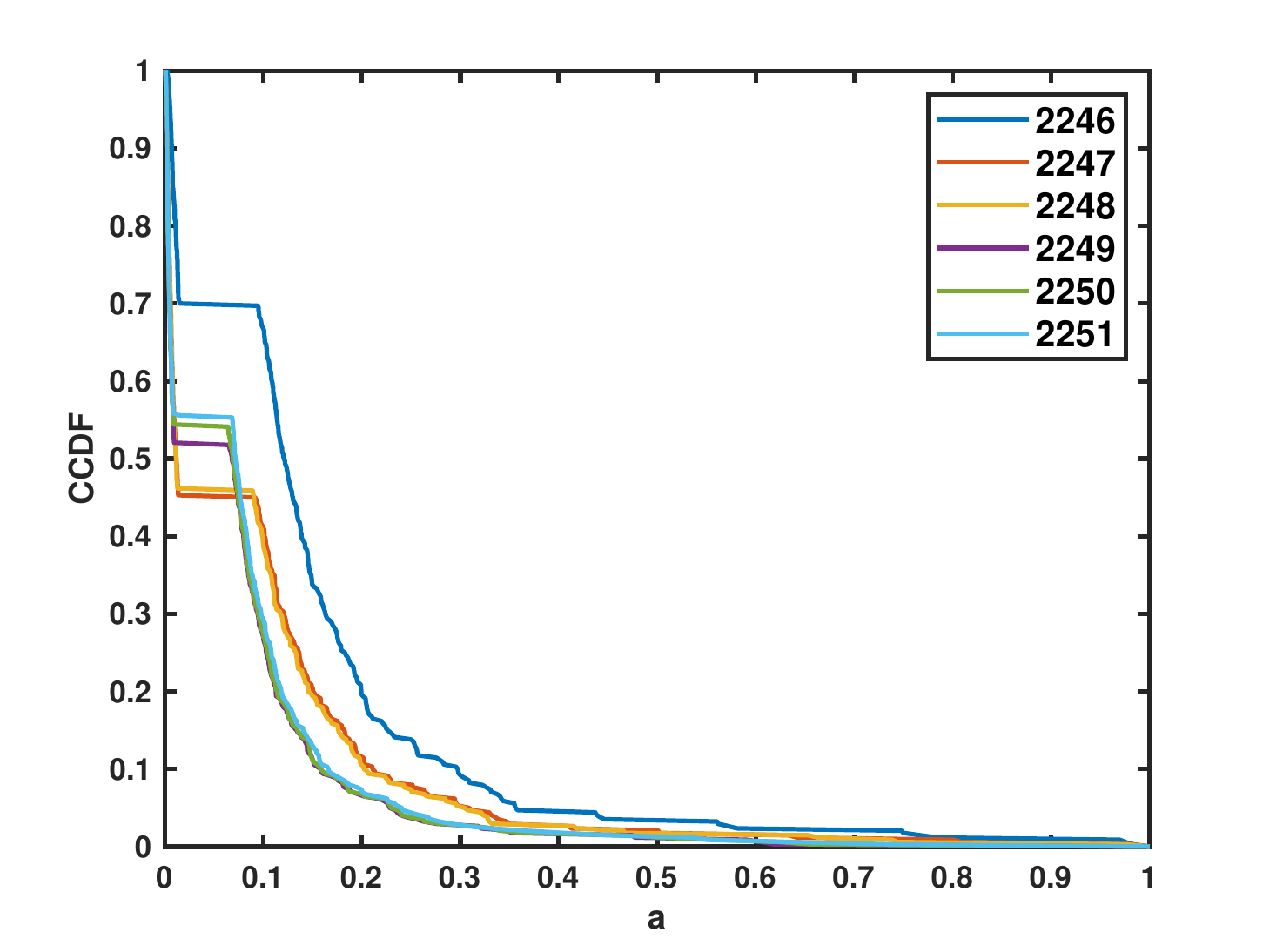}
  \caption{CCDF curve results for car following scenario in Fig. \ref{fig33} using RCP-RF model.}\label{fig36}
\end{figure}
\begin{figure}[hptb]
  \centering
  \includegraphics[width=\linewidth]{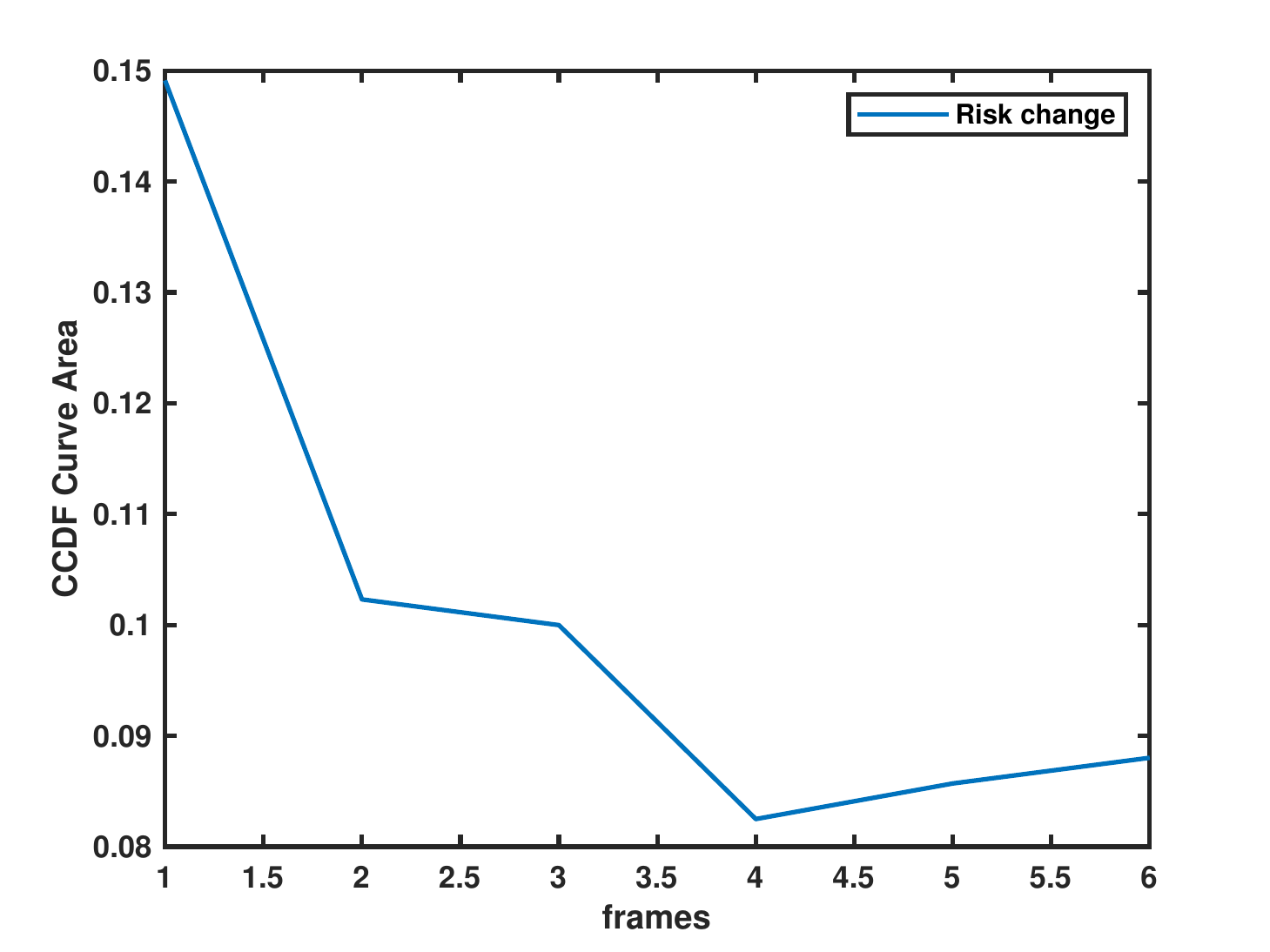}
  \caption{CCDF curve area change results for car following scenario in Fig. \ref{fig33} using RCP-RF model.}\label{fig37}
\end{figure}
For results clarification and avoiding repetition, we do not provide the visual risk distribution results at this section, which are much similar to Fig. \ref{fig23}.
Instead, CCDF curve results of six consecutive car-following frames using Li \cite{LiGan-423} and RCF-RF model are provided in Fig. \ref{fig35} and Fig. \ref{fig36}, which can be used to reflect the risk changes.
It is consistent with the previous results discussed in section \ref{subsubsec4.1.3} that the CCDF curves of Li's model have almost the same shape, while the CCDF curevs of RCP-RF can better reflect the risk changes during driving.
As disscussed at the end of the section \ref{subsec3.4}, the goal of the CCDF curves is to explain the influence of obstacles' risk and guide ego's driving decision.
The CCDF curves can provide risk information from multi aspects, such as its change rate and the areas between curves and coordinates. 
To simply give an illustration of how CCDF can describe the change of risk, the areas change between curves and coordinates of Fig. \ref{fig36} is provided in Fig. \ref{fig37}.
It is obvious that the result in Fig. \ref{fig37} conform to previous discussion that obstacle's risk is highest at the beginning, then declining.
Area change can give a rough description that how obstacle's risk changing, another factors are also need to be considered for final precise results, but these will not be discussed in this paper.

\subsubsection{Real-world car cut-in analysis}\label{subsubsec4.2.3}
\begin{figure}[hptb]
  \centering
  \includegraphics[width=\linewidth]{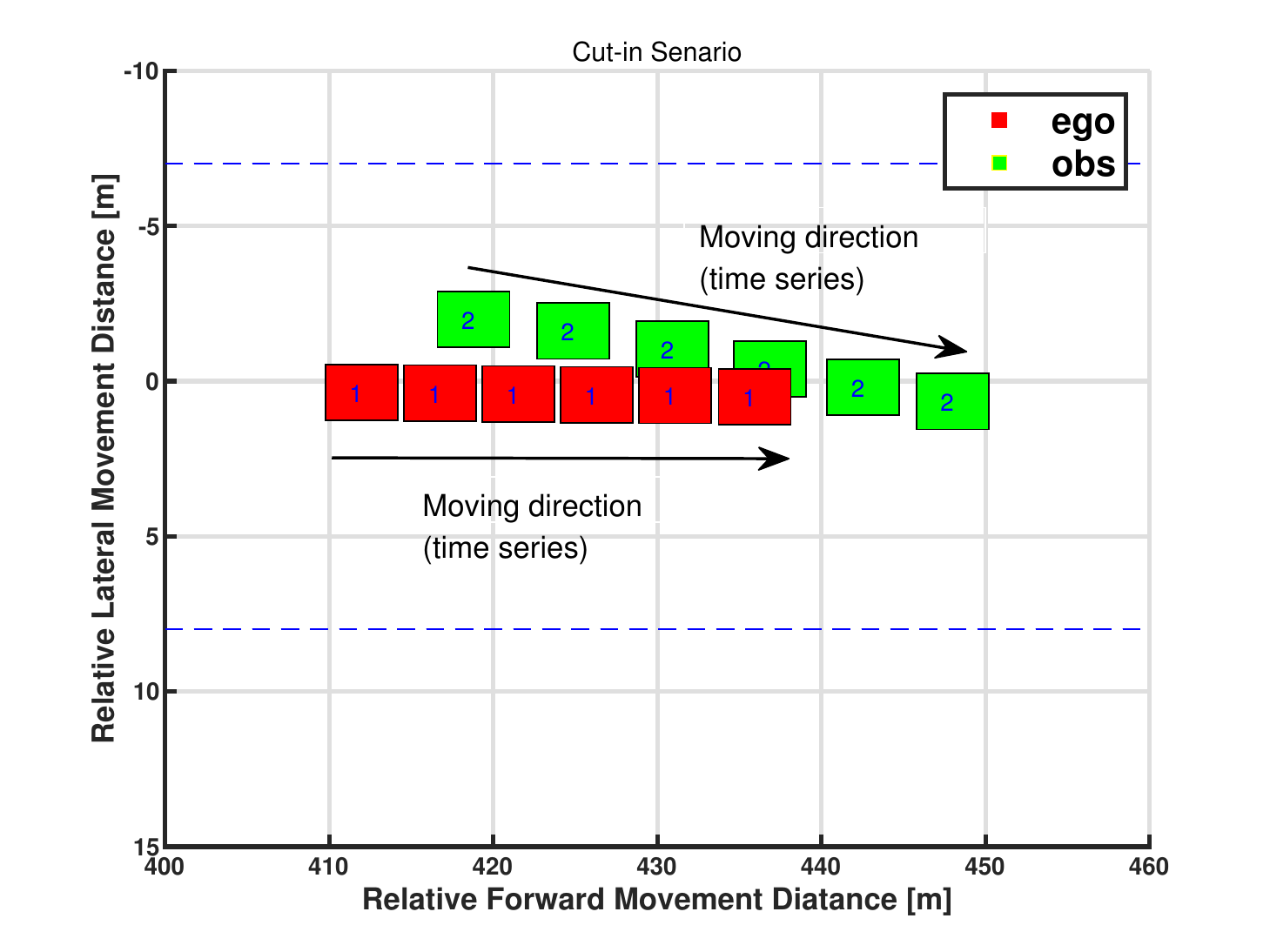}
  \caption{Real-world car cut-in trajectoryies in consecutive six frames.Red is the ego car kept moving forward; green is the obstacle car cut-in from nearby lane.}\label{fig38}
\end{figure}

Real-world car cut-in scenario risk analysis results are discussed in this section.
A cut-in segment from the whole driving series is selected to evaluate the risk model, Fig. \ref{fig38} shows the trajectories of the two cars in consecutive six frames. 
Red ego car kept moving straight, when the green obstacle car cut-in from the nearby lane and moved into the ego car's lane at four or five frames.
\begin{figure}[hptb]
  \centering
  \includegraphics[width=\linewidth]{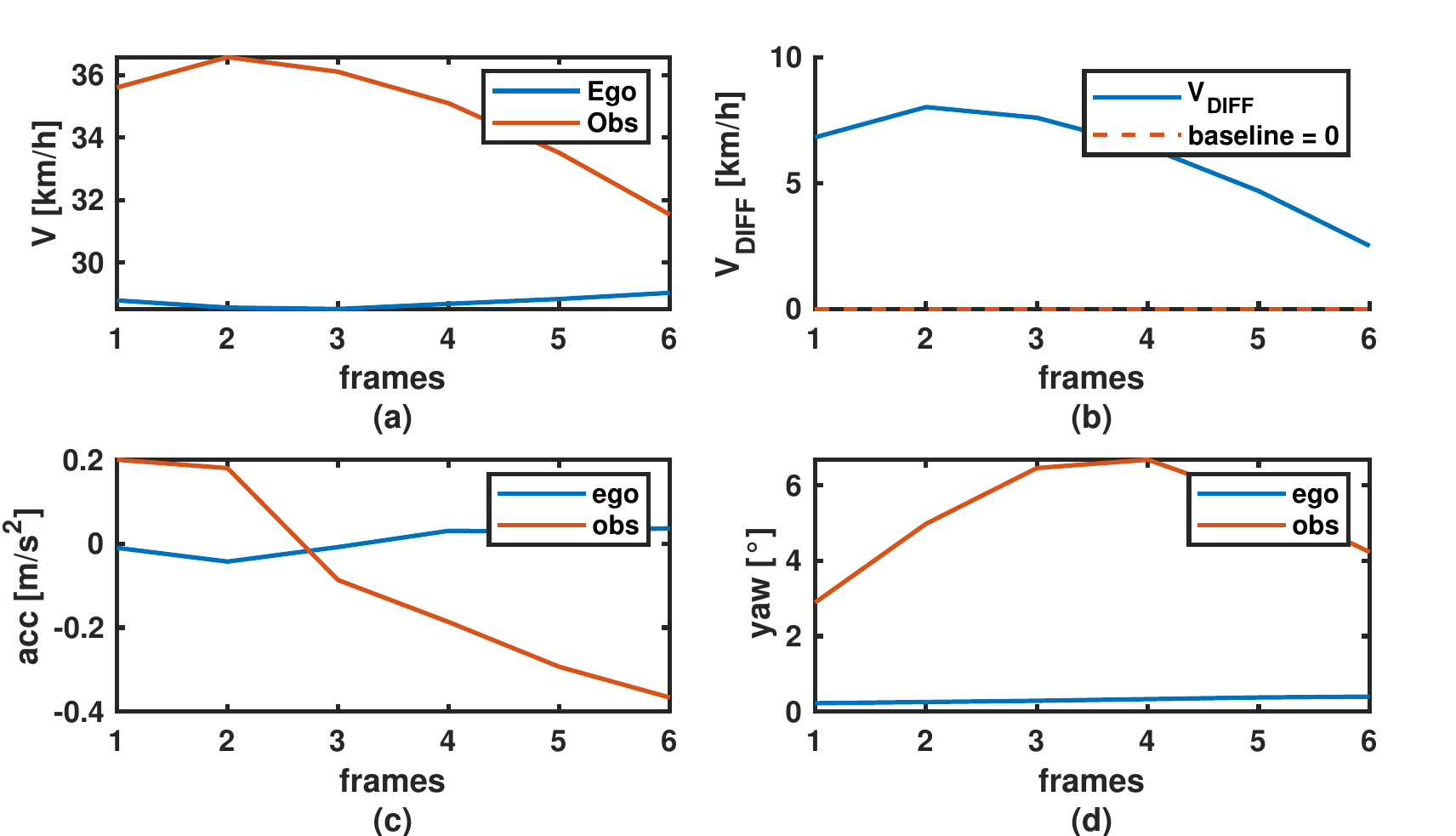}
  \caption{Relative motion state relationships.$(a)$ and $(b)$ shows the velocity relationship of two cars; $(c)$ shows the acceleration changes of two cars; $(d)$ is two cars' heading angle changing.}\label{fig39}
\end{figure}

Fig. \ref{fig39} describes the motion states of two cars during cut-in process, the ego kept stable motion state while the obstacle's motion kept changing. 
$(a)$, $(b)$ shows the obstacle car's velocity was always faster then the ego to finish the cut-in task.
$(c)$ shows that the obstacle car's acceleration changed from positive to negative; $(d)$ gives the heading yaw changes of the obstacle car.
The obstacle car started at the nearby lane, then moved to the ego's lane to be the preceding vehicle.
\begin{figure}[hptb]
  \centering
  \includegraphics[width=\linewidth]{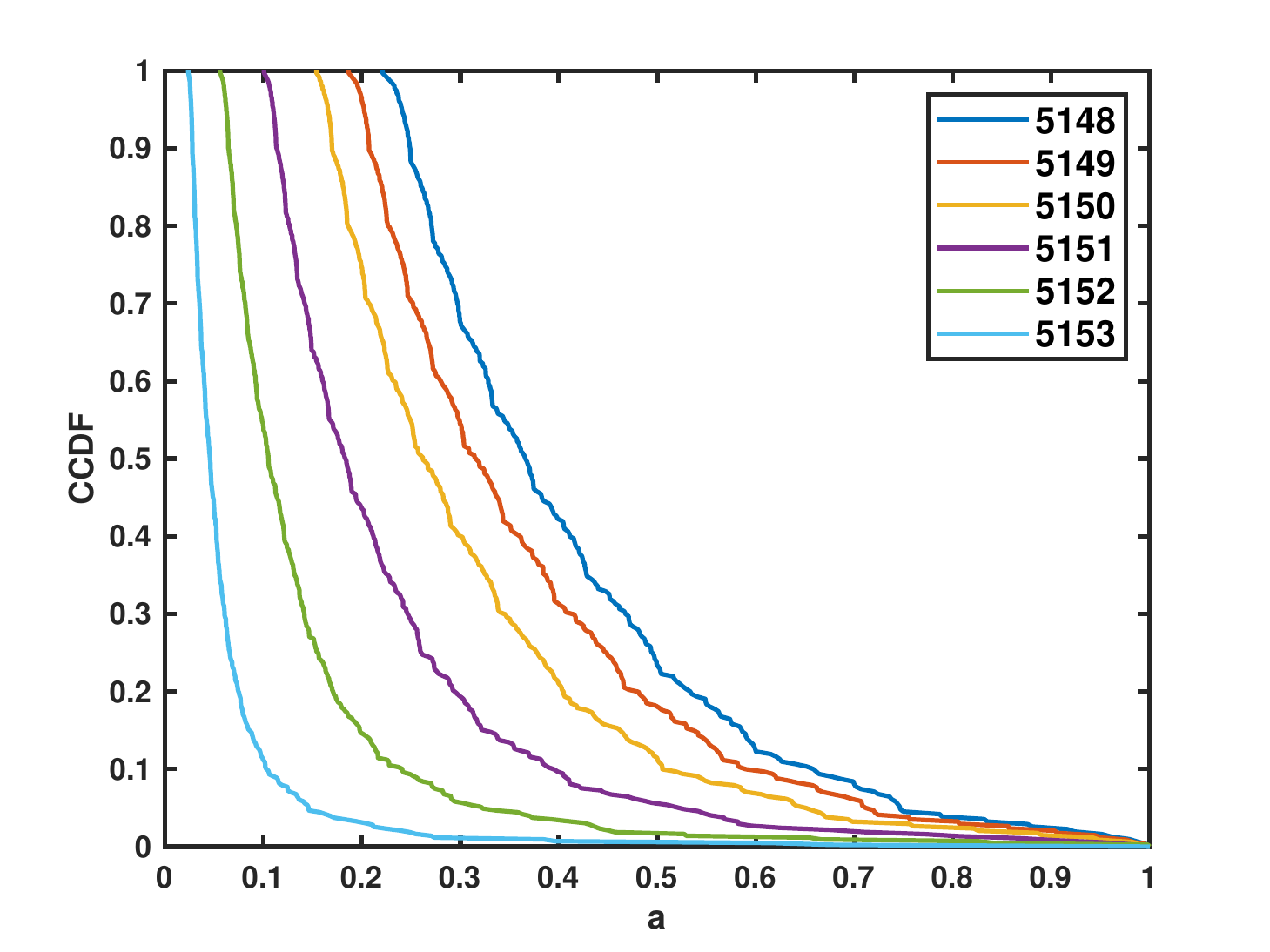}
  \caption{CCDF curve results for car cut-in scenario in Fig. \ref{fig38} using Li model.}\label{fig40}
\end{figure}
\begin{figure}[hptb]
  \centering
  \includegraphics[width=\linewidth]{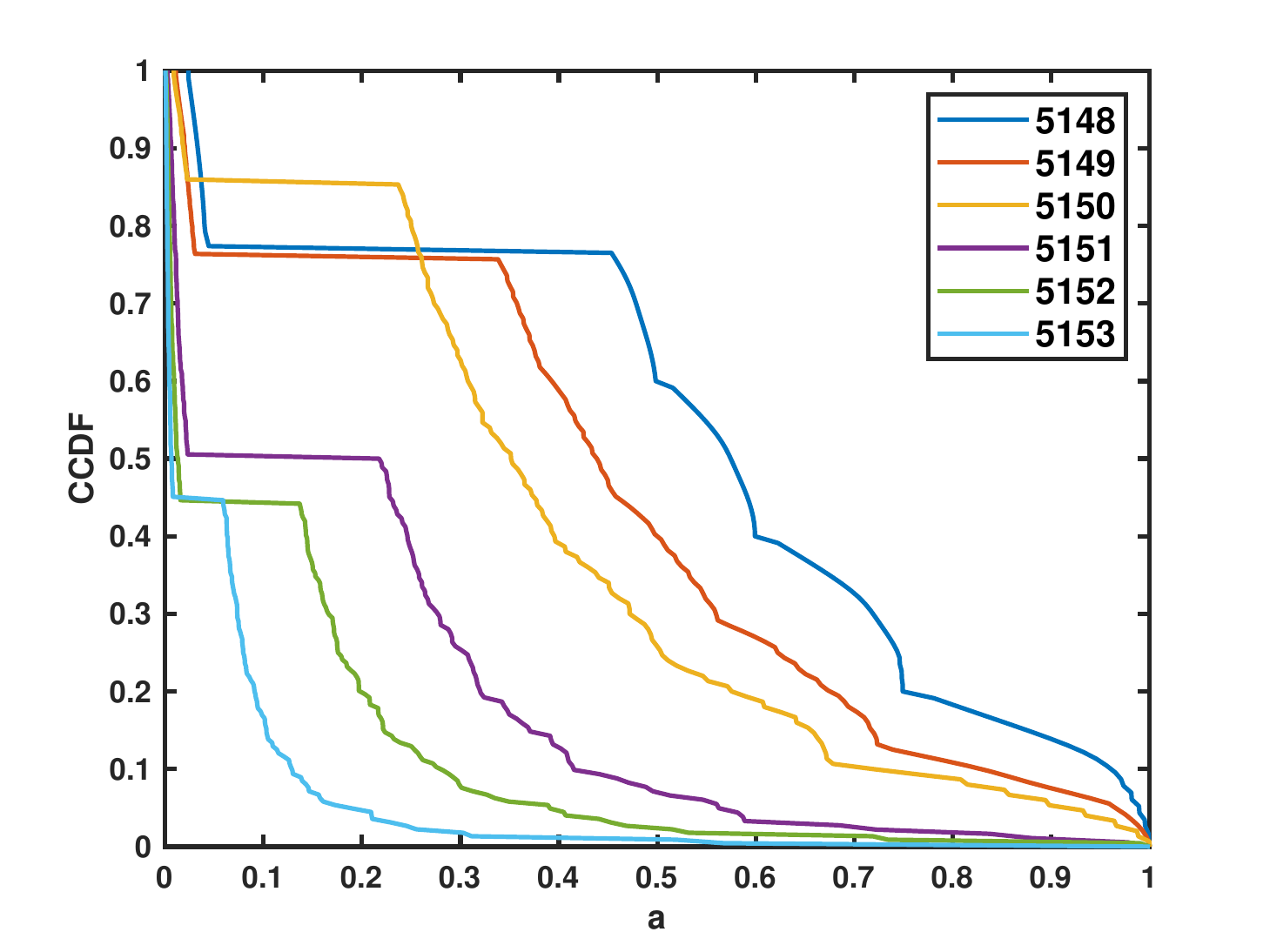}
  \caption{CCDF curve results for car cut-in scenario in Fig. \ref{fig38} using RCP-RF model.}\label{fig41}
\end{figure}
\begin{figure}[hptb]
  \centering
  \includegraphics[width=\linewidth]{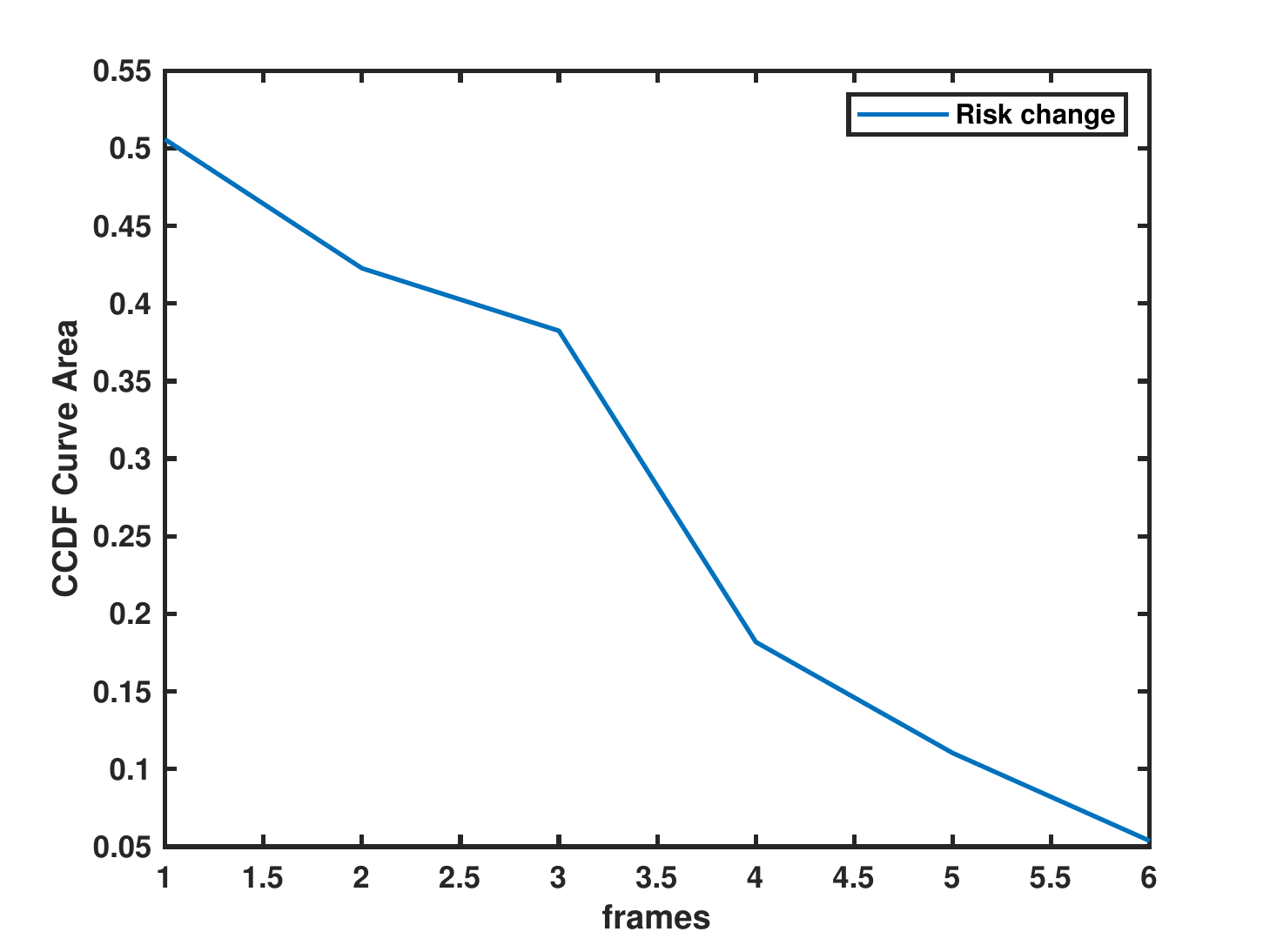}
  \caption{CCDF curve area change results for car cut-in scenario in Fig. \ref{fig38} using RCP-RF model.}\label{fig42}
\end{figure}

Same with the previous section \ref{subsubsec4.2.2}, only CCDF results of Li's and RCP-RF model are provided in Fig. \ref{fig40} and Fig. \ref{fig41}.
And the results are also consistent with previous discussion.
Compared to Li's results, CCDF curves of RCP-RF can better represents the risk changes caused by motion states changes and can provides more information of the environmental risk.
The illustration of CCDF curve areas changes using Fig. \ref{fig41} is also given in Fig. \ref{fig42}.
The overall tendency is descending. At the first three frames, green car was in the different lane with the ego and had a tendency to approach the ego, so the risks were relative higher.
After the forth frame, the green car moved to the ego's lane, and since its speed was faster than the ego, it tended to leave away, so the risk had a obvious decline.

\subsection{Analysis of pedestrian risk metric}\label{subsec4.3}
 
\begin{figure*}[hptb]
\centering
\includegraphics[width=\linewidth]{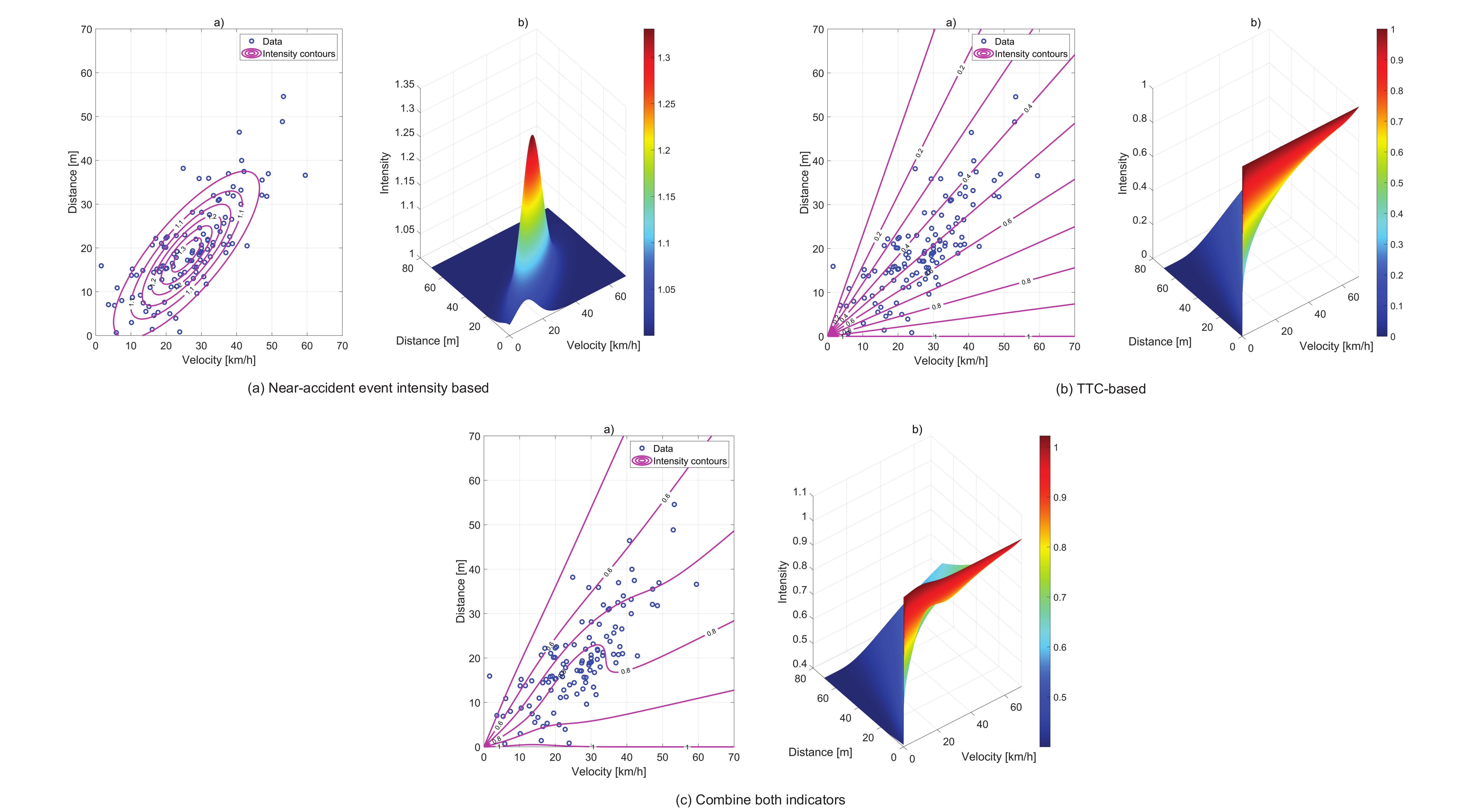}
\caption{Pedestrian risk metric visualization. (a) is the risk metric only considers the near-accident event intensity; (b) is the result only considers the TTC indicator; (c) is the result combines the near-accident event intensity and TTC.}\label{fig43}
\end{figure*}

As described in section \ref{subsec3.3}, the risk intensity of pedestrian risk is a linear function \ref{equ 14} depended on TTC and  the intensity of near-accident event. Fig. \ref{fig43} shows the pedestrian risk distribution for specific simulated data, and the data used for this experiment is generated using the algorithm described in Shen \cite{ShenRaksincharoensak-553}. All the results are got based on this assumption that ego-car is at the position $(0,0)$, and  $x$ coordinate represents the velocity of ego when the pedestrian appears, and $y$ coordinate represents the distance between ego and the pedestrian. The parameter vector of the Pearson model shown in equation \ref{equ 16} in this experiment is:

\begin{equation}
\theta_{p o}=[3.8301, 0.0090, 25.7824, 0.0105,\\
18.3901, 0.7999, 5.7181, 617.3393]^{T}.
\label{equ 20}
\end{equation}

Part $(a)$ shows the risk distribution only considers the near-accident event intensity, part $(b)$ gives the risk distribution based on traditional TTC metric, and part $(c)$ is the final risk metric combined both near-accident event intensity and TTC metric. It is obviously that the metric in $(c)$ can reflect the real word situation better. The risk is most severe when pedestrian is press close to the ego, and when the distance increases, the risk decreases. The risk tend to increase when the speed of ego increases, but high speed of ego also force pedestrian to stop, which will reduce the risk. The highest risk is likely to occur at the position where distance and velocity are in the middle value, which may confuse pedestrians whether they need to stop or pass, then causing ego hard to make right decision. The results of the pedestrian metric satisfy both the result in Shen's paper \cite{ShenRaksincharoensak-553} and the real world situation.

\section{Conclusion}\label{sec5}
This paper proposed a comprehensive driving risk field management framework, which integrates the characteristics of roads, vehicles and pedestrians. Compared to other previous related work, the vehicle risk field model introduced in this paper can reflect the vehicles' motion tendency better and the generated risk field is more close to the real data, which can help autonomous car to avoid over-estimating the risk and making too conservative  driving decision. Meanwhile, a new simple risk metric based CCDF risk curve is introduced to quantify the risk indicator. What's more, a specific form of pedestrian risk model is integrated into the driving risk field model, which combining near-accident event and TTC index. And the pedestrian risk metric is also proposed and the effectiveness has been evaluated. The proposed RCP-RF risk field model can explain the interaction between ego-car and other traffic participants better, and the algorithm time complexity is limited to $O(n^2)$, which is adaptable to real industry scene.

According to the above summary, the following future work can be done to improve the current work:

$1.$ More real work traffic data need to be prepared to verify the effectiveness of the proposed risk model, including different traffic scenarios and different traffic environment, and it will be more convinced if the risk field model can be validated in a real autonomous car driving in a real traffic environment.

$2.$ More factors should be considered into building the risk field model, such as the weather, vehicle kinetic model, road condition, etc. It is clearly that more accurate information we can get and add into the model, the model results can fit the real world better. However, adding more factors will definitely influence the accuracy and effectiveness of the risk model, so how to select the most important factors to analyze the model is a difficult problem.

\section*{Acknowledgments}

This work was supported in part by Key Science and Technology Project of Anhui (Grant No.
202103a05020007) and National Key Research
and Development Program of China (No. 2020AAA0108103).

\section*{Declarations}

\section*{Conflict of interest} 

The authors have no conflicts of interest to declare that are relevant to the content of this article.

\noindent

{\appendix[Algorithm time complexity analysis]
The part will discuss the general algorithm time complexity of the whole model program. Although the algorithm also can be realized by Python and Matlab, for engineering feasibility, the time complexity analysis is based on C++14 programming language code. 
\begin{figure}[h]%
\centering
\includegraphics[width=\linewidth]{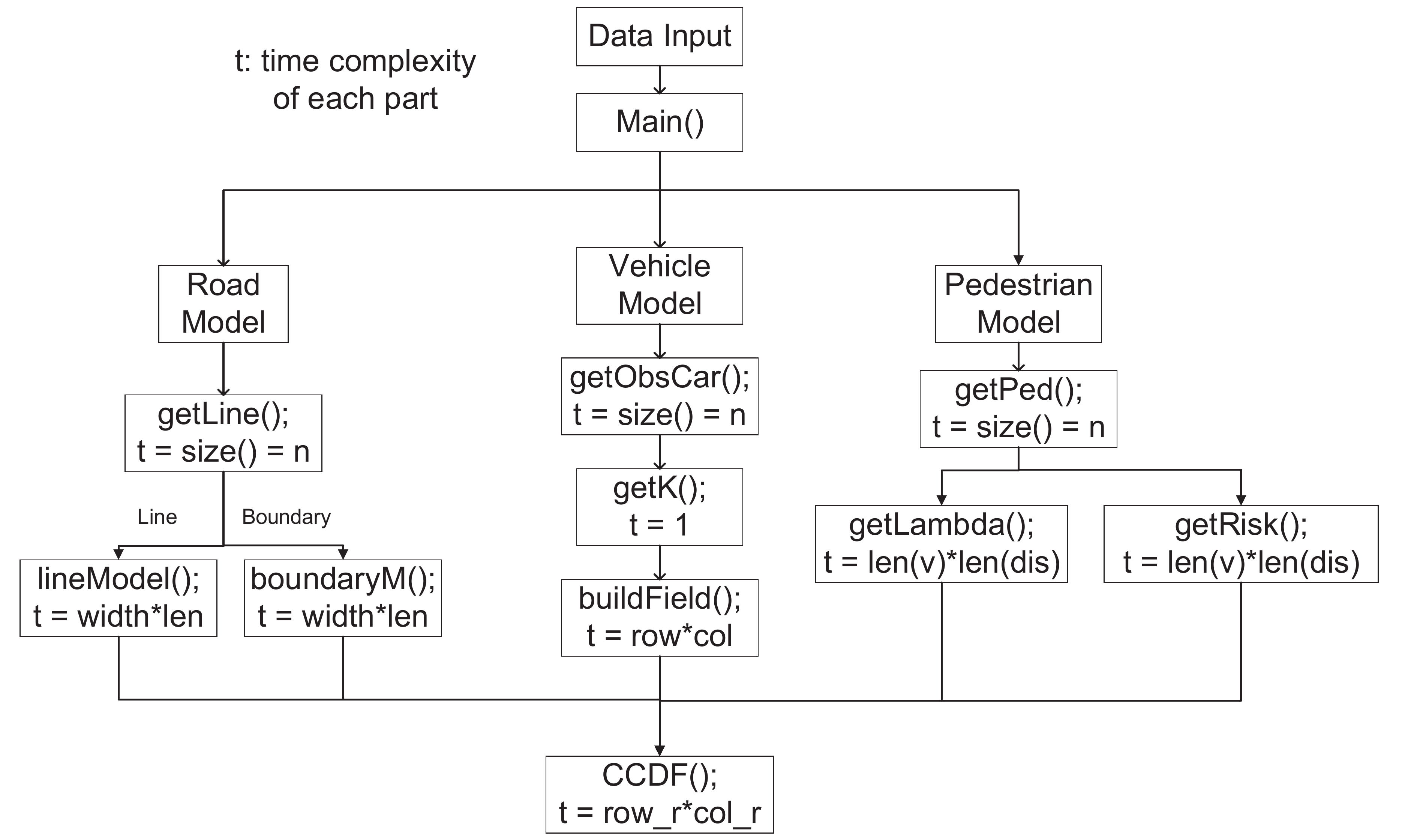}
\caption{Algorithm flow and time complexity for whole risk framework.}\label{fig44}
\end{figure}

Fig. \ref{fig44} shows the rough algorithm flow of the program, and $t$ is the time complexity of each code part. When the needed  input data is got, the program starts at $Main()$ interface, then it divides into three branches based on the type of obstacles.  As shown in the picture, for the road risk model, the time complexity formula can be written as $O(n*width*len)$, where $n$ is the number of detected lines, and $width$ and $len$ are the width and length of the whole lane; for vehicle model, is time complexity is $O(n*1*row*col)$, $n$ is the number of detected obstacle cars, and $row$ and $col$ are the perception map size; for the pedestrian model, the time complexity formula is $O(n*len(v)*len(dis))$, also $n$ is the number of detected pedestrians, $len(v)$ and $len(dis)$ are the number of sample $(v, dis)$ data, which is used to calculate the $\lambda$(the  intensity of near-accident event), this data size is selected by users. For the last part, the calculation of risk metric based on CCDF curve, the time complexity is $O(row_r*col_r)$, $row_r$ and $col_r$ are the map size selected for calculating CCDF. The map is selected to avoid too mush zero values in the risk field to effect the result of CCDF curve. And the range of the selected map is depended on the specific obstacle object and ego-car's position, and the risk distribution range of that object, in order to get the most heuristic risk metric. $row_r$ and $col_r$ are strictly smaller than $row$ and $col$. Therefore, the total time complexity of the program can be expressed as 
{$O(max(n*width*len, n*row*col, n*len(v)*len(dis))+ row_r*col_r)$}, which is depended on the total number of obstacles can be detected and the basic perception range need to be considered for safety.}\label{secA1}

\vfill

\end{document}